\documentclass[10pt,twocolumn,letterpaper]{article}

\usepackage[pagenumbers]{cvpr} 

\usepackage[table,xcdraw,dvipsnames]{xcolor}

\usepackage{color}
\usepackage{caption}
\usepackage{booktabs}
\usepackage{siunitx} 

\newcommand{\rev}[1]{{#1}}

\newcommand{\best}{\cellcolor[HTML]{ff9090}}
\newcommand{\scnd}{\cellcolor[HTML]{ffd088}}
\newcommand{\thrd}{\cellcolor[HTML]{fffeb4}}

\definecolor{bestc}{HTML}{f26060}
\definecolor{scndc}{HTML}{f7b34c}
\definecolor{thrdc}{HTML}{f7f45d}

\definecolor{cvprblue}{rgb}{0.21,0.49,0.74}
\usepackage[pagebackref,breaklinks,colorlinks,citecolor=cvprblue]{hyperref}

\def\paperID{67} 
\def\confName{3DV\xspace}
\def\confYear{2024\xspace}

\title{3D Pose Estimation of Two Interacting Hands from a Monocular Event Camera} 

\author{Christen Millerdurai\textsuperscript{1,2} 
\qquad
Diogo Luvizon\textsuperscript{1} 
\qquad
Viktor Rudnev\textsuperscript{1,2} 
\qquad
André Jonas\textsuperscript{3} \\ 
Jiayi Wang\textsuperscript{1} 
\qquad\quad\;\;
Christian Theobalt\textsuperscript{1} 
\qquad\quad
Vladislav Golyanik\textsuperscript{1} \\
\textsuperscript{1}MPI for Informatics, SIC 
\qquad
\textsuperscript{2}Saarland University, SIC 
\qquad
\textsuperscript{3}RPTU Kaiserslautern-Landau
}

\begin{document}
\maketitle

%
\begin{abstract}
3D hand tracking from a monocular video is a very challenging problem due to hand interactions, occlusions, left-right hand ambiguity, and fast motion. Most existing methods rely on RGB inputs, which have severe limitations under low-light conditions and suffer from motion blur. 
In contrast, event cameras capture local brightness changes instead of full image frames and do not suffer from the described effects. Unfortunately, existing image-based techniques cannot be directly applied to events due to significant differences in the data modalities. 
In response to these challenges, this paper introduces the first framework for 3D tracking of two fast-moving and interacting hands from a single monocular event camera. 
Our approach tackles the left-right hand ambiguity with a novel semi-supervised feature-wise attention mechanism and integrates an intersection loss to fix hand collisions.
To facilitate advances in this research domain, we \rev{release} a new synthetic large-scale dataset of two interacting hands, \mbox{Ev2Hands-S}, and a new real benchmark with real event streams and ground-truth 3D annotations, Ev2Hands-R. 
Our approach outperforms existing methods in terms of the 3D reconstruction accuracy and generalises to real data under severe light conditions\footnote{\url{https://4dqv.mpi-inf.mpg.de/Ev2Hands/}}. 
\end{abstract}

%
%
\section{Introduction}\label{sec:intro} 
\begin{figure}[t]
\centering
   \includegraphics[width=0.99\linewidth]{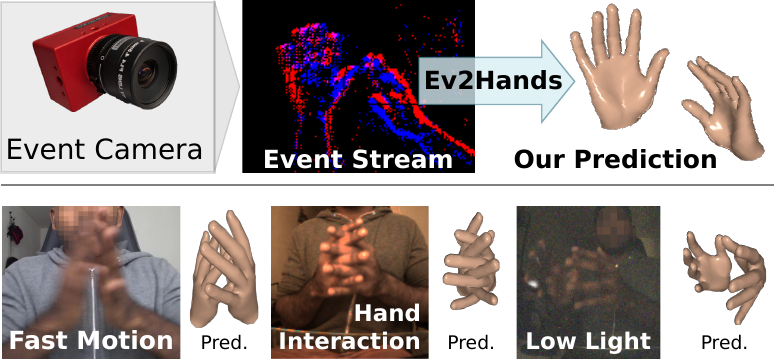}
   \caption{
    We propose \textbf{Ev2Hands}, the first method for the 3D reconstruction of two interacting hands from a single event camera. 
    Our method operates on event clouds and outputs the shape, pose, and position of both hands in 3D space. The RGB images (bottom) are \textit{not used by our method} and are shown only for reference. 
    } 
    \vspace{-0.5cm} 
\label{fig:teaser}
\end{figure}
Live 3D hand tracking from visual streams is a challenging problem arising in many applications \cite{rautaray2015vision,cheok2019review,lei2019applications,al2018systematic,gesture_888469,glove_4539650,DecafTOG2023} such as human-computer interaction and automatic sign language translation, among others. 
Existing works address it predominantly with RGB sensors \cite{Baek2019, boukhayma20193d, Ge2019, hassan2019resolving, wang2020rgb2hands, Moon_2020_ECCV_InterHand2.6M}. 
However, human hands can move fast and can be observed under low-light conditions, which often makes these 3D hand reconstruction scenarios impractical due to apparent motion blur and the limited temporal resolution (or under-exposure) of RGB sensors. 
\par
In contrast to synchronously operating RGB sensors (\textit{i.e.,} recording absolute brightness values of all pixels in a frame at pre-defined time intervals), event cameras record per-pixel brightness changes asynchronously. 
Event cameras have already been successfully used in several complex tasks, including image reconstruction~\cite{rebecq2019high, scheerlinck2018continuous,rudnev2023eventnerf}, optical flow estimation~\cite{paredes2021back}, human action recognition~\cite{sabater2022event, plizzari2022e2}, and human motion capture~\cite{eventCap2020CVPR}, even under low lighting conditions. 
Despite these successes, event-based vision for 3D hand tracking is still in its infancy \cite{TretschkNonRigidSurvey2023}: 
Only a few works~\cite{rudnev2021eventhands, Nehvi2021, xue2022_evnonrigid} addressed it for single hands, and none explicitly accounts for the complex interactions and occlusions frequently occurring during two-hand interactions. 
Note that tracking two hands from asynchronous events is more challenging than the single-hand scenario, as events can be caused by both moving hands simultaneously and by background motion or noise, leading to left-right hand confusion and spurious predictions.
Moreover, there are currently no datasets in the literature with real event stream observations, 3D annotations for both hands, two-hand interactions and fast hand movements. 
\par
In response to the reviewed challenges, this paper proposes Ev2Hands, a new, and the first of its kind, event-based tracker for reconstructing high-speed two-hand interactions in 3D from a single event stream (see Fig.~\ref{fig:teaser}). 
Our new neural architecture operates on spatio-temporally unrolled events {aggregated as event point clouds}, which are segmented in a semi-supervised manner into events caused by {left hand, right hand, or background}. 
{Our method leverages a feature-wise attention mechanism that helps to resolve the left-right event ambiguity, while not requiring ground-truth segmentation labels from real data.}
To encourage plausible 3D estimates and prevent collision and self-penetration in the predicted hand meshes, we also introduce inter-hand and intersection losses. 
Finally, we acquire \textit{Ev2Hands-S}, \textit{i.e.,} a new large-scale dataset with events synthesised from sparse real hand pose annotations, and \textit{Ev2Hands-R}, \textit{i.e.,} a new two-hand pose benchmark with events captured by a real event camera with calibrated and synchronised 3D ground-truth hand poses for both interacting hands. 
\par
In summary, our technical contributions are as follows:
\begin{itemize}\setlength{\itemsep}{1pt}
    \item The first approach for 3D tracking of two interacting hands from a single event camera; 
    \item A new neural architecture with feature-wise attention {coupled with individual event point segmentation that helps to resolve handedness based on a semi-supervised learning strategy};
    \item Two new datasets for training and evaluation: {\mbox{Ev2Hands-S}, derived from sparse 3D hand pose annotations, and Ev2Hands-R, a new real data benchmark} with a synchronised and calibrated event stream, RGB views and 3D hand annotations for two interacting hands.
\end{itemize} 
\par
Our comparisons with the event-based baseline \hbox{EventHands}~\cite{rudnev2021eventhands} and recent RGB-based methods \cite{boukhayma20193d, li2022interacting, Moon_2023_CVPR_InterWild} show that the proposed architecture and the training strategy result in steadily more precise 3D predictions compared to them, especially in challenging cases. 
The readers are urged to watch the accompanying video. 
\section{Related Work}\label{sec:related_work} 
This section focuses on recent approaches for the 3D reconstruction of two hands from a monocular input and reviews event-based methods for 3D vision. 
\noindent\textbf{3D Reconstruction of Two Hands.}
Existing work focuses mainly on single-hand pose estimation from RGB images~\cite{zimmermann2017learning, Mueller_2018_CVPR, Iqbal_2018_ECCV, zhou2019monocular, boukhayma20193d, Baek2019, Zhang2019, Ge2019} or depth maps~\cite{Yuan_2018_CVPR, malik2020handvoxnet, sridhar2015fast}. These methods are often limited by the low frame rate of conventional colour cameras and depth sensors, suffer from blurry images, and fail in the more challenging problem of estimating two interacting hands.
Furthermore, most available datasets are for single-hand estimation from RGB~\cite{Zimmermann2019, Mueller_2018_CVPR, Kulon_2020_CVPR}, depth~\cite{yuan2017bighand2, FirstPersonAction_CVPR2018, OccludedHands_ICCV2017}, or synthetic events~\cite{rudnev2021eventhands}, and only a very few RGB~\cite{Moon_2020_ECCV_InterHand2.6M} and depth~\cite{mueller2019real} datasets provide 3D annotations for both hands.
\par
Some methods for two-hand pose estimation decompose the problem into more tractable tasks~\cite{wang2020rgb2hands, fan2021learning,Moon_2023_CVPR_InterWild} and leverage large-scale annotated data to learn a hand pose predictor~\cite{Moon_2020_ECCV_InterHand2.6M, Moon_2023_CVPR_InterWild}.
Parametric hand models~\cite{MANO:SIGGRAPHASIA:2017, HTML_eccv2020, Moon2020_DeepHandMesh, corona2022lisa} offer the possibility to predict more plausible hands interactions, which can be coupled with intermediate supervisions~\cite{zhang2021interacting}, attention mechanisms~\cite{li2022interacting, hampali2022keypoint}, or modelled in a probabilistic manner~\cite{wang2022handflow}.
Hands segmentation is also particularly useful as an intermediate task for hands estimation from RGB~\cite{fan2021learning} or depth maps~\cite{taylor2017articulated, mueller2019real, li2019point}. However, obtained segmentation from events is a hard problem still in its infancy ~\cite{sun2022ess} with no available dataset for hands.
Differently from previous methods, our approach does not require RGB or depth data and predicts two interacting hands directly from the event stream with extremely high temporal resolution, while enforcing plausible predictions by penalising collisions between interacting hands in 3D space.
\noindent\textbf{Event-based Methods for 3D Reconstruction.}
\begin{figure*}[t]
\begin{center}
   \includegraphics[width=1.0\textwidth]{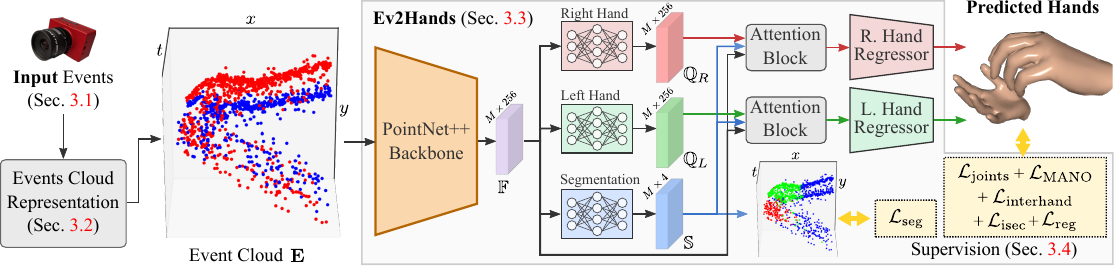}
\end{center}
    \vspace{-10pt} 
    \caption{
        \textbf{Workflow of our Ev2Hands approach}. Our framework converts a time-sliced event stream into an \textit{event cloud}, which is fed as input to Ev2Hands that regresses the MANO parameters of the left and right hands with their global rotations and translations. The grey blocks are non-trainable and dashed lines indicate components only used during training. Note that the segmentation branch is only supervised with synthetic data and trained in a semi-supervised manner on real data.
    } 
\label{fig:ev2hands}
\end{figure*}
Traditional image-based techniques cannot be directly applied to events. 
To cope with this, Xu \textit{et al.}~\cite{eventCap2020CVPR} formulates a model-fitting term using a collection of ``close events'' to refine an initial pose. 
Other methods use \textit{event frames}, \textit{i.e.,} an image-space representation by aggregating events from a fixed time interval. 
This enables the usage of learning-based techniques that take advantage of the inductive bias of CNNs. 
Thus, 
Rudnev \etal~\cite{rudnev2021eventhands} presented the first method for the estimation of a single 3D hand \rev{pose} from events with the proposed LNES representation and a temporal Kalman filtering stage. LNES aggregates events in a 2D image considering a temporal sliding window\footnote{While designing Ev2Hands, we observed that LNES does not work well for two hands due to its inherent ambiguities when the hands occlude each other, and we had to utilise another representation (\textit{i.e.,} event cloud).}. 
Nehvi \etal~\cite{Nehvi2021} track non-rigid 3D objects by propagating through a new differentiable event stream simulator and Xue \etal~\cite{xue2022_evnonrigid} presents an Expectation Maximisation (EM) framework where the parameters of a hand model are optimised by associating events to the mesh faces, assuming that events are typically caused by moving edges.
However, such methods do not take advantage of the sparsity of the event streams and must process each 
event image entirely. 
Additionally, they are limited to a single hand.
When multiple hands are interacting in the scene, events triggered by different hands are entangled in the event frame representation, making it harder to estimate both hands separately.
\par
One way to preserve data sparsity is to represent event streams as a space-time Event Point Cloud (EPC).
This representation has been applied for gesture recognition~\cite{Wang_WACV_2019}, where event points are processed by PointNet~\cite{qi2017pointnet}.
Chen~\etal~\cite{chen2022EPP} extended EPC with a Rasterised Event Point Cloud (REPC) representation for 2D human pose estimation. 
This includes a re-sampling strategy to ensure that PointNet can operate on fixed time windows. 
Our approach leverages an EPC representation and processes it with PointNet++~\cite{qi2017pointnetplusplus} to solve the much more complex task of \emph{3D reconstruction} of two self-similar hands. Differently from previous methods, we propose a feature-wise attention mechanism coupled with semi-supervised training that benefits from synthetic event segmentation and real data without segmentation labels, which helps the network to focus on only relevant events while generalising to real event streams. 
%

\section{Method} \label{sec:method}
Given a monocular event stream (Sec.~\ref{subsec:evcameramodel}) observing two interacting hands, we reconstruct the shape, pose, and position of each hand in the global 3D space in two stages. 
First, we process the raw monocular event stream into an event point cloud representation (Sec.~\ref{subsec:eventcloudrepresentation}), which preserves relevant information for a given time window while compressing redundant events from the same pixel in a single point.
Second, we propose \textit{Ev2Hands} (Sec.~\ref{sec:architecture}), an end-to-end attention-based neural network that takes as input an event cloud and regresses the hand model parameters~\cite{MANO:SIGGRAPHASIA:2017} along with the global translation and rotation of the left and right hands.
The key component in the proposed architecture is a feature-wise attention module that learns to use semi-supervised segmentation features fed to hand-specific regressors. 
See Fig.~\ref{fig:ev2hands} for the method overview. 
\subsection{Event Camera Model} \label{subsec:evcameramodel}
Event cameras produce a stream of asynchronous events used to record changes in brightness. These \textit{raw events} are represented by tuples
\begin{equation}
\label{eqn:rawevents} 
\mathbf{e}_i=(x_i,y_i,t_i,p_i),
\end{equation}
where the $i$-th event corresponding to the pixel location $(x_i,y_i)$ is triggered at time $t_i$ with polarity $~{p_i\in\{-1,1\}}$.
An event is emitted at time $t_i$ when the change in logarithmic brightness $L$ crosses the threshold $C$, \ie, $|L(x_i,y_i,t_i)-L(x_i,y_i,t_i-t_p)|\geq{C}$,  where $t_p$ is the previous triggering time at the same pixel.
Due to the sparse, asynchronous nature of the event stream, it is challenging to directly process the raw event stream using a neural network. 
Next, we present an event cloud representation that is better suited for training our network architecture. 
\subsection{Event Cloud Representation} \label{subsec:eventcloudrepresentation} 
Our goal in this stage is to process the input stream of asynchronous events in a more stable and efficient representation.
Previous works employ 2D representations of events by projecting the temporal information to the image plane
~\cite{maqueda2018event, rudnev2021eventhands, rebecq2017real} so that CNNs can be directly applied.
However, this aggregation collapses the temporal information and creates \rev{inefficient} and sparse image representations. Alternatively, we treat time as a third data dimension and conceptualize events as point clouds in a similar manner to REPC~\cite{chen2022EPP}. 
\par
Specifically, let us consider a time window of size $T$ where the first raw event $\mathbf{e}_0$ occurs at time $t_0=0$ (relative to the given time window) and all events at the same pixel location $(x,y)$ are combined into an event point $\mathbf{E}_k$: 
\begin{equation}
\label{eq:event_cloud} 
\mathbf{E}_k = (x_k, y_k, t_k, P_k, N_k),
\end{equation}
where $t_k$ is the average time of the combined events and $P_k$ and $N_k$ are the number of positive and negative events in the time interval considered and normalised by the total number of events in the pixel. 
When all the raw events in the time window $T$ are combined, we obtain an \textit{event cloud} $\mathbf{E}\in\mathbb{R}^{M\times5}$, where $M$ is the resulting number of events.
\subsection{The Proposed Ev2Hands Approach} \label{sec:architecture}
Given the event cloud $\mathbf{E}$, we estimate the shape, pose, and position of each hand corresponding to the end of the time window $T$. 
In what follows, we provide the details of the hand model we use and the proposed attention-based neural network for \rev{processing $\mathbf{E}$} for both hands. 
\subsubsection{3D Hand Model} \label{subsubsec:handmodel} 
We use MANO \cite{MANO:SIGGRAPHASIA:2017} for human hand mesh parameterisation, which 
includes 
a hand pose vector $\boldsymbol \theta \in \mathbb{R}^6$ and shape vector $\boldsymbol \beta \in \mathbb{R}^{10}$; both vectors are coefficients obtained through PCA decomposition. 
We also encode the rigid transformation parameters, \textit{i.e.,} translation $\textbf{t} \in \mathbb{R}^3$ and the rotation  $\mathbf{R} \in \mathbb{R}^3$ for each hand. 
We can obtain the sparse hand joints from MANO with $~{\mathbf{J}=\mathcal{J}(\boldsymbol \theta, \boldsymbol \beta, \textbf{t}, \mathbf{R})}$, where $\mathcal{J}$ is a function that regresses the joint locations and applies rotation and translation, and $\mathbf{J}\in\mathbb{R}^{N_J\times{3}}$ are the joint locations of the regressed 3D hand.
For simplicity, we refer to the hand parameters in the same way for both left and right hands, unless explicitly indicated otherwise.
\subsubsection{Ev2Hands Model} \label{subsubsec:ev2hands} 
Considering that both hands and the background can trigger independent events and some events are spurious (noisy signals), we need a model that can determine which events are more relevant to each hand prediction and which events should be ignored. 
In addition, the model architecture has to be specifically designed to handle the event cloud input and to predict the parameters for both hands, as previously discussed.
To achieve this goal, we leverage PointNet++~\cite{qi2017pointnetplusplus} as our backbone, which takes as input the event cloud $\mathbf{E}$ and outputs the \textit{event features} $\mathbb{F}\in\mathbb{R}^{M\times{256}}$.
The event features are then processed by a multilayer perceptron (MLP) that produces individual features 
relevant for the segmentation task and the hand regression tasks. 
A feature-wise attention block is individually applied to the features from both hands, where the predicted segmentation labels are used as keys. This allows the two-hand regressor models to take segmentation results into account to predict the left- and right-hand parameters.
A diagram of Ev2Hands is shown in Fig.~\ref{fig:ev2hands}.
In what follows, we explain its components. 
\par
\noindent \textbf{Hand Branches}.
Given the event features $\mathbb{F}$, we want to obtain a per-hand feature vector that will be further used to compute hand-specific features. To achieve this, we use the left- and right-hand branches, respectively depicted in green and red in Fig.~\ref{fig:ev2hands}, which extract two feature vectors  $\mathbb{Q}_{L},\mathbb{Q}_{R}\in\mathbb{R}^{M\times{256}}$ through two shallow MLP networks that are individually applied to each point in $\mathbb{F}$. 
\par
\noindent \textbf{Segmentation Branch}.
In addition to the hand-specific features, we want our model to reason about whether points belong to the left hand, right hand, or background events.
Hence, we introduce a segmentation branch, which also uses a shallow MLP applied to each point in $\mathbb{F}$. Differently from the hand branches, the segmentation branch predicts the logits associated with each point in the event cloud, which are represented by $\mathbb{S}\in\mathbb{R}^{M\times{4}}$ and encode the classes \textit{left hand}, \textit{right hand}, \textit{background}, and \textit{no class}. 
The left and right-hand labels correspond to events directly produced by one of the hands and the background labels correspond to events produced by non-hand objects, like torso or arm movements, or by changes in the background. The extra \textit{no class} label is used to indicate when an event point combines multiple events with different labels in the time window, \eg when a left-hand event and a background event are triggered at the same pixel in the same time window.
\par
\noindent \textbf{Feature-wise Attention Block}.
Inspired by attention mechanisms \cite{bahdanau2014neural, AttentionisAllYouNeed}, we want our model to extract features that are relevant to each hand individually. To this end, we have a feature-wise attention module defined by:
\begin{equation}
\small
\label{eqn:attnetioneqn}
  \mathrm{Attention}(\mathbb{Q}_{(\cdot)}, \mathbb{S}, \mathbb{F})=%
  \mathbb{F}\left(\mathrm{Softmax}\left(\frac{\mathbb{Q}_{(\cdot)}^{T}\mathbb{S}}{\sqrt{d_s}}\right)\right),
\end{equation}
where the hand features, $\mathbb{Q}_L$ or $\mathbb{Q}_R$, are masked by the key values $\mathbb{S}$, and the \textit{Softmax} operates as a linear combination of the event features $\mathbb{F}$. Note that Eq.~\eqref{eqn:attnetioneqn} is applied individually to each hand, which produces a two hand attention features $\mathbf{H}_L, \mathbf{H}_R \in \mathbb{R}^{M\times{d_s}}$; $d_s$ is the dimension of $\mathbb{S}$.
\par
The feature-wise attention mechanism is designed to allow the attention features $\mathbf{H}$ to be a function of the hand features $\mathbb{Q}$, but also to respond to the segmentation predictions from $\mathbb{S}$. This helps the model to identify features that are more relevant to specific hands or conditions of interactions, as demonstrated in our experiments.
In addition, we also show that the feature-wise attention mechanism can refine the segmentation prediction $\mathbb{S}$, which is semi-supervised with synthetic data and trained end-to-end on real event data by supervising the hand parameters only.
\par 
\noindent \textbf{Hand Parameters Regressor}.
Finally, given the hand-specific attention features, we predict the parameters $\{\boldsymbol \theta$, $\boldsymbol \beta$, $\textbf{R}$, $\textbf{t}\}$ for each hand. For this task, we use a \rev{\mbox{mini-PointNet}} \cite{qi2017pointnet} followed by an MLP. Since each hand regressor takes as input its own hand-specific features, there is no weight sharing between both models.
\subsection{Loss Functions}\label{sec:losses} 
\par
\noindent \textbf{3D Joint Loss}. This loss penalises deviations between the regressed 3D joints from the hand model and the ground truth 3D joint annotations of each hand. For simplicity, we omit the left- and right-hand indexes:
\begin{equation}
\label{eq:lossjoints}
\mathcal{L}_{\text{joints}} = \frac{1}{N_J}\sum_{i=1}^{N_J} \lVert \hat{\mathbf{J}}_{i} - \mathbf{J}_{i} \rVert,
\end{equation}
where $\hat{\mathbf{J}}_{i}$ and $\mathbf{J}_{i}$ are the predicted and ground-truth $i$-th 3D joints, and $N_J$ is the number of hand joints.
\par
\noindent \textbf{MANO Loss}. This term penalises deviations between the predicted and reference hand parameters. The PCA pose and shape coefficients $\boldsymbol \theta$ and $\boldsymbol \beta$, and the global rotation $\textbf{R}$ are penalised using $\ell_2$ distance, while the rigid translation $\textbf{t}$ is supervised with using $\ell_1$ distance:
\begin{equation}
\mathcal{L}_{\text{MANO}}=%
  \lVert \hat{\boldsymbol \theta} - \boldsymbol \theta \rVert^2%
  +\lVert \hat{\boldsymbol \beta} - \boldsymbol \beta \rVert^2%
  +\lVert \hat{\mathbf{R}} - \mathbf{R} \rVert^2%
  +\lVert \hat{\mathbf{t}} - \mathbf{t} \rVert,
\end{equation}
where $\hat{\boldsymbol \theta}, \hat{\boldsymbol \beta}, \hat{\mathbf{R}}, \hat{\mathbf{t}}$ are predicted by the hand regressors and $\boldsymbol \theta, \boldsymbol \beta, \mathbf{R}, \mathbf{t}$ are the reference hand parameters.
\par
\noindent \textbf{Segmentation Loss.} The segmentation loss is intermediate supervision used to penalise wrong event classes in $\mathbb{S}$, which is the case in our experiments \textit{only} when the model is trained on synthetic data. When training on real event data, the segmentation branch is indirectly supervised only by the gradients propagated from the supervision of the hand parameters. The segmentation loss reads: 
\begin{equation}
\mathcal{L}_{\text{seg}} = \mathrm{CrossEntropy} (\mathrm{Softmax}(\mathbb{S}), \mathbf{c}),
\end{equation}
where $\mathbf{c}$ are the event class labels considering \textit{left hand}, \textit{right hand}, and \textit{background}. Note that the key corresponding to \textit{no class} is not supervised.
\par
\noindent \textbf{Inter-hand Loss}. This loss term considers both hands simultaneously and penalises deviations between left- and right-hand shape parameters and the relative position between both hands. The inter-hand loss is expressed as:
\begin{equation}
\mathcal{L}_{\text{interhand}} =%
  \lVert \boldsymbol \beta_{\text{left}} - \boldsymbol \beta_{\text{right}} \rVert ^2
+ \mathcal{I}_J 
+ \mathcal{I}_T,\;\text{where}
\end{equation}
\begin{equation}
\mathcal{I}_J =%
  \frac{1}{N_J}\sum_{i=1}^{N_J} \lVert (\hat{\mathbf{J}}_{\text{left},i} - \hat{\mathbf{J}}_{\text{right},i}) -  (\mathbf{J}_{\text{left},i} - \mathbf{J}_{\text{right},i}) \rVert ^2
\end{equation}
\begin{equation}
\mathcal{I}_T = %
  \lVert (\hat{\textbf{t}}_{\text{left}} - \hat{\textbf{t}}_{\text{right}}) - (\textbf{t}_{\text{left}} - \textbf{t}_{\text{right}}) \lVert ^2.
\end{equation}
The inter-3D joint $\mathcal{I}_J$ and the inter-translation $\mathcal{I}_T$ terms account respectively for the relative articulation errors and the relative distance errors considering the left and right hands.
\par
\noindent \textbf{Intersection Loss}. We avoid physically invalid predictions by penalising intersections, both due to articulation and hand-hand interactions. 
We adopt the conic distance fields approximation of meshes~\cite{Tzionas:IJCV:2016} 
for our collision loss. 
This is done by first finding the set of colliding triangles using bounding volume hierarchies \cite{Karras:2012:MPC:2383795.2383801}. 
For each triangle, a 3D cone is constructed, defined by a circumscribing circle and the face orientation. 
The distance to the surface of the cone can be calculated for each query point, and the sum of these distances over all triangles under consideration approximates the distance field of the hand. 
The value of the distance field represents the amount of the repulsion $\mathcal{L}_{\text{isec}}$ that is needed to penalise the intrusion.
For the exact definition of $\mathcal{L}_{\text{isec}}$, we refer the reader to Tzionas \etal~\cite{Tzionas:IJCV:2016}. 
\par
\noindent \textbf{MANO Regularisation}. In addition to the losses introduced above, we also regularise the hand predictions in the PCA parameter space of MANO through the use of Tikhonov regulariser~\cite{mueller2019real}:
\begin{equation}
\mathcal{L}_{reg} = %
  \lambda_{\theta} \lVert \hat{\boldsymbol \theta} \rVert^2 
  + \lambda_{\beta} \lVert \hat{\boldsymbol \beta} \rVert^2,
\end{equation}
where $\lambda_{\theta}=0.025$ and $\lambda_{\beta}=25$. %
This term penalises statistically unlikely MANO parameters.
\par
\noindent \textbf{The Total Loss.} Overall, our total loss reads: 
\begin{multline}
\mathcal{L} = 
  \lambda_{\text{joints}} \mathcal{L}_{\text{joints}} 
+  \lambda_{\text{MANO}} \mathcal{L}_{\text{MANO}}
+  \lambda_{\text{seg}} \mathcal{L}_{\text{seg}}\\
+  \lambda_{\text{interhand}} \mathcal{L}_{\text{interhand}}
+  \lambda_{\text{isec}} \mathcal{L}_{\text{isec}}
+  \mathcal{L}_{\text{reg}},
\end{multline}
where the weights $\lambda_{\text{joints}}{=}0.01$, $\lambda_{\text{MANO}}{=}10$, $\lambda_{\text{seg}}{=}1$, $\lambda_{\text{interhand}}{=}100$, $\lambda_{\text{isec}}{=}100$ are chosen empirically to account for the different magnitudes of each loss term.
\par
\noindent \textbf{Training on Real Data.}
Although our large-scale synthetic data\rev{set} (Sec. \ref{subsec:syntheticdataset}) contains a high variability of poses, our model can further benefit from fine-tuning on real event streams to reduce the domain gap.
However, real data does not provide event segmentation labels and the intrinsics from DAVIS346C can differ from the synthetic data.
To cope with the first problem, we remove $\mathcal{L}_{\text{seg}}$ from our losses and indirectly learn 
$\mathbb{S}$ based only on the hand parameters supervision.
To mitigate the variations in the intrinsics that could cause discrepancies in 3D joint positions, we replace the 3D joint loss with a 2D projection joint loss:
\begin{equation}
\begin{split}
\mathcal{L}_{\text{real}} = &
  \lambda_{\text{joints2D}} \mathcal{L}_{\text{joints2D}} + \lambda_{\text{interhand}} \mathcal{L}_{\text{interhand}} + \\
 & \lambda_{\text{isec}} \mathcal{L}_{\text{isec}}
+  \mathcal{L}_{\text{reg}},\;\text{where}\; 
\label{eq:real_loss}
\end{split}
\end{equation}
\begin{equation}
\mathcal{L}_{\text{joints2D}} = \frac{1}{N_J}\sum_{i=1}^{N_J} \lVert \Pi_{S}({\hat{\mathbf{J}}_{i}}) - \Pi_{R}(\mathbf{J}_{i}) \rVert,
\end{equation}
and $\Pi_{S}$ and $\Pi_{R}$ are the camera projection operators for the simulated and real event cameras.
$\mathcal{L}_{\text{real}}$ enables training on real data while the semi-supervised segmentation branch is still optimised through our attention mechanism.
\section{The Ev2Hands Datasets}\label{sec:dataset} 
\begin{figure}[t]
\centering
   \includegraphics[width=0.99\linewidth]{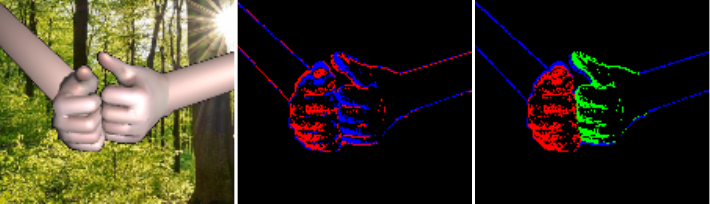}
   \caption{
    Sample from Ev2Hands-S with input image (left), event stream (middle), and segmentation labels (right).
} 
\label{fig:dataset_synt}
\end{figure}

\begin{figure}[t]
\centering
   \includegraphics[width=0.99\linewidth]{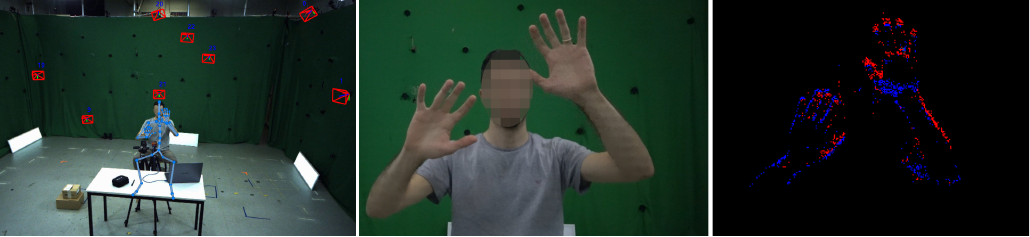}
   \caption{
    Sample from Ev2Hands-R with motion tracking setup (left) used for obtaining the reference hand poses, RGB stream (middle), and event stream (right).
} 
\label{fig:dataset_real}
\end{figure}

To train our model, we need to provide event data and labels to supervise the loss functions described in Sec.~\ref{sec:losses}.
However, \rev{no} dataset with annotated two-hand interactions and an event stream is available.
To solve this issue, we synthesise a large-scale event stream dataset and record a real-world dataset using one DAVIS346C event camera. Both datasets have synchronised RGB videos along with the event streams. We refer to the synthetic and real datasets as \textit{Ev2Hands-S} and \textit{Ev2Hands-R}. See Figs.~\ref{fig:dataset_synt} and \ref{fig:dataset_real} for examples of our data.
To the best of our knowledge, these are the first event stream datasets to model two-hand interactions. 
Please also see our video for dynamic visualisations.
We next provide details on our datasets. 
\subsection{Ev2Hands-S Dataset} \label{subsec:syntheticdataset} 
We generate our synthetic dataset by rendering synthetic videos of two interacting hands, which are then fed to the event stream simulator VID2E~\cite{Gehrig_2020_CVPR}. 
\par
\noindent \textbf{Interacting Hand Animation}. We obtain realistic hand motion by leveraging the InterHand2.6M dataset~\cite{moon2022neuralannot,Moon_2020_ECCV_InterHand2.6M}, providing ground-truth MANO parameters for two hands. 
For each sequence from InterHand2.6M, we linearly interpolate the annotations to obtain smooth sequences with a higher framerate and to fill in possible gaps in the annotations (which are mainly due to sporadically missing ground truth). 
This also allows us to vary the animation frame rate by re-sampling the sequences, which provides more variability to the training data. 
In total, we obtain $3.12 \cdot 10^{8}$ events from the simulated videos. 

\noindent \textbf{Scene Modelling}. We use MANO~\cite{MANO:SIGGRAPHASIA:2017} to model the hands and attach cylinders onto the base to fit the forearms. We texture the hands with HTML~\cite{HTML_eccv2020} and render the models with Pyrender~\cite{pyrender}.
Please see the supplementary material for more details.
\par
\noindent \textbf{Rendering and Event Stream Generation}. We render the scenes using a perspective camera with \ang{30} as a vertical field of view and a resolution of $346 \times 240$ to emulate the specifications of DAVIS346C~ \cite{davis_346}. We apply the RGGB Bayer filter~\cite{rebecq2019high} to the rendered image, as we use a colour event camera in our experiments. We convert the ``Bayered'' image $B_{i}$ at frame $i$ to log intensity $L(t_i)$ at time $t_i$ by:
\begin{equation}
L(t_i) = \log(B_{i} + \epsilon), 
\end{equation}
where $\epsilon$ is set as $10^{-4}$ for numerical stability. We pass the log intensity image to an event stream simulator~\cite{Gehrig_2020_CVPR} with the event threshold $C$. We estimate $C$ to match our DAVIS346C event camera following the procedure adopted by Rudnev \etal~\cite{rudnev2021eventhands} and obtain $C=0.4$.
\subsection{Ev2Hands-R Dataset}  \label{subsec:realdataset} 

\begin{table*}[htbp] 
\centering
\resizebox{\textwidth}{!}{
\begin{tabular}{@{}lcccccccccccccccccccc@{}}
\toprule
& \multicolumn{2}{c}{Palm wave} & \multicolumn{2}{c}{Dorsal wave} & \multicolumn{2}{c}{Wrist rot.} & \multicolumn{2}{c}{Articulation} & \multicolumn{2}{c}{Clap} & \multicolumn{2}{c}{Intersection} &  \multicolumn{2}{c}{Occlusion} & \multicolumn{2}{c}{Free Style} & \multicolumn{2}{c}{Avg.}\\
& \scriptsize{R-AUC} & \scriptsize{RR-AUC} & \scriptsize{R-AUC} & \scriptsize{RR-AUC} & \scriptsize{R-AUC} & \scriptsize{RR-AUC} & \scriptsize{R-AUC} & \scriptsize{RR-AUC} & \scriptsize{R-AUC} & \scriptsize{RR-AUC} & \scriptsize{R-AUC} & \scriptsize{RR-AUC} & \scriptsize{R-AUC} & \scriptsize{RR-AUC} & \scriptsize{R-AUC} & \scriptsize{RR-AUC} & \scriptsize{R-AUC} & \scriptsize{RR-AUC} \\ \cmidrule(lr){2-19}
\dag \text{Boukhayma \etal \cite{boukhayma20193d}}$^*$  & 0.50 & --   & \scnd{0.49} & -- & 0.33 & -- & 0.45 & -- & 0.29 & -- & 0.27 & -- & 0.28 & -- & 0.45 & -- & \thrd{0.38} & -- \\ 

\dag \text{Li \etal~\cite{li2022interacting}}          & \best{0.67} & \scnd{0.43} & \best{0.69} & \scnd{0.42} & \thrd{0.57} & \thrd{0.34} & \thrd{0.62} & \scnd{0.43} & \thrd{0.59} & \scnd{0.41} & \thrd{0.51} & \thrd{0.38} & \thrd{0.55} & \thrd{0.37} & \best{0.63} & \scnd{0.46} & \scnd{0.60} & \scnd{0.40} \\ 

\dag \text{Moon \etal~\cite{Moon_2023_CVPR_InterWild}}          & \thrd{0.65} & \thrd{0.33} & \best{0.69} & \thrd{0.35} & \best{0.67} & \thrd{0.34} & \best{0.65} & 0.35 & \scnd{0.6} & \thrd{0.36} & \best{0.62} & \scnd{0.42} & \best{0.64} & \scnd{0.44} & \thrd{0.61} & \thrd{0.33} & \best{0.64} & \thrd{0.36} \\ \hline

\text{EventHands~\cite{rudnev2021eventhands}}$^*$       & 0.41 & 0.31 & \thrd{0.48} & 0.21 & 0.31 & \scnd{0.40} & 0.43 & \thrd{0.42} & 0.45 & 0.29 & 0.28 & 0.35 & 0.28 & 0.33 & 0.43 & 0.28 & \thrd{0.38} & 0.32 \\

\text{\textbf{Ev2Hands (Ours)}}                         & \scnd{0.66} & \best{0.53} & \best{0.69} & \best{0.54} & \scnd{0.63} & \best{0.52} & \scnd{0.63} & \best{0.52} & \best{0.65} & \best{0.54} & \scnd{0.60} & \best{0.50} & \scnd{0.63} & \best{0.54} & \scnd{0.62} & \best{0.50} & \best{0.64} & \best{0.52} \\
\bottomrule
\end{tabular}
} 
\caption{
Comparison on Ev2Hands-R. ``\dag''\hspace{0.02cm} denotes RGB-based and ``*'' denotes single-hand methods. Ev2Hands outperforms existing approaches in most of the activities by a fair margin while estimating hands with a much higher temporal resolution. 
\textcolor{bestc}{\textbf{Red}}, \textcolor{scndc}{\textbf{orange}} and 
\textcolor{thrdc}{\textbf{yellow}} denote the highest, the second-highest and the third-highest AUC scores, respectively. 
}
\label{tab:sotabenchmark_full} 
\end{table*} 

To evaluate our method with an actual event camera and to further bridge the gap between synthetic and real data, we collected the Ev2Hands-R dataset.
\par
\noindent
\textbf{Dataset composition}.
The dataset focuses on the everyday usage of hands with a high range of motions and variations among the participants, who were instructed to perform a set of actions in an unconstrained manner. 
We recorded in total eight sequences with five different subjects, including persons of different sizes and different skin colours, resulting in variations in the shape and aspect of the hands.
The sequences depict various hand motions performed with progressively increasing complexity: \textit{palm wave}, \textit{dorsal wave} (back of the hand), \textit{wrist rotation}, \textit{hand articulation}, \textit{clap}, \textit{intersection}, \textit{occlusion}, and \textit{free style}.
\par
\noindent
\textbf{Data Capture Setup}. 
We capture all sequences with event camera DAVIS346C and a high-speed RGB camera Sony RX0. 
The reference 3D hand poses are obtained by a commercial multi-view human motion tracking system~\cite{captury} with $29$ external cameras (see Fig.~\ref{fig:dataset_real}). All the $31$ cameras are synchronised and jointly calibrated with a reprojection error below $3$mm (which is substantially more accurate than the accuracy of monocular 3D hand pose estimation techniques). 
The acquired events and RGB frames (along with the corresponding 3D hand poses) span $20.1$ minutes. 
\section{Experiments}\label{sec:experiments} 
We evaluate Ev2Hands on two-hand 3D reconstruction considering Ev2Hands-S and Ev2Hands-R datasets.
\par
\noindent \textbf{Ev2Hands-S Dataset}. We follow the original train and test split from InterHand2.6M~\cite{Moon_2020_ECCV_InterHand2.6M} and evaluate our method in the test split using the synthesised event stream as input. We compare our predictions with the interpolated reference poses, as described in Sec.~\ref{subsec:syntheticdataset} and in the supplement. 
\begin{figure*}[t]
  \centering
  \includegraphics[width=1.0\textwidth]{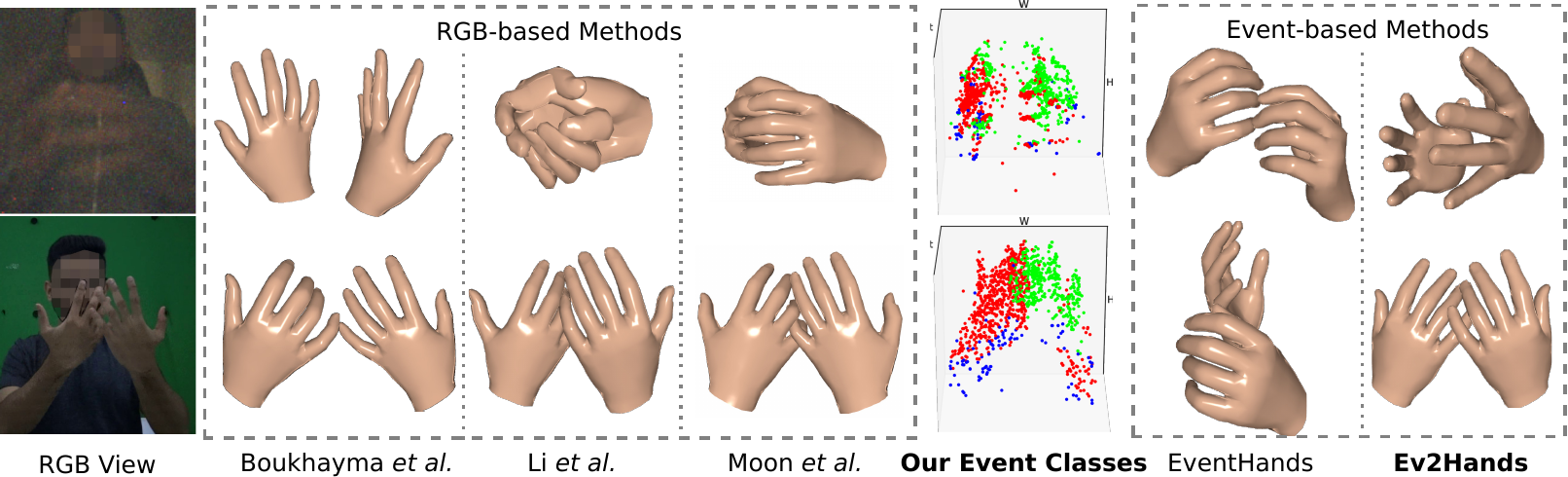}
  \caption{
  A qualitative comparison of our Ev2Hands method with previous RGB-based \cite{boukhayma20193d, li2022interacting, Moon_2023_CVPR_InterWild} and 
  event-based \cite{rudnev2021eventhands} methods. Note how EventHands and Boukhayma \etal~fail to predict interacting hands and how Li \etal~fail with low-light RGB. 
  } 
\label{fig:qualitative}
\end{figure*}
\par
\noindent \textbf{Ev2Hands-R Dataset}. We use two subjects to fine-tune our method on real data and three subjects for evaluation.
To establish a reference for the performance metrics, we evaluate {three} RGB methods that claim to work on in-the-wild data~\cite{li2022interacting, boukhayma20193d, Moon_2023_CVPR_InterWild} on the synchronised RGB video stream.
Note that although we only evaluate event predictions that have corresponding RGB frames, our method works even when RGB methods have no input due to the much higher temporal resolution of our method. 
This advantage is not reflected in the performance metrics.
\par
\noindent \textbf{Evaluation Metrics.} For our quantitative comparisons, we use the Percentage of Correct Keypoints (PCK) and the area under the PCK curve (AUC) with thresholds ranging from 0 to 100 millimetres (mm). Following Li \textit{et al.}~\cite{li2022interacting}, we report the relative PCK (R-PCK) and AUC (R-AUC) scores to evaluate the performance of 3D hand pose estimation of each hand individually. To evaluate the performance of localisation of each hand with respect to the other, we extend the mean relative-root position error  \cite{Moon_2020_ECCV_InterHand2.6M} and report the relative-root PCK (RR-PCK) and AUC (RR-PCK) scores.
Unlike R-PCK, which is computed after performing root-joint alignment of each hand individually, the RR-PCK aligns the entire two-hand configuration to the right-hand root of the reference pose.
Thus, the RR-PCK metric better evaluates the relative 3D position of the hands.
To evaluate the mesh penetration of interacting hands, we take the collision percentage "Coll\%" of each mesh. This metric takes the percentage of mesh triangles that intersect with each other. 
We report the mean over frames where the hands are positioned less than $50$mm to each other, and a lower value indicates less mesh penetration of interacting hands. 
\par
\noindent \textbf{Implementation.} 
We implement our network in PyTorch \cite{paszke2019pytorch} and use Adam optimiser~\cite{kingma2014adam} with a learning rate of $5 \cdot 10^{-5}$ and a mini\rev{-}batch of $128$.
We train for $8 \cdot 10^{5}$ iterations with \textit{Ev2Hands-S} and fine-tune for $1.5 \cdot 10^{4}$ iterations on \textit{Ev2Hands-R}.
The network is supervised by the reference pose considering the last frame in the time window $T$, emulating the position of hands closest to the current time. 
The temporal resolution of the event stream is set to $1000$~FPS, achieved by using a $2$~ms window time with $1$~ms overlap.
\subsection{Comparisons to State of the Art}
\begin{figure}[t]
\centering
   \includegraphics[width=0.99\linewidth]{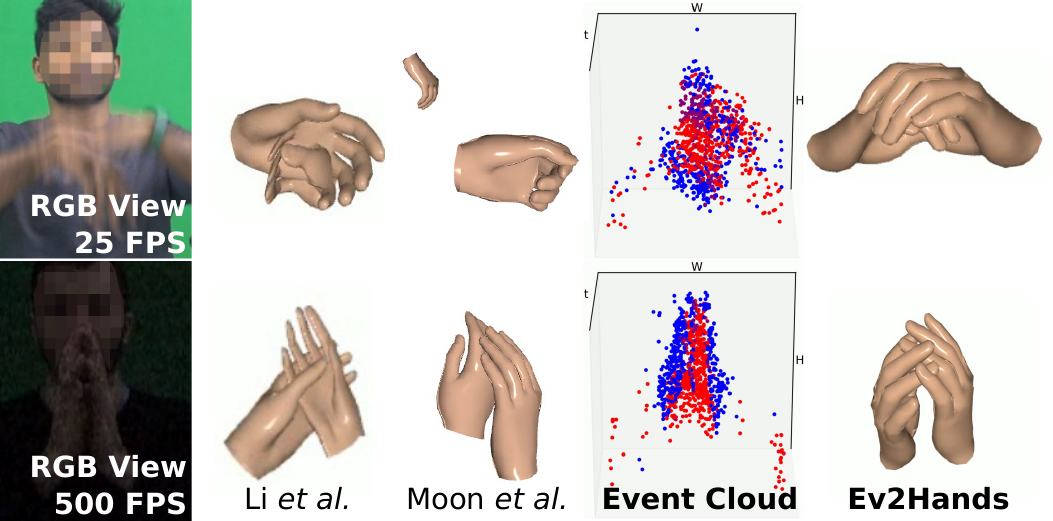}
   \caption{{Our event-based method performs well, whereas RGB-based methods~\cite{li2022interacting, Moon_2023_CVPR_InterWild} fail, notably due to motion blur and scene underexposure under low lighting conditions. 
}} 
\label{fig:qualitative_rgb_events}
\end{figure}

In Table~\ref{tab:sotabenchmark_full}, we compare our method with state-of-the-art pose estimation methods on Ev2Hands-R. We evaluate single-hand \cite{boukhayma20193d,rudnev2021eventhands} and two-hand \cite{li2022interacting, Moon_2023_CVPR_InterWild} methods.
Ev2Hands significantly outperforms both the single-hand event and single-hand RGB approaches, presumably because they do not handle heavy occlusions. 
The proposed approach also outperforms the RGB-based two-hand methods \cite{li2022interacting, Moon_2023_CVPR_InterWild} {in most cases}, {especially} when considering the more challenging RR-AUC metric, while operating at significantly higher temporal resolution. 
For RGB-based methods \cite{li2022interacting, boukhayma20193d}, we provide \textit{ground-truth} cropped hands as the input, while \cite{Moon_2023_CVPR_InterWild} and our method do not require hand crops. For the event-based method \cite{rudnev2021eventhands}, we use our approach to generate event labels for each hand. 
The latter are then given as input to \cite{rudnev2021eventhands} to reconstruct the position and pose of the hands. 
As Boukhayma \textit{et al.}~\cite{boukhayma20193d} use a scaled orthographic projection for each hand, the RR-PCK and RR-AUC metrics cannot be evaluated. 
Note in Fig.~\ref{fig:qualitative} how EventHands~\cite{rudnev2021eventhands} fails to reconstruct hand interactions and how the RGB-based methods fail with low-light images. 
We also compare our approach to RGB-based methods on images captured under different camera frame rates in Table~\ref{tab:highfpsexp} and Fig.~\ref{fig:qualitative_rgb_events} {emulating low lighting arising from the high shutter speed and motion blur due to the fast motion of the hands.} 
{Under these conditions, RGB-based methods fail drastically, while our method outputs reasonable predictions with much higher temporal resolution.} 
\begin{table}[htbp] 
\centering
\resizebox{\columnwidth}{!}{
\begin{tabular}{@{}lcccccc@{}}
\hline
 & \multicolumn{3}{c}{High Shutter Speed ($500$ FPS)} & \multicolumn{3}{c}{Fast Motion ($25$ FPS)} \\ \hline
\textbf{}                & R-AUC  & RR-AUC & Coll\% & R-AUC & RR-AUC & Coll\% \\
Li \etal \cite{li2022interacting}                & \multicolumn{1}{r}{0.24}           & 0.12                       & \multicolumn{1}{r}{19.99}            & \multicolumn{1}{r}{0.38}           & 0.25                       & \multicolumn{1}{r}{12.67}          \\
Moon \etal \cite{Moon_2023_CVPR_InterWild} & \multicolumn{1}{r}{0.41}  & 0.23                       & \multicolumn{1}{r}{6.71}            & \multicolumn{1}{r}{0.4}           & 0.23                       & \multicolumn{1}{r}{6.82}          \\
\textbf{Ev2Hands (Ours)} & \multicolumn{1}{r}{\textbf{0.47}} & \textbf{0.30}             & \multicolumn{1}{r}{\textbf{0.57}} & \multicolumn{1}{r}{\textbf{0.53}} & \textbf{0.36}             & \multicolumn{1}{r}{\textbf{4.09}} \\ \hline
\end{tabular}
}
\caption{{Ev2Hands outperforms Li \etal \cite{li2022interacting} and Moon \etal \cite{Moon_2023_CVPR_InterWild} on under-exposed high-speed ($500$~FPS) videos and blurry fast motions ($25$~FPS) by a fair margin while estimating 3D hand poses with a much higher temporal resolution ($1000$~FPS).}} 
\label{tab:highfpsexp} 
\end{table}
\subsection{Ablation Study} \label{seq:ablation}

\begin{table}[htbp] 
\centering
\resizebox{\columnwidth}{!}{
\begin{tabular}{@{}lccccc@{}}
\toprule
   & R-AUC $\uparrow$ & RR-AUC $\uparrow$  & Coll\% $\downarrow$ \\
 \cmidrule(lr){2-4}
 \text{LNES \cite{rudnev2021eventhands}}                 &  0.56       &    0.46     & \thrd{8.13} \\
 \text{Raw events}                                       &  0.62       &    0.46     & \scnd{7.98} \\
 \text{Event Cloud (EC) adapted from~\cite{chen2022EPP}} &  0.66       &    0.53     & 8.39 \\ \hline
 \text{EC$+$Attention}                                   & \thrd{0.69} & \thrd{0.57} & 8.38 \\
 \text{EC$+$Attention$+\mathcal{L}_{\text{seg}}$}        & \best{0.75} & \best{0.66} & 8.39 \\
 \text{EC$+$Attention$+\mathcal{L}_{\text{seg}}+$IAL}    & \scnd{0.72} & \scnd{0.63} & \best{6.69} \\
\bottomrule
\end{tabular}
}
\caption{Ablation study on Ev2Hands-S with different event representations (top) and different components of our method (bottom). ``IAL'' refers to the Intersection Aware Loss $\mathcal{L}_{isec}$. 
} 
\label{tab:ablation_representation} 
\end{table} 

\begin{table}[htbp] 
\centering
\resizebox{\columnwidth}{!}{
\begin{tabular}{@{}lccccc@{}}
\toprule
   & R-AUC $\uparrow$ & RR-AUC $\uparrow$  & Coll\% $\downarrow$ \\
 \cmidrule(lr){2-4}
 \text{Event Cloud (EC)}                                   &  0.35       &    0.37     & 7.53 \\
 \text{EC$+$Attention}                                     & 0.38        &  0.40 & 7.62  \\
 \text{EC$+$Attention$+\mathcal{L}_{\text{seg}}$}          & \scnd{0.41} & \scnd{0.43} & \thrd{7.52} \\
 \text{EC$+$Attention$+\mathcal{L}_{\text{seg}}+$IAL}      & \thrd{0.39} & \thrd{0.41} & \scnd{7.49} \\
 \text{EC$+$Attention$+\mathcal{L}_{\text{seg}}+$IAL$+$FT} & \best{0.64} & \best{0.52} & \best{5.12} \\
\bottomrule
\end{tabular}
}
\caption{Ablation on Ev2Hands-R with different losses and training strategies. ``FT'' refers to \textit{fine-tuning} on real data.
}
\label{tab:ablation_training} 
\end{table} 
\noindent \textbf{Influence of Representation}.
We systematically examine the impact of different event representations in Tab.~\ref{tab:ablation_representation}-(top). 
Although LNES~\cite{rudnev2021eventhands} outperforms event point cloud representations in Coll\%, it fails to provide precise pose estimation of interacting hands (R-PCK metric). The EC representation provides the best pose estimation scores.
\begin{figure}[t]
\centering
   \includegraphics[width=0.89\linewidth]{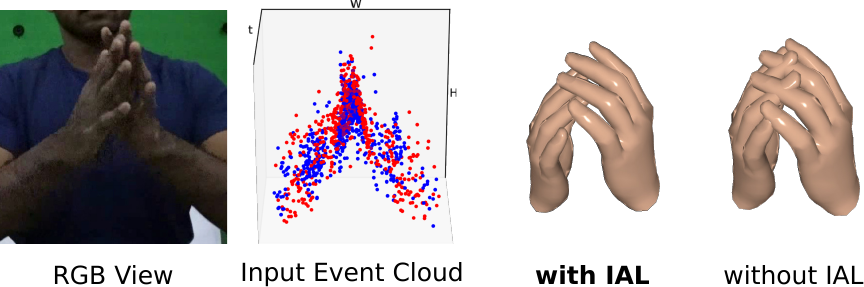}
   \caption{
    Influence of the Intersection Aware Loss (IAL).
} 
\label{fig:qualitative_intersection}
\end{figure}
\begin{figure}[t]
\centering
   \includegraphics[width=0.9\linewidth]{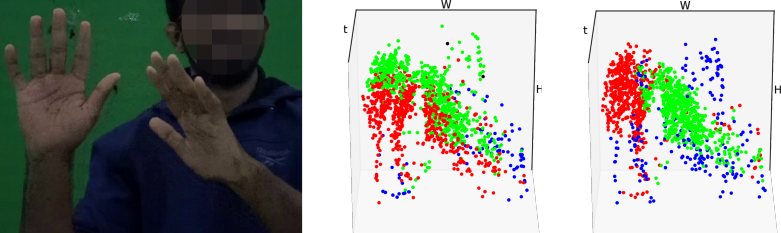}
   \caption{
    Predicted segmentation without (middle) and with fine-tuning on real data without segmentation labels (right). 
} 
\label{fig:qualitative_segmentation}
\end{figure}
\par
\noindent \textbf{Influence of Attention}. 
We conduct ablations on the proposed architecture and losses on both synthetic and real datasets (Tables~\ref{tab:ablation_representation} and \ref{tab:ablation_training}).
We see that the feature-wise attention mechanism, even without supervision, improved the method's performance (EC$+$Attention).
With additional segmentation supervision from the synthetic data, a substantial performance improvement can be observed. 
Interestingly, we observe that the attention mechanism learns plausible segmentation values even on real data when fine-tuned using only pose annotations; see Table~\ref{tab:ablation_training} and Fig.~\ref{fig:qualitative_segmentation}. 
\par
\noindent \textbf{Influence of the Intersection Loss}. 
By modelling hand intersections explicitly (experiments$+$IAL), the amount of interpenetration decreases as indicated by Coll\%. 
Although the pose estimation performance slightly drops, we theorise this is because small deviations in pose prediction can cause a lot of interpenetration in heavy interaction cases. 
However, since the physical plausibility of the interaction is essential for many applications, this trade-off could still be advantageous. 
A qualitative comparison is shown in  Fig.~\ref{fig:qualitative_intersection}. 
\section{Conclusion}\label{sec:discussion} 
We presented \textit{Ev2Hands}, the first method for two-hand 3D hand pose estimation from event streams. 
Our event cloud representation, when combined with the novel attention-based segmentation mechanism and collision mitigation loss, regresses reasonable 3D poses of two interacting hands and outperforms the related methods on our proposed benchmark dataset, \textit{Ev2Hands-R}, on real event streams. 
This is enabled by our new synthetic dataset, \textit{Ev2Hands-S}, which provides 3D pose, segmentation labels, and corresponding RGB images, all of which are difficult to obtain for real event streams. 
Furthermore, Ev2Hands works well in low illumination conditions and can estimate high-speed 3D hand motions. 
Our Ev2Hands assumes a fixed camera. While this does not pose an issue in traditional RGB methods, a moving (portable) event camera would generate a large amount of background clutter. 
Future work could investigate how to extract events caused by the object of interest. 
Another exciting avenue for future research would be to combine the RGB and event streams, thereby increasing the visual fidelity of the data while preserving the low latency of the event stream.
This will make high-quality textured 3D reconstructions of fast-moving hands possible. 
As this is also the first work on the 3D reconstruction of more than one non-rigid object from a single event stream, we believe it could inspire future research. 

\noindent\textbf{Acknowledgement.} 
This work was supported by the ERC consolidator grant 4DReply (770784). 

\clearpage

{
    \small
    \bibliographystyle{ieeenat_fullname}
    \bibliography{references}
}

\clearpage 
\setcounter{figure}{0} 
\setcounter{table}{0} 
\renewcommand*\thetable{\Roman{table}}
\renewcommand*\thefigure{\Roman{figure}}
\maketitlesupplementary
\appendix

This supplement provides additional details about our datasets in Section~\ref{sec:suppmap_dataset} and our experimental settings in Section~\ref{sec:suppmap_experiments}. 
Additional results are presented in Section~\ref{sec:suppmap_results}.
We use Roman numerals to reference figures and tables. 
\section{Dataset Details} \label{sec:suppmap_dataset}

\subsection{Ev2Hands-S Dataset}

\noindent \textbf{Adaptation from InterHand2.6M.} We obtain realistic hand motion by leveraging the InterHand2.6M dataset \cite{Moon_2020_ECCV_InterHand2.6M, moon2022neuralannot}  providing ground-truth MANO parameters for two hands. We use the provided MANO annotations, which are originally available at $5$ frames per second (FPS), and interpolate each sequence to achieve $30$ FPS. The interpolated hand models are rendered and composed with background scenes to generate the synthetic RGB frames. The RGB sequences are fed into the event stream simulator VID2E \cite{Gehrig_2020_CVPR}, which outputs the corresponding event streams. 
We follow train and test splits as Moon \etal~\cite{Moon_2020_ECCV_InterHand2.6M}.

\noindent \textbf{Material and Lighting.}
To synthesise realistic images that will be fed to the event stream simulator, we need to model the material properties and the lighting conditions in our simulations. 
To provide clear transitions in the boundaries of the hands as well as realistic colour changes in the interior parts of the hands, we apply a metallic-roughness material model \cite{pharr2016physically, pyrender} to the hand models.

The scene is illuminated with ambient light along with five-point light sources with positions and intensities randomly perturbed for each sequence.
We also rendered each sequence with nine different backgrounds.

\noindent \textbf{Hand Surface Modeling.}
In addition, we add a Gaussian noise with the std.~dev.~of $3$mm to the MANO vertices (hand surfaces) before rendering, resulting in more realistic event streams and helping the model to generalise to real data. 

\noindent \textbf{Data Augmentation.} We augment the event stream with noise by perturbing the existing events with random position and time offsets, and polarity swaps.
The augmented events emulate noise (\textit{i.e.,} spurious events) and event patterns due to possible changes in the illumination.

\subsection{Ev2Hands-R Dataset}

\noindent \textbf{Dataset composition.} The dataset comprises eight recorded sequences with  five subjects. 
The subjects encompass a spectrum of physical attributes, including variations in both body size and skin colour. The cumulative duration of the recorded sequences amounts to $19.4$~minutes. We use two subjects for fine-tuning our method and three subjects for evaluation purposes.

\begin{figure}[t]
\centering
   \includegraphics[width=0.9\linewidth]{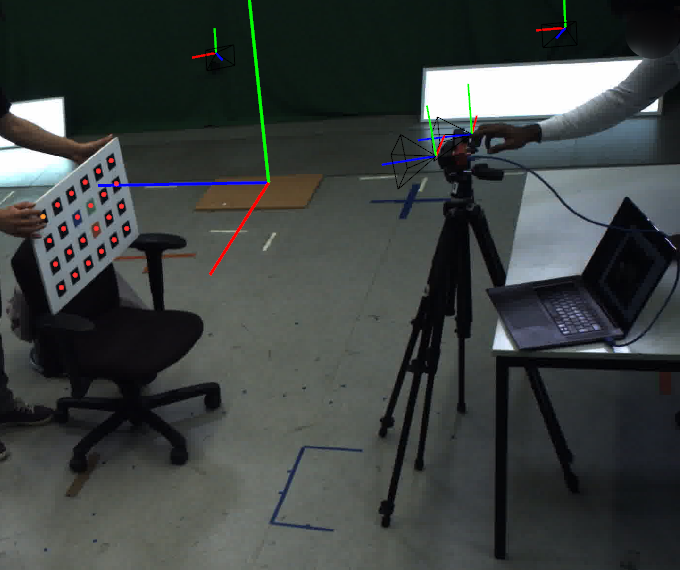}
   \caption{
    Recording setup for our Ev2Hands-R dataset with an event camera (\mbox{DAVIS 346C}) and an RGB camera (Sony RX0), which are synchronised and calibrated with respect to the 
    multi-view markerless motion capture system~Captury \cite{captury}. 
}
\label{fig:calibration}
\end{figure}

\noindent \textbf{Event+RGB Camera Setup.}
Our setup for real data capture includes an event camera (\mbox{DAVIS 346C}) and a high-speed RGB camera (\mbox{Sony RX0}); see Fig.~\ref{fig:calibration}.
All the cameras are synchronised and calibrated, and the RGB video stream is used only to compare our method with the existing RGB-based approaches and for reference. 

Synchronisation is achieved through a sequence of claps both at the start and the end of the recording.
We transform the stream of events into event frames with a time window of 20~ms matching the frame time of a RGB frame.
Subsequently, we manually align the timestamp of the event frame
with the corresponding timestamp of the RGB frame containing the clap sequence.

\noindent \textbf{Reference 3D Hand Pose.}
The 3D reference hand poses for \textit{Ev2Hands-R} are obtained by a three-step process.
First, we calibrate the camera setup to obtain the intrinsic and extrinsic parameters of the event and  RGB cameras. The extrinsic parameters are obtained w.r.t.~the markerless motion capture system~\cite{captury} using a chequerboard. 
Second, we ask the target actor to perform hand sequences in front of our Event+RGB camera setup. The motion capture system produces full-body 3D human poses of the actor in global space, including the body and both hands.
Finally, the 3D hand markers are projected onto the camera views of the event and RGB cameras.
Table \ref{fig:datasetvisualization} shows the list of actions performed by one of the subjects. The way each individual carries out actions is affected by their personal style, resulting in small differences in each sequence. These slight variations contribute to the dataset diversity. 

\begin{figure}[t]
\centering
   \includegraphics[width=0.8\linewidth]{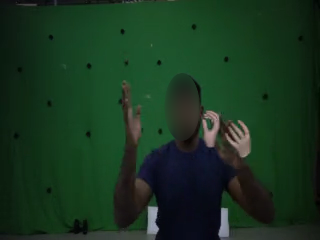}
   \caption{
    The predictions of RGB2Hands \cite{wang2020rgb2hands}.
    Due to the differences between the training regime and assumptions of RGB2Hands and our setup, the obtained predictions are poor.
}
\label{fig:rgb2handprediction}
\end{figure}

\section{Experimental Settings} \label{sec:suppmap_experiments}

\noindent \textbf{Comparisons to EventHands.}  We retrain and evaluate \mbox{EventHands} ~\cite{rudnev2021eventhands} on \textit{Ev2Hands-R} by individually inferring on the left- and right-hand events. The left- and right-hand event labels are obtained by the predicted segmentation labels using Ev2Hands, since the ground-truth event labels are not available for real event streams.

\begin{figure*}[t]
    \centering
    \begin{minipage}{.5\textwidth}
      \centering
      \includegraphics[width=0.9\linewidth]{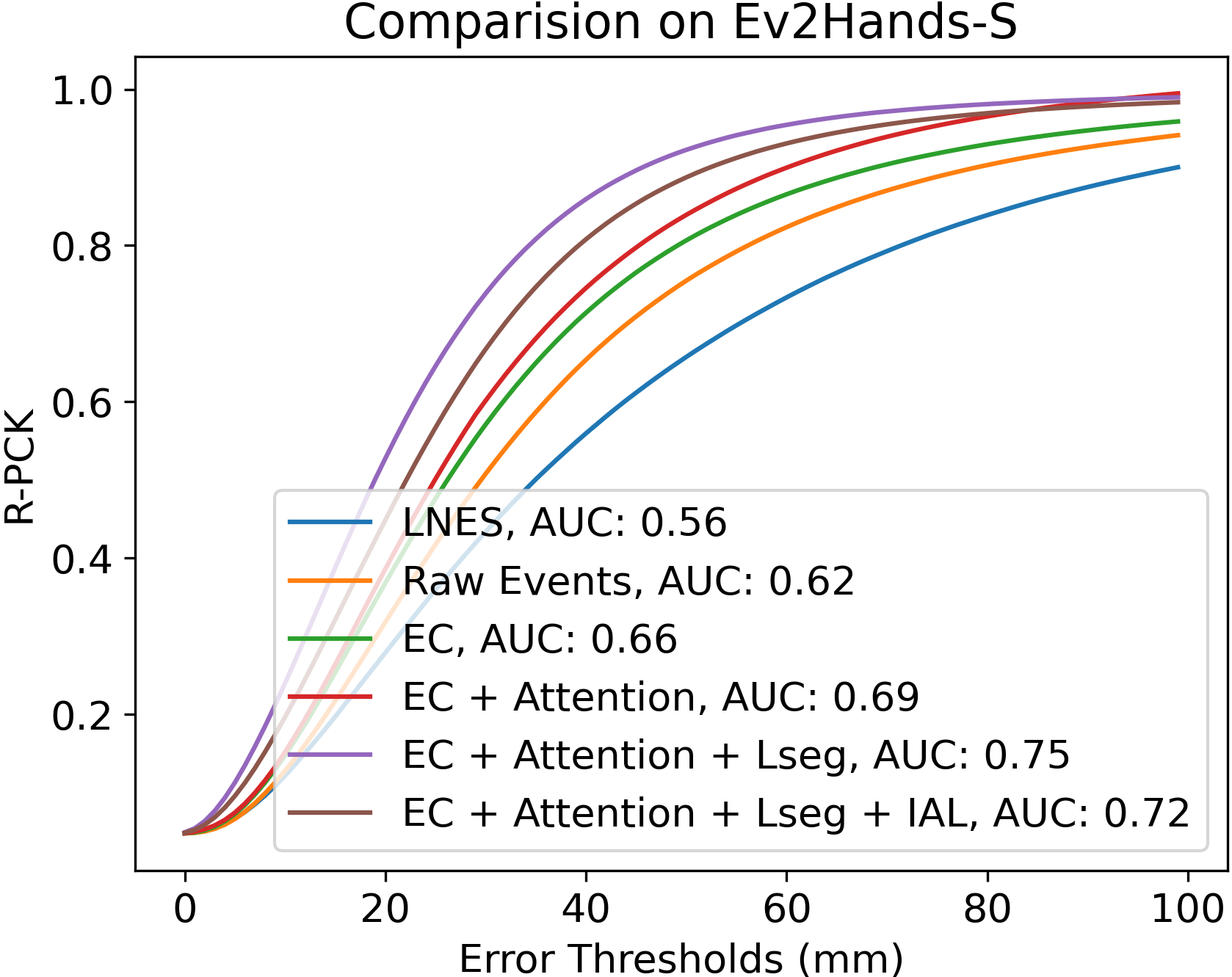}
    \end{minipage}%
    \begin{minipage}{.5\textwidth}
      \centering
      \includegraphics[width=0.9\linewidth]{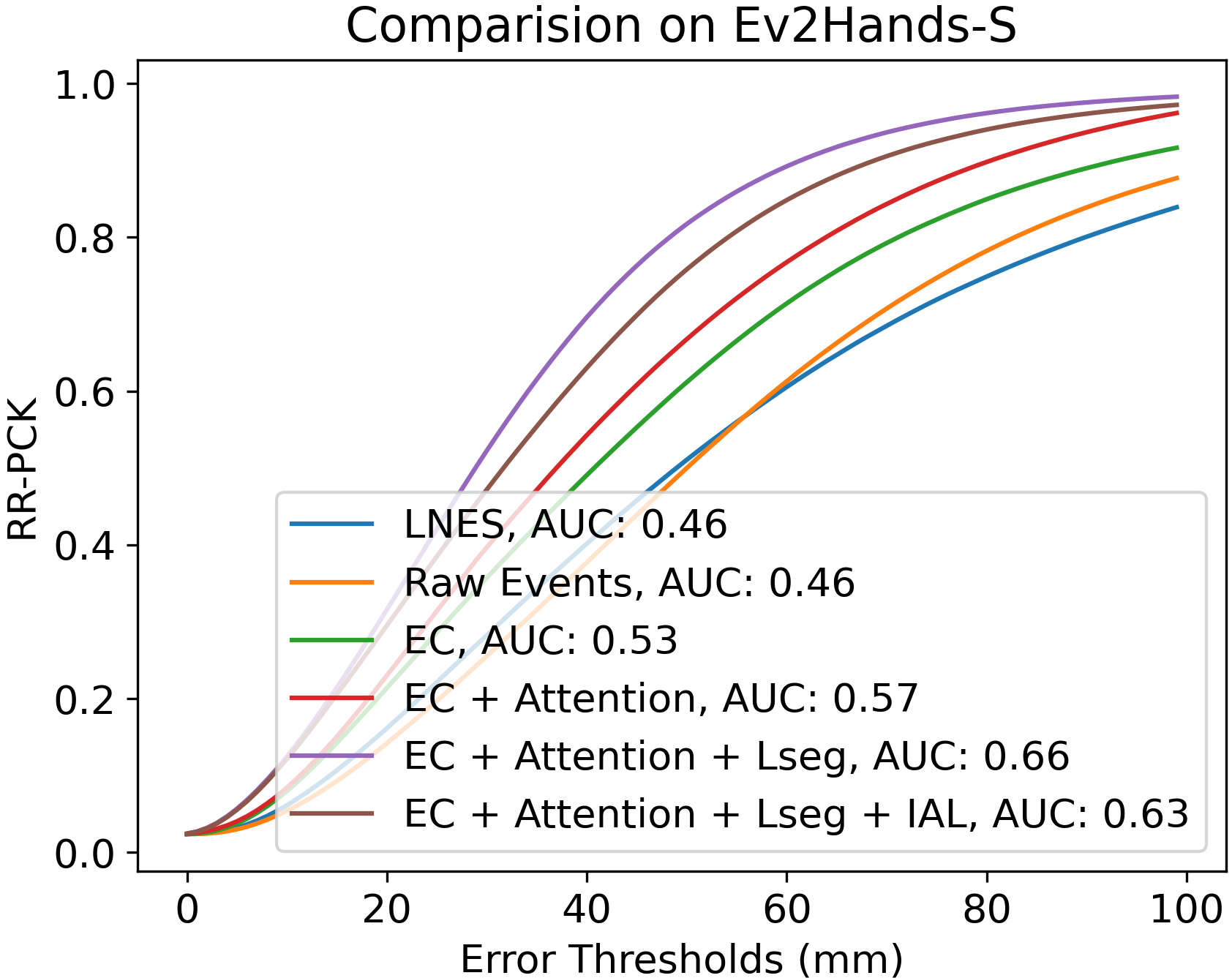}
    \end{minipage}
    \hspace{-2mm}
    \vskip\baselineskip
    \begin{minipage}{.5\textwidth}
      \centering
      \includegraphics[width=0.9\linewidth]{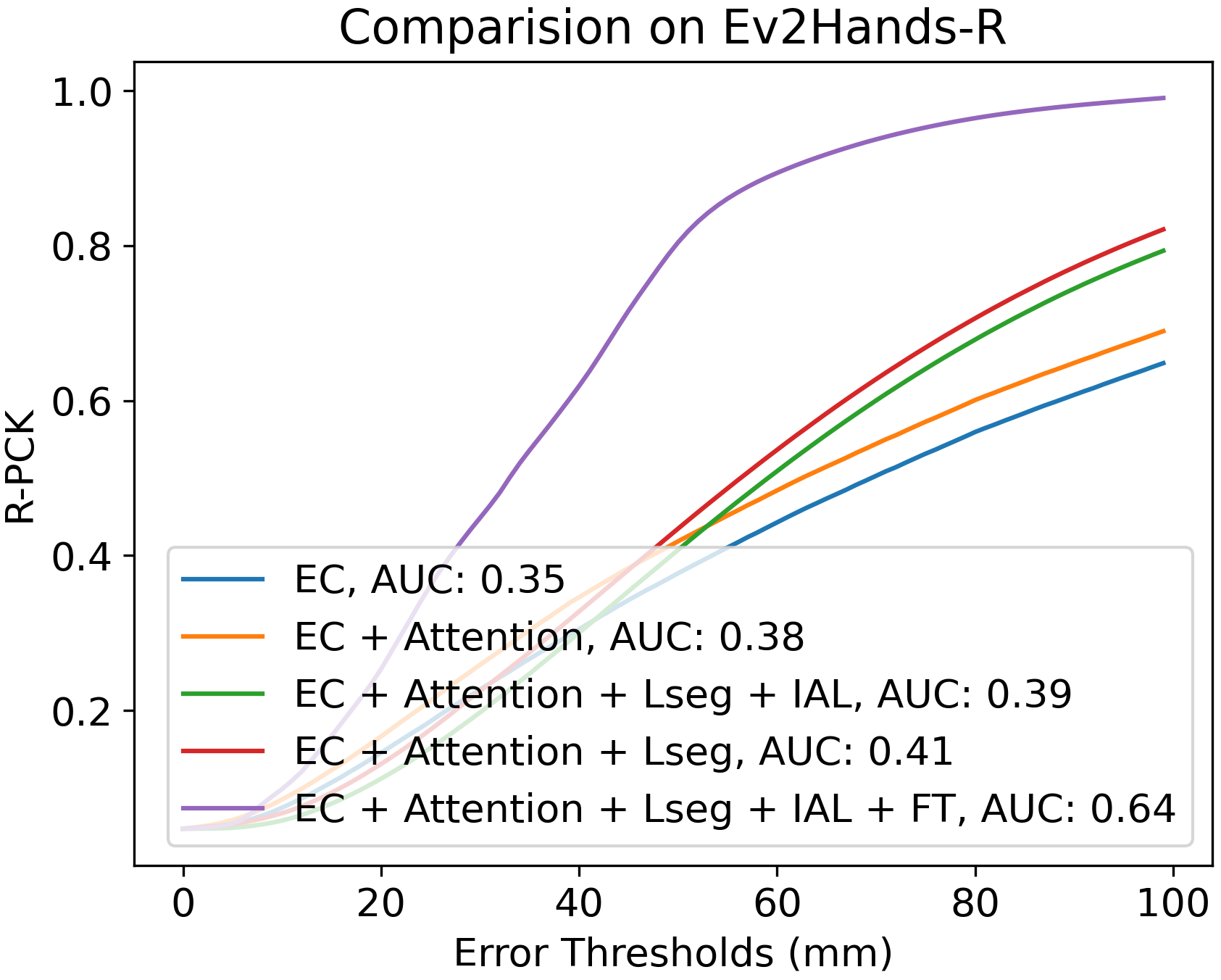}
    \end{minipage}%
    \begin{minipage}{.5\textwidth}
      \centering
      \includegraphics[width=0.9\linewidth]{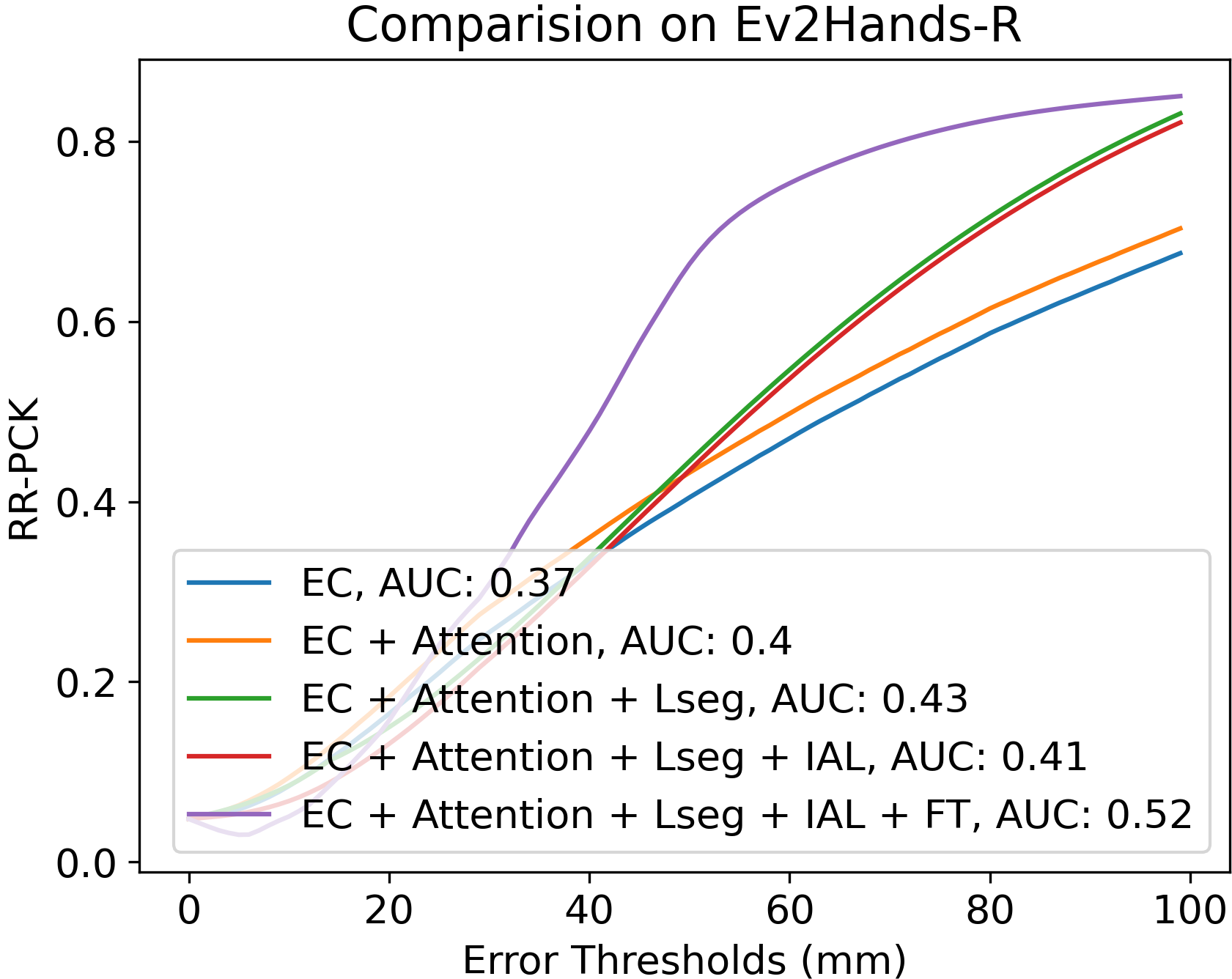}
    \end{minipage}
    \caption{
    PCK curves for the ablation experiments on the synthetic \textit{Ev2Hands-S} and real \textit{Ev2Hands-R} datasets. The curve is plotted for relative PCK (R-PCK) and relative root PCK (RR-PCK).
    }
\label{fig:ablation}
\end{figure*}

\begin{figure*}[t]
    \centering
    \begin{minipage}{.5\textwidth}
      \centering
      \includegraphics[width=0.9\linewidth]{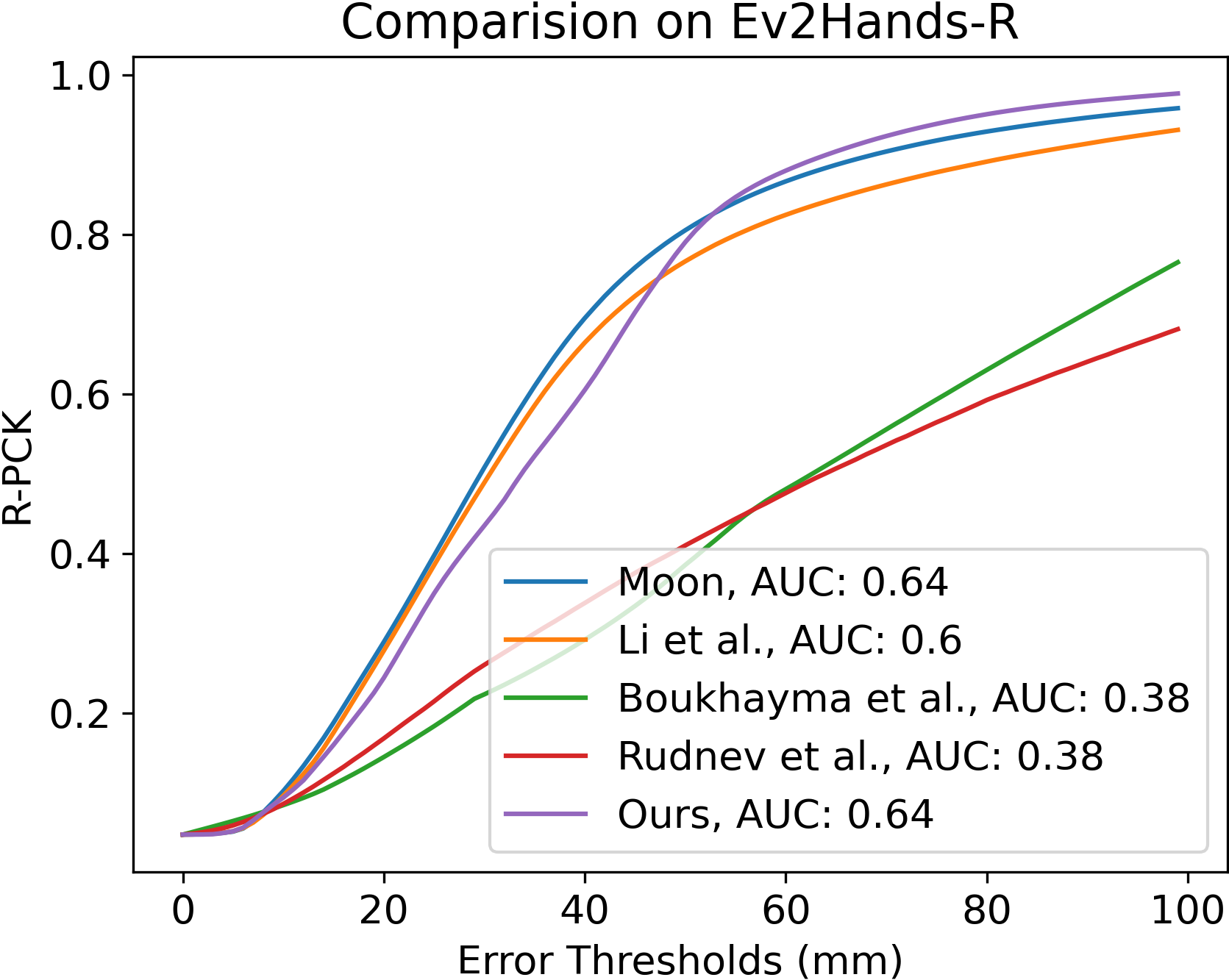}
    \end{minipage}%
    \begin{minipage}{.5\textwidth}
      \centering
      \includegraphics[width=0.9\linewidth]{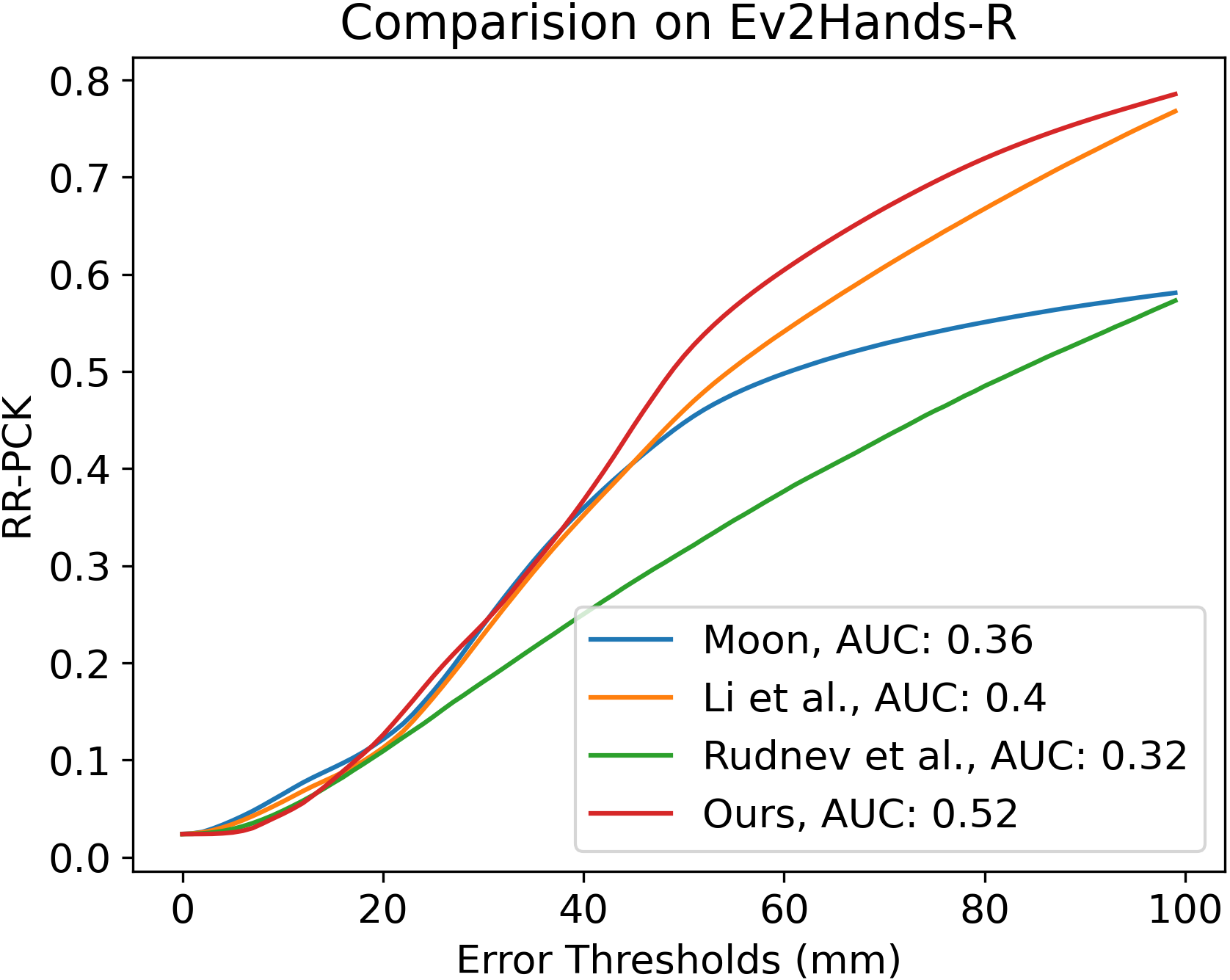}
    \end{minipage}
    \hspace{-2mm}
    \caption{
    PCK curves for Moon~\cite{Moon_2023_CVPR_InterWild}, Li~\etal~\cite{li2022interacting}, Boukhayma~\etal~\cite{boukhayma20193d}, Rudnev~\etal~\cite{rudnev2021eventhands} and Ev2Hands (Ours) on real \textit{Ev2Hands-R} dataset. The curve is plotted for relative PCK (R-PCK) and relative root PCK (RR-PCK).
    }
\label{fig:sotapck}
\end{figure*}

\noindent \textbf{Comparisons to Additional RGB-based Methods.} 
In addition to the methods evaluated in the main paper, \ie., Moon \cite{Moon_2023_CVPR_InterWild}, Li \etal~\cite{li2022interacting} and Boukhayma \etal~\cite{boukhayma20193d},
we also test other RGB-based methods for 3D hand pose estimation of two hands from monocular videos~\cite{wang2020rgb2hands, zhang2021interacting}. 
However, due to the differences in the training regime of these methods and our setup, the obtained predictions are poor. 
Therefore, we do not report quantitative metrics; see Fig.~\ref{fig:rgb2handprediction} for qualitative results for one of these RGB-based methods. 
\noindent \textbf{Ablative Study.} To investigate the contributions of the key components of our method, we conduct an ablation study on the \textit{Ev2Hands-R} and \textit{Ev2Hands-S} datasets. The PCK curves and AUC are shown in Fig.~\ref{fig:ablation}. The PCK curves and AUC of our method compared to Rudnev \etal~\cite{rudnev2021eventhands}, Li \etal~\cite{li2022interacting}, and Boukhayma \etal~\cite{boukhayma20193d} are shown in Fig.~\ref{fig:sotapck}.
When assessing performance using the R-PCK metric with \textit{Ev2Hands-S}, the use of raw events as input surpasses the performance of the LNES \cite{rudnev2021eventhands} approach as demonstrated by an AUC value of $0.62$. However, compared to the event clouds, raw events exhibit inferior performance. Incorporating the feature-wise attention mechanism leads to a higher AUC value of $0.69$. This is further improved by the segmentation supervision, resulting in the highest AUC score of $0.75$.
The introduction of Intersection Aware Loss (IAL) leads to a marginal reduction in the AUC score to $0.72$. Importantly, this is accompanied by a decrease in the rate of collisions down to a value of $6.69\%$, enhancing the plausibility of valid hand poses during the scenarios with highly interacting hands.
When assessing the performance of our method 
on \textit{Ev2Hands-R}, the same trend is observed. Additionally, Fine Tuning (FT) our method with \textit{Ev2Hands-R} increases the R-PCK AUC score from $0.41$ to $0.64$.    
The same trend is also reflected in RR-PCK values.

\noindent \textbf{Temporal Stability.} We reduce the jitter of high-speed motions generated by our method with the 1€ filter~\cite{casiez20121} for visualisations in the accompanying supplementary video. 
For a fair comparison, the same procedure is also performed for all the methods we evaluate. 
Note that the temporal filtering of the motions as shown in the video (5:23) produces a slight lag when observing the predictions at $1000$ FPS. 
\section{Additional Results}  \label{sec:suppmap_results}

\subsection{Comparison with RGB Camera Methods}
To demonstrate the robustness of our method to low-light and high-speed motion sequences, we compare it with RGB-based methods for 3D reconstruction of interacting hands \cite{li2022interacting, Moon_2023_CVPR_InterWild} on fast motion sequences captured by Sony RX0. 
We consider two scenarios, \textit{i.e.,} $25$ FPS and at a high shutter speed, which is $500$ FPS in our case. 
The two sequences are captured in different setups: 
Background activity from the subject, the distance between the subjects and the cameras are different compared to \textit{Ev2Hands-R}; calibration of the event camera, \textit{i.e.,} p-n bias\footnote{
The "p-n bias" refers to the sensitivity of firing positive and negative events. 
} settings and background noise filter settings are also different compared to \textit{Ev2Hands-R}. 
This shows our method can generalise to different recording setups. 
Unlike RGB-based hand pose estimation methods, which mostly fail on poorly lit frames (due to fast shutter speed or motion blur induced by fast hand motion), our method infers more accurate articulations and inter-hand distances; see Table~\ref{fig:highfpsexp_qualitative}. 

\subsection{Additional Visualisations}
We provide additional visualisations for our method in well-lit scenarios along with results for low-light and high-speed motion cases; see Table~\ref{fig:additionalresults}. The predicted segmentations, as shown in our experiments, help the network to disambiguate left and right hands even under very challenging conditions. Hence, our Ev2Hands approach is robust to hand occlusions in a wide range of scenarios, making it also suitable for high-speed two-hand interactions.
Please see our video (1:23) for further qualitative results.
\noindent \textbf{Segmentation Supervision.}
We next compare the performance of our method with pretrained and non-pretrained segmentation branches.
Both experiments are fine-tuned with real data.
In the supplementary video (6:03), we observe that segmentation supervision makes our method robust to ambiguities arising due to intense hand interactions. 

\newcolumntype{C}{>{\centering\arraybackslash}m{19.6em}}
\begin{table*}\sffamily
\centering
\begin{tabular}{l*8{C}@{}}
\toprule
Action & RGB & Events \\ \midrule

Palm Wave &
\includegraphics[width=12em]{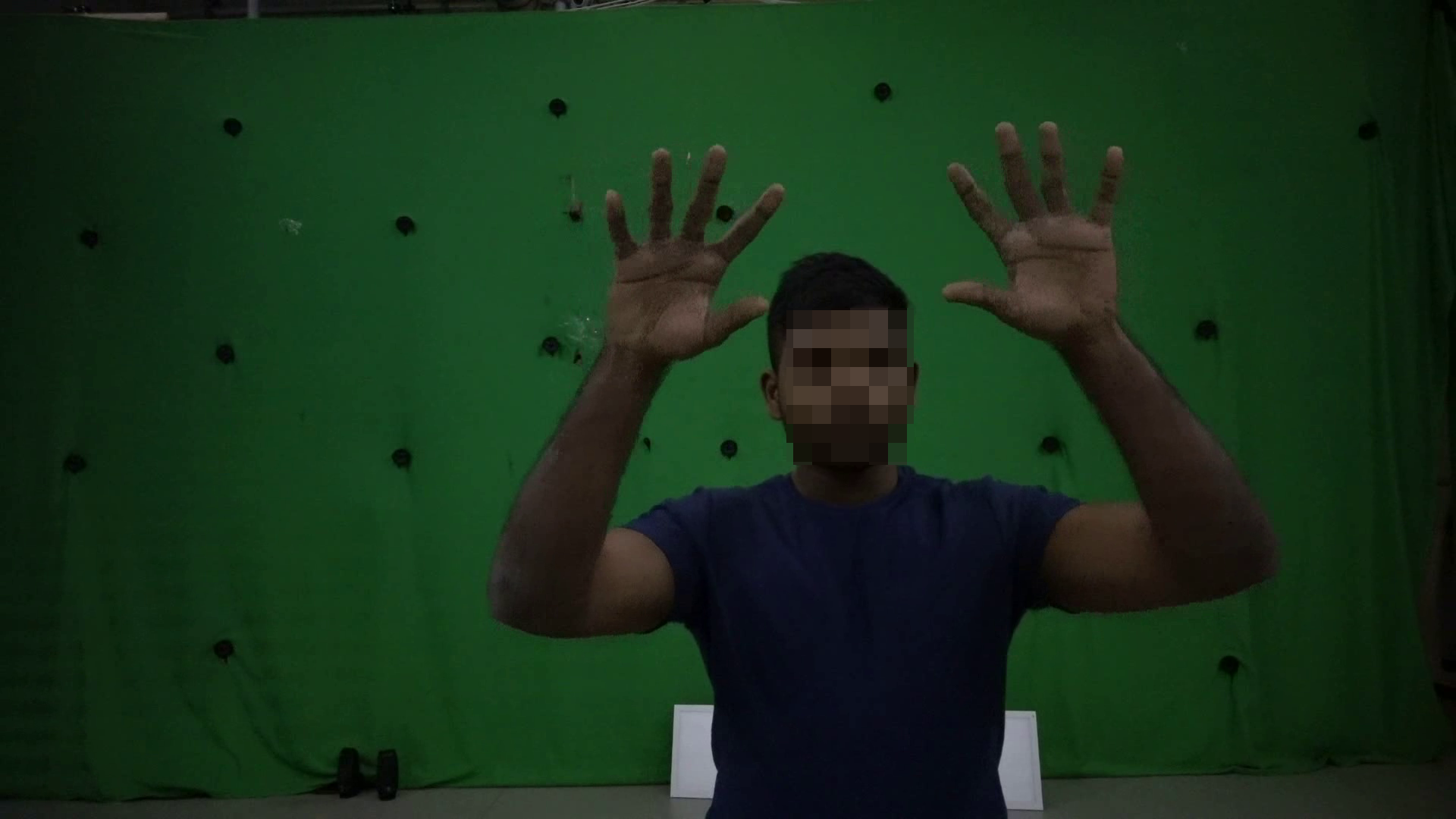}  &
\includegraphics[width=9em]{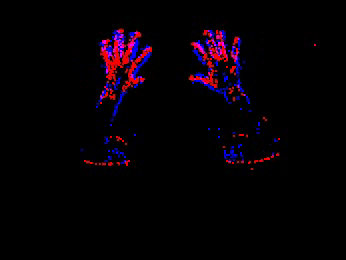} \\

Dorsal Wave &
\includegraphics[width=12em]{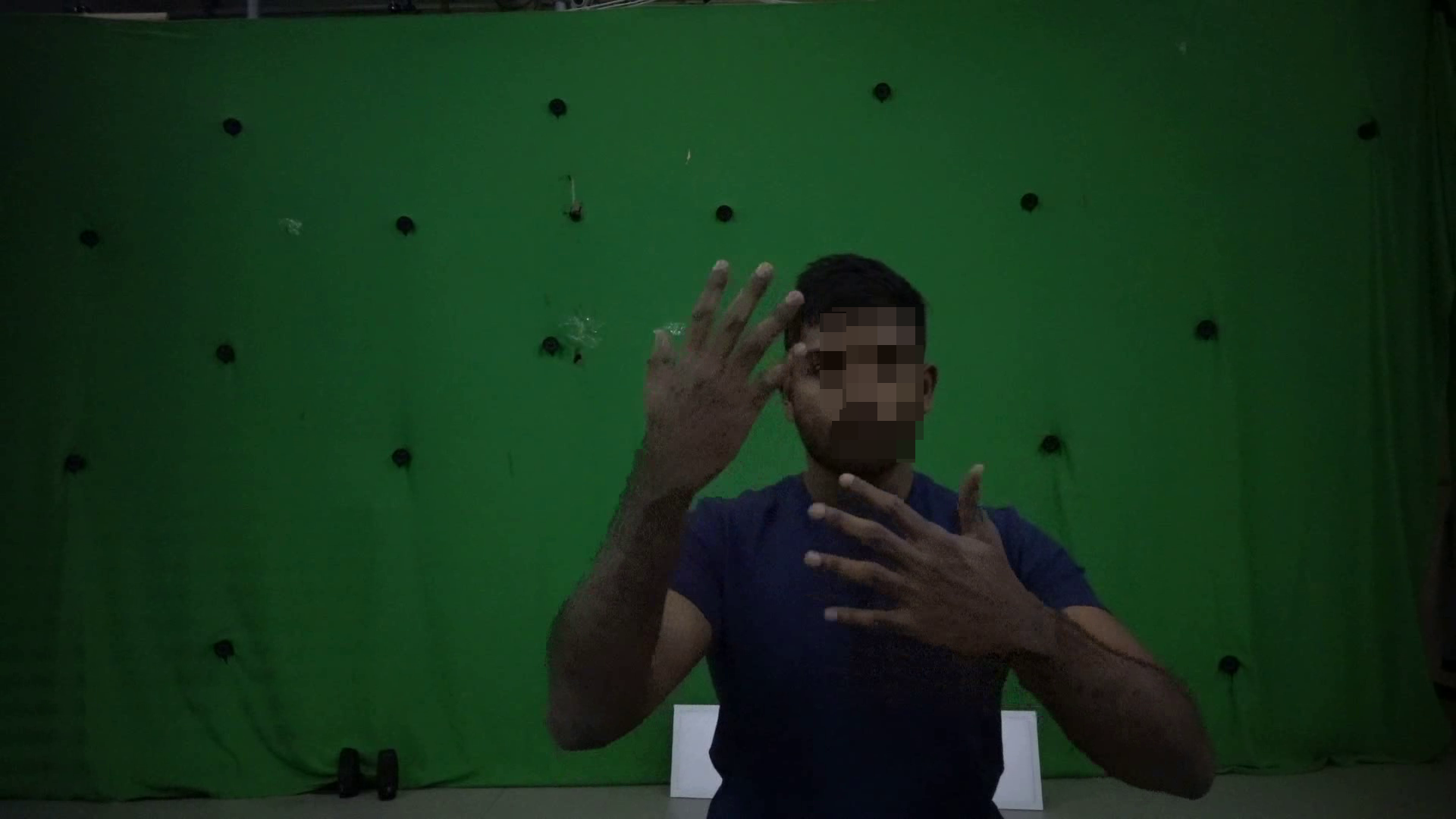}  &
\includegraphics[width=9em]{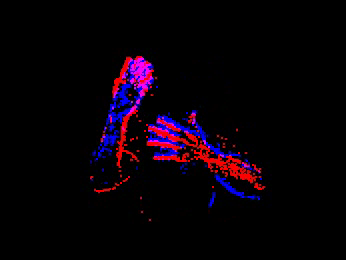} \\

Wrist Rotation &
\includegraphics[width=12em]{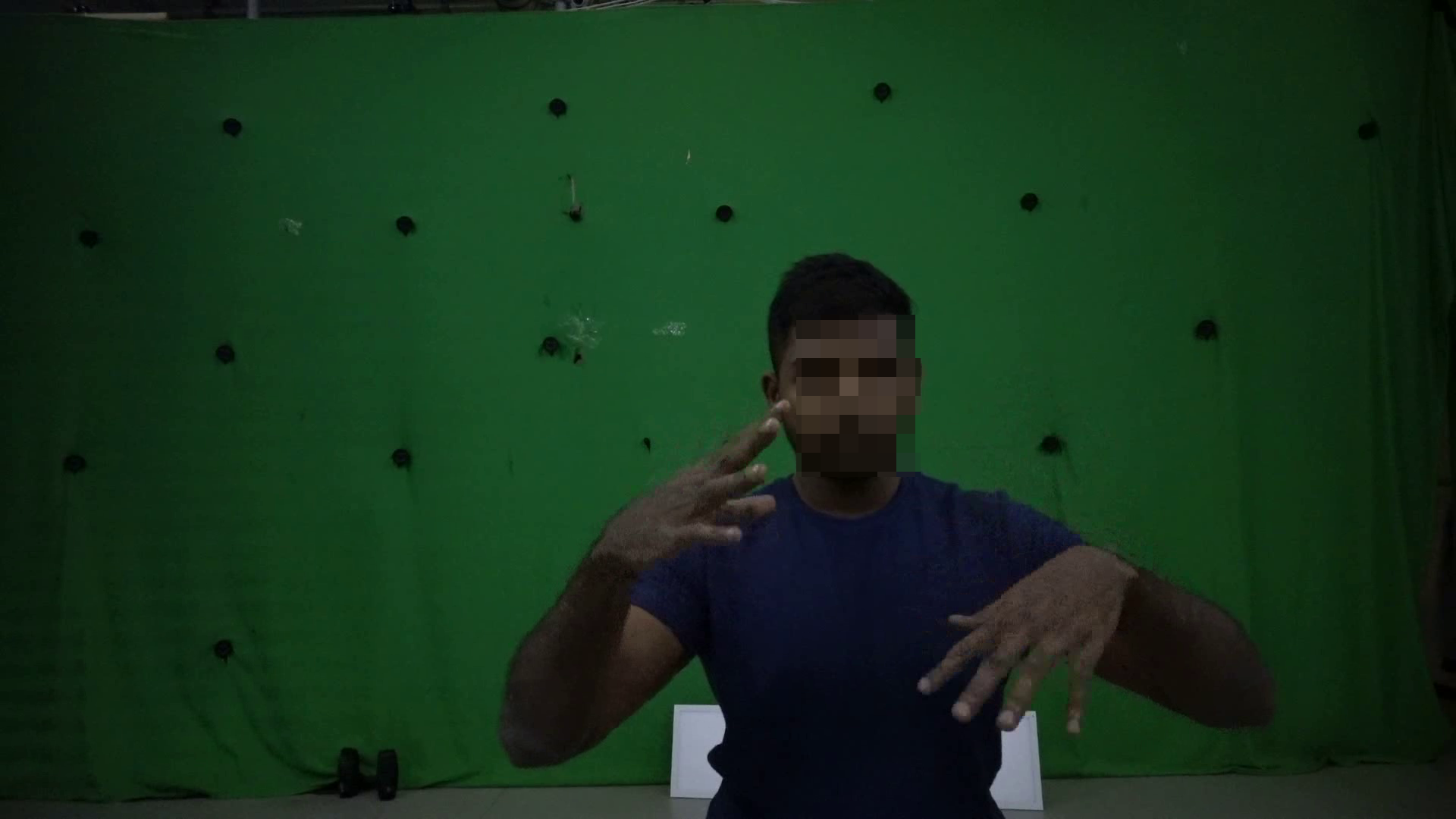}  &
\includegraphics[width=9em]{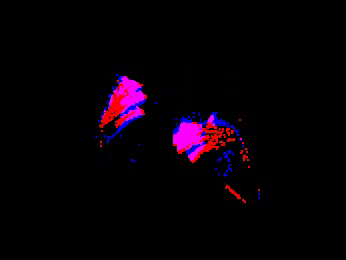} \\

Articulation &
\includegraphics[width=12em]{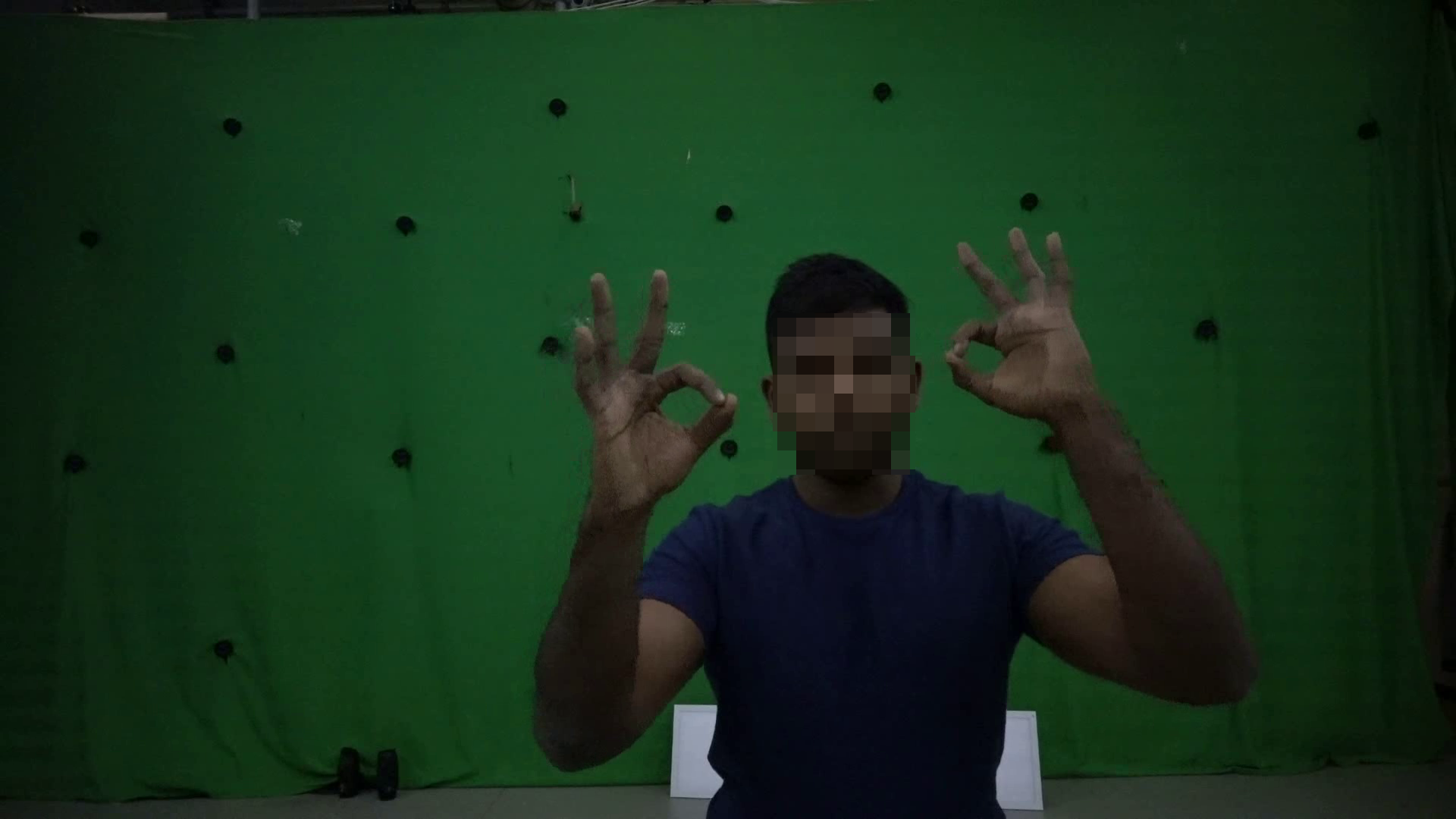}  &
\includegraphics[width=9em]{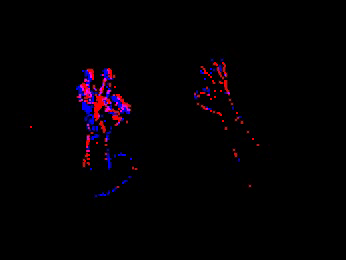} \\

Clap &
\includegraphics[width=12em]{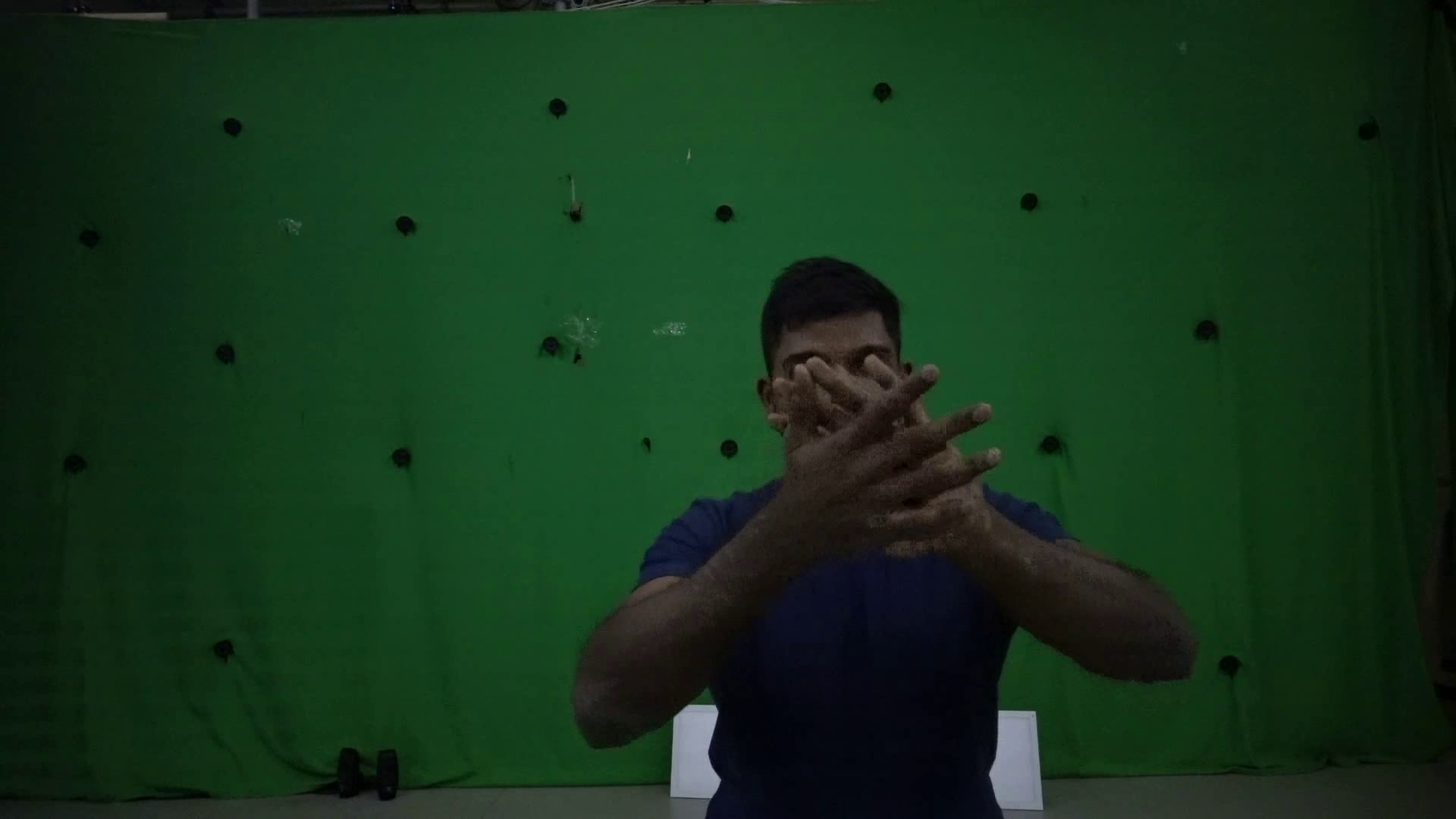}  &
\includegraphics[width=9em]{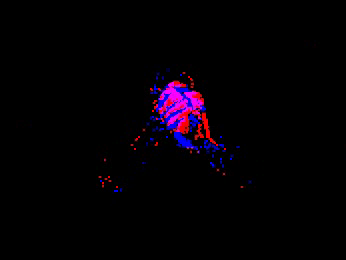} \\

Intersection &
\includegraphics[width=12em]{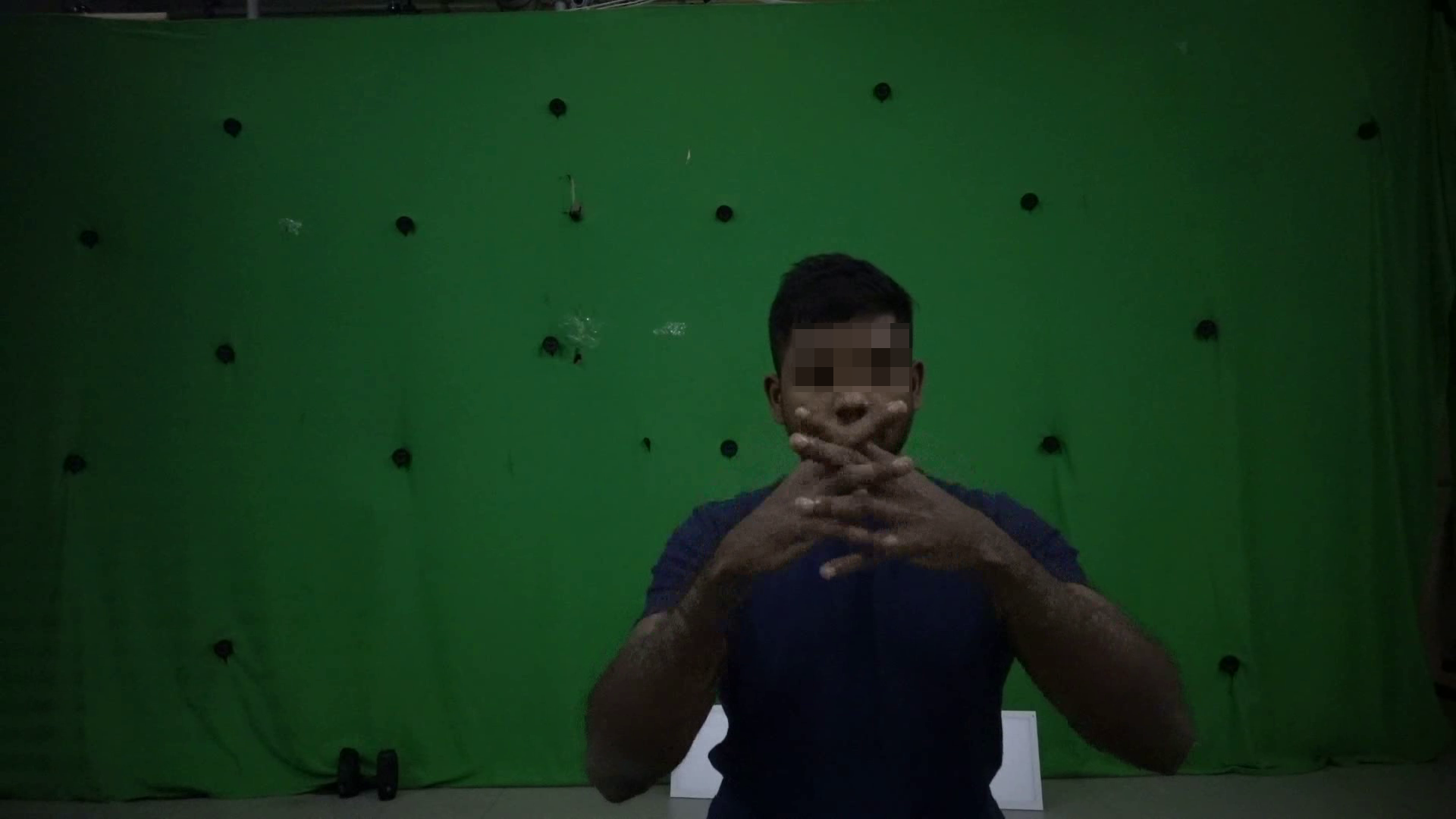}  &
\includegraphics[width=9em]{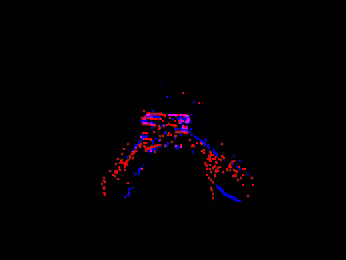} \\

Occlusion &
\includegraphics[width=12em]{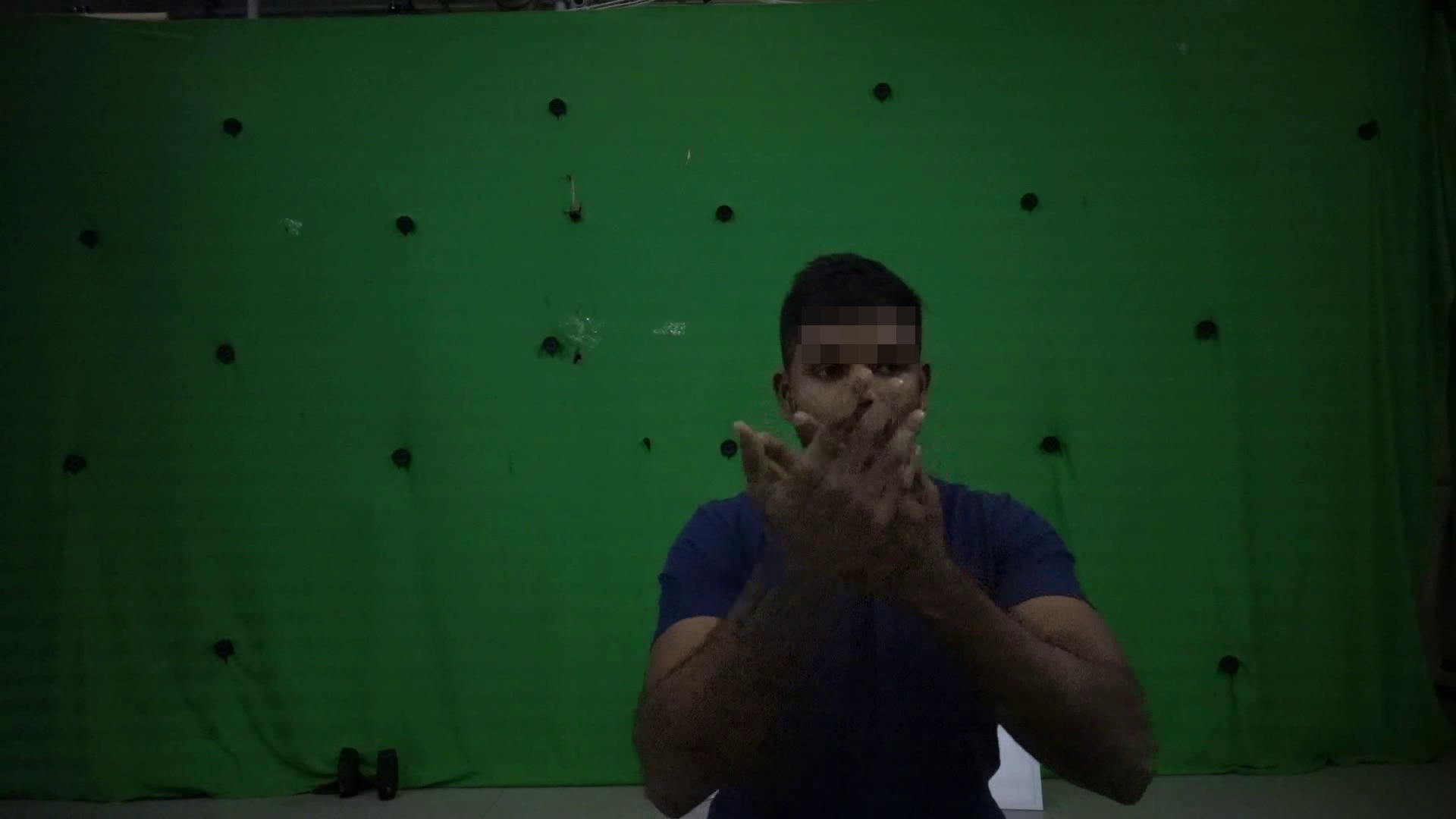}  &
\includegraphics[width=9em]{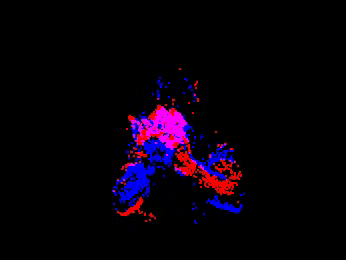} \\

Free Style &
\includegraphics[width=12em]{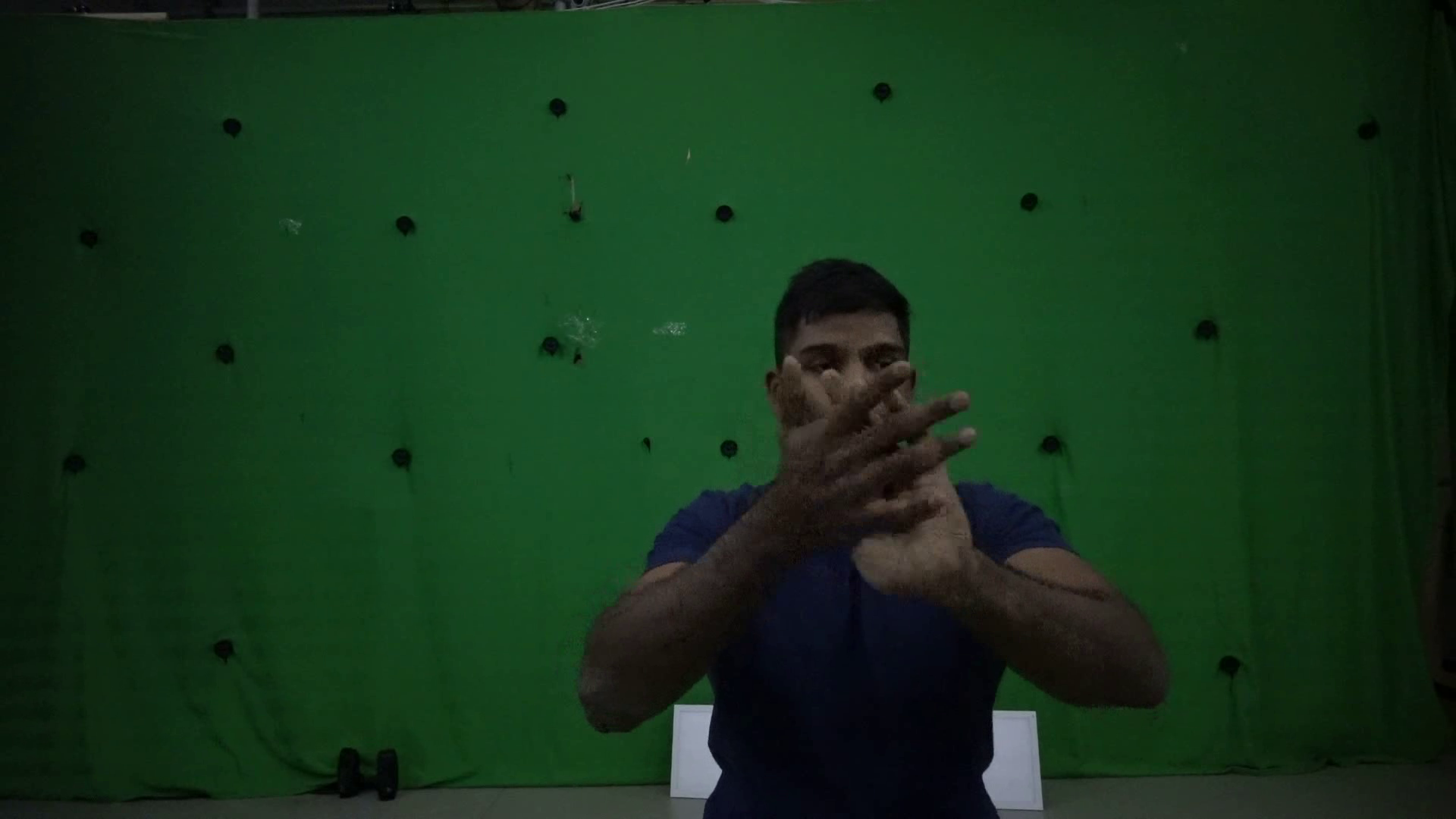}  &
\includegraphics[width=9em]{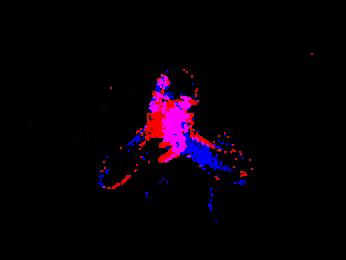} \\

\bottomrule
\end{tabular}
\caption{
The Ev2Hands-R dataset consists of eight different actions performed by each of the five different actors captured with synchronised RGB and event cameras.  
}
\label{fig:datasetvisualization}
\end{table*}

\newcolumntype{C}{>{\centering\arraybackslash}m{0.176\linewidth}}

\begin{table*}
\centering
\sffamily
\setlength\extrarowheight{-3cm}
\begin{tabular}{*{5}{C}}
\toprule
\multicolumn{1}{c}{RGB} & \multicolumn{1}{c}{Events} & \multicolumn{1}{c}{Li \etal} & \multicolumn{1}{c}{Moon} & \multicolumn{1}{c}{Ev2Hands (Ours)} \\
\midrule

\includegraphics[width=\linewidth]{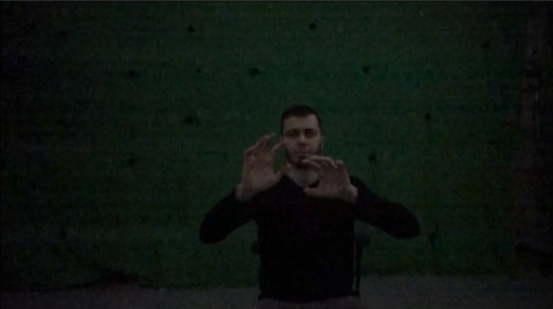} &
\includegraphics[width=\linewidth]{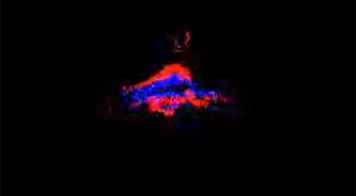} &
\includegraphics[width=\linewidth]{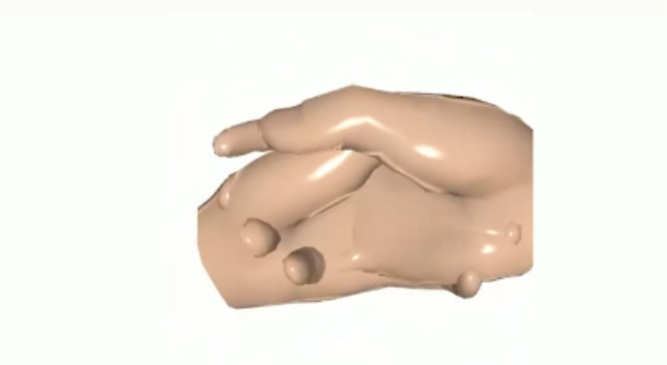} &
\includegraphics[width=0.8\linewidth]{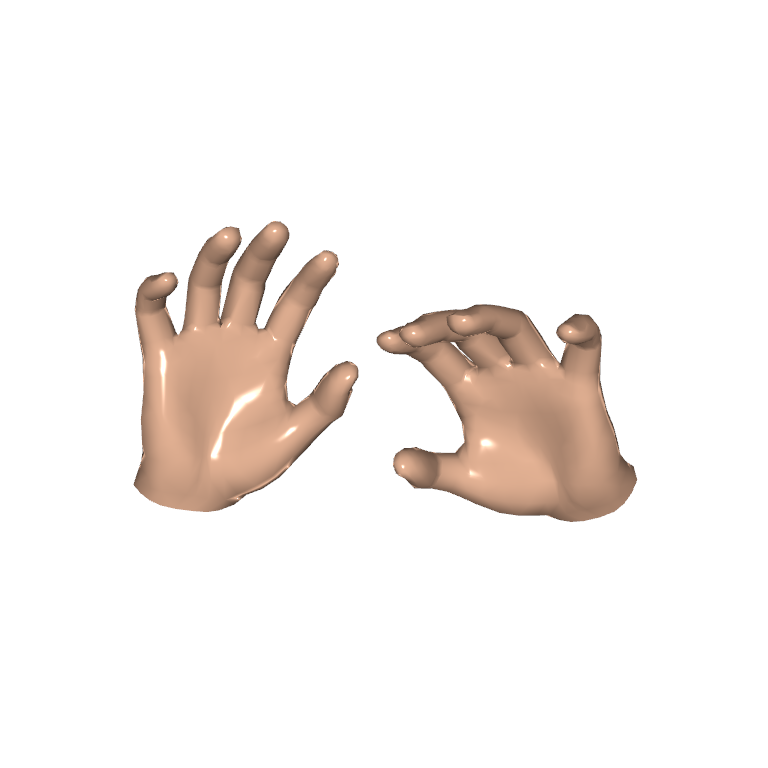} &
\includegraphics[width=\linewidth]{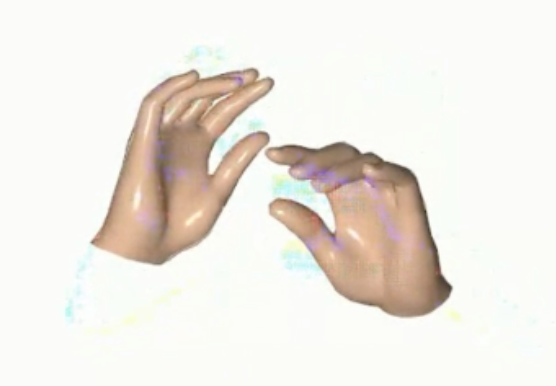} \\

\includegraphics[width=\linewidth]{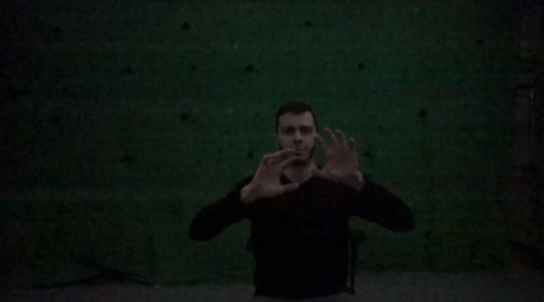}  &
\includegraphics[width=\linewidth]{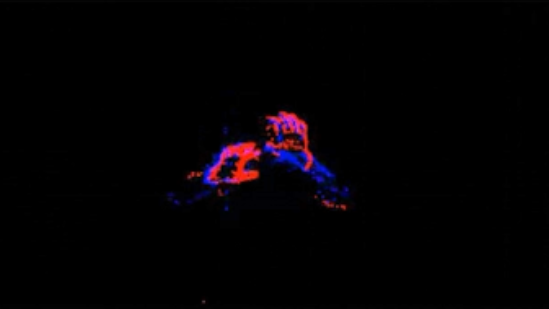} &
\includegraphics[width=\linewidth]{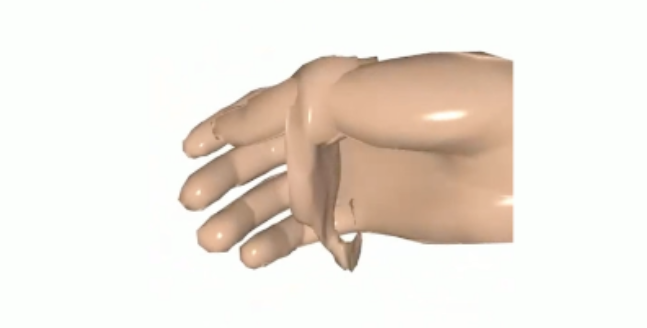} &
\includegraphics[width=0.8\linewidth]{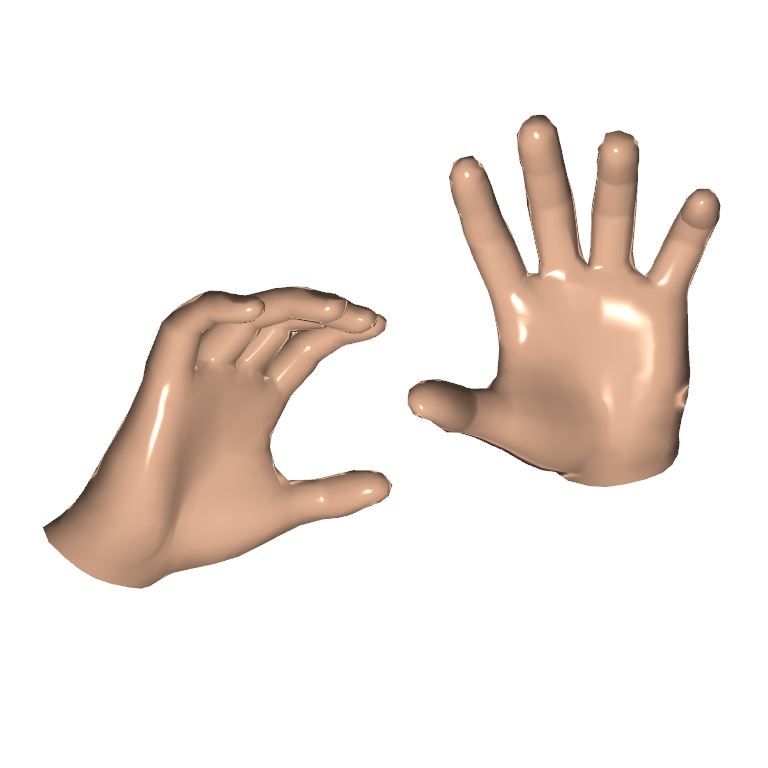} &
\includegraphics[width=\linewidth]{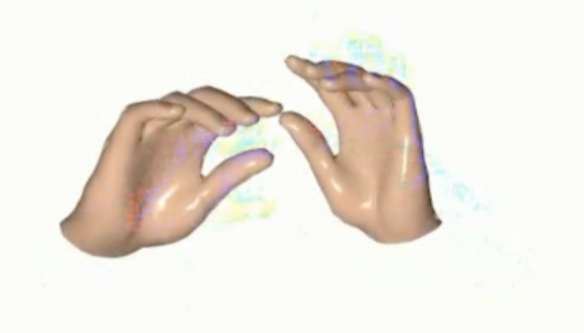} \\

\includegraphics[width=\linewidth]{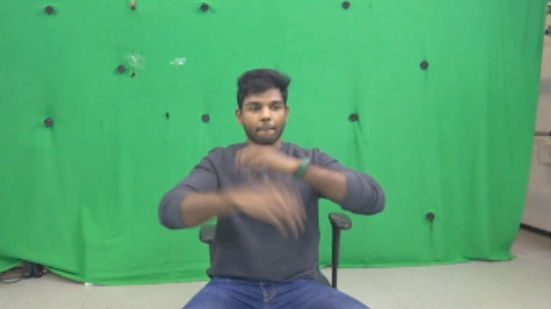}  &
\includegraphics[width=\linewidth]{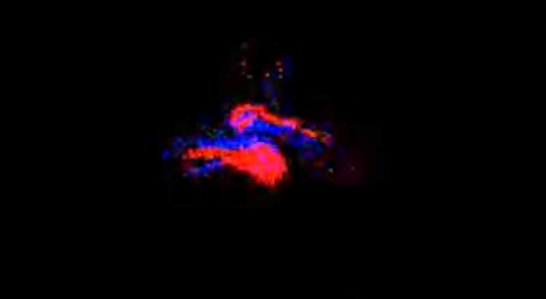} &
\includegraphics[width=\linewidth]{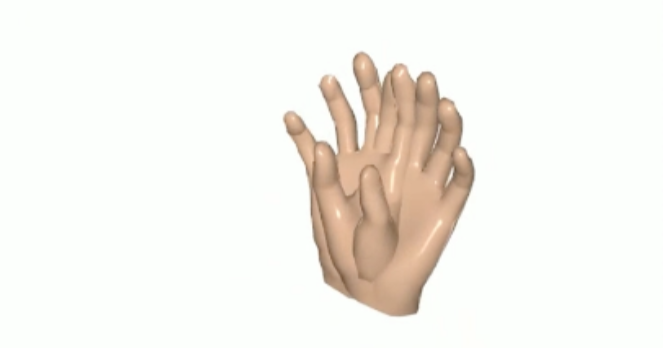} &
\includegraphics[width=0.8\linewidth]{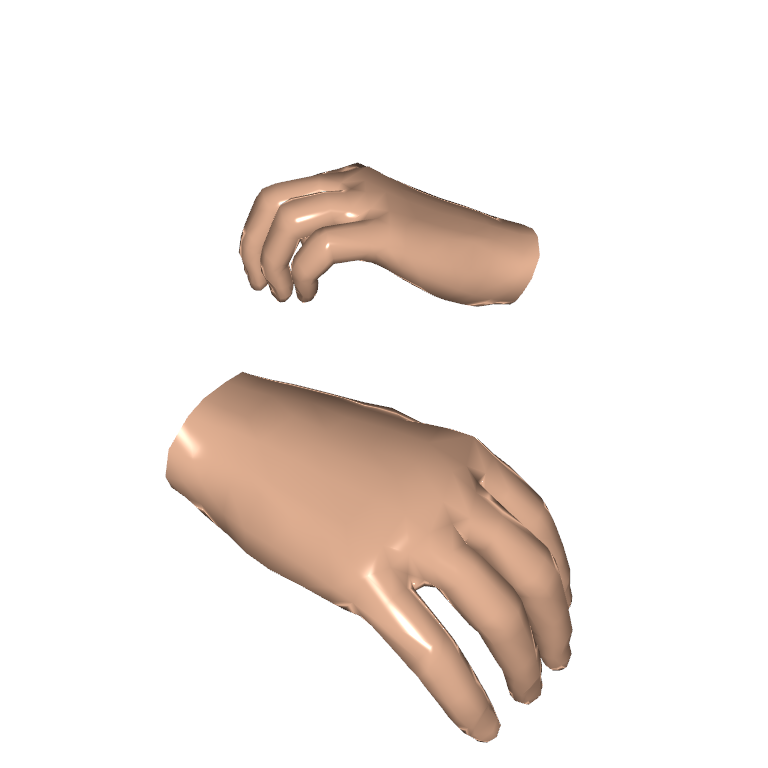} &
\includegraphics[width=\linewidth]{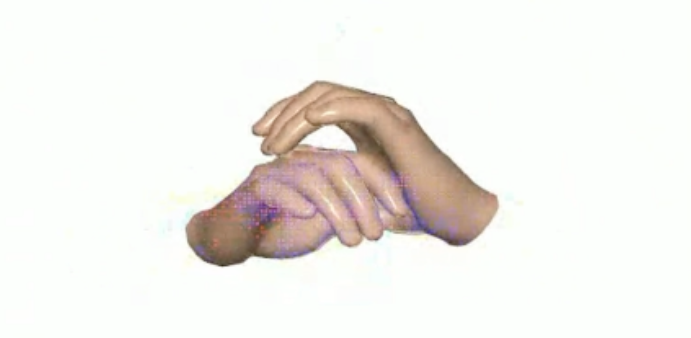} \\

\includegraphics[width=\linewidth]{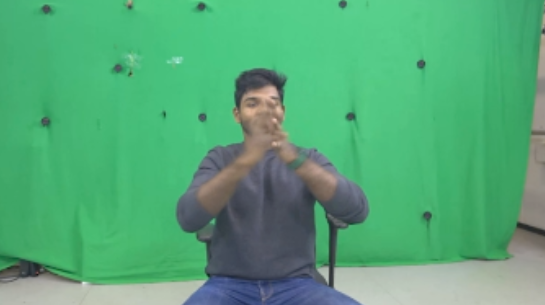}  &
\includegraphics[width=\linewidth]{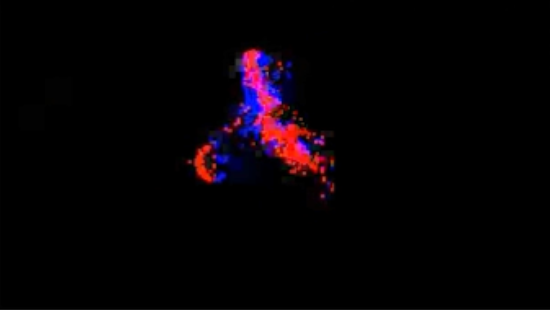} &
\includegraphics[width=\linewidth]{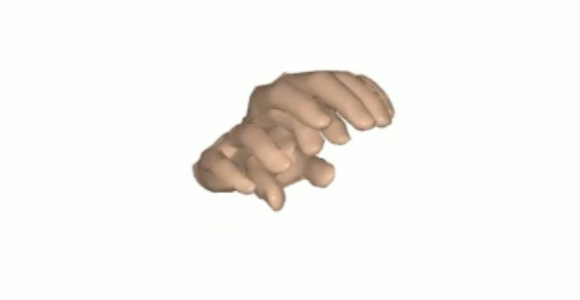} &
\includegraphics[width=0.8\linewidth]{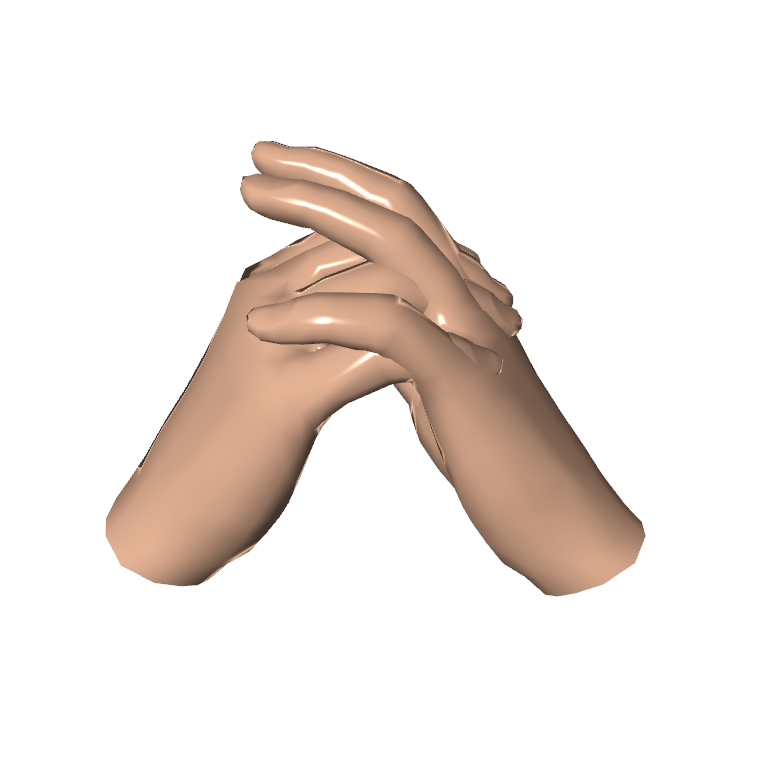} &
\includegraphics[width=\linewidth]{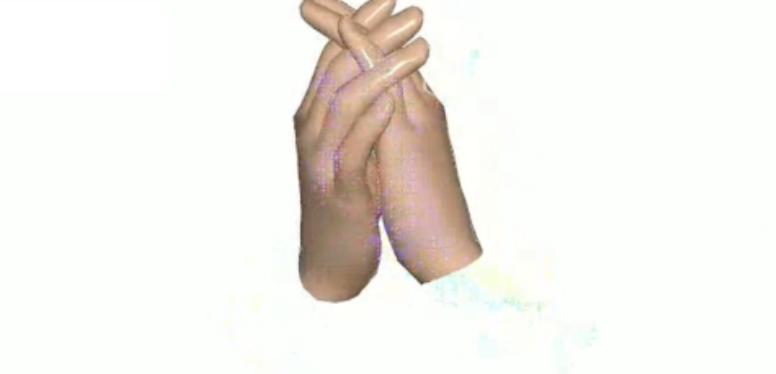} \\

\includegraphics[width=\linewidth]{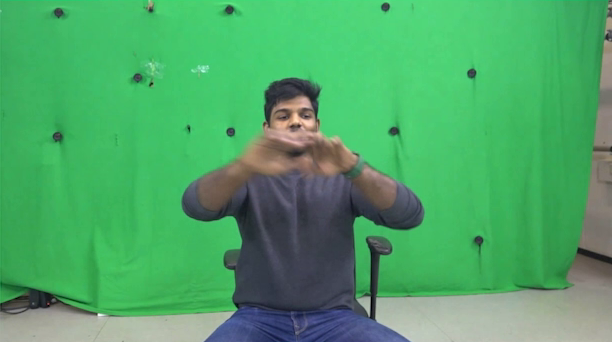}  &
\includegraphics[width=\linewidth]{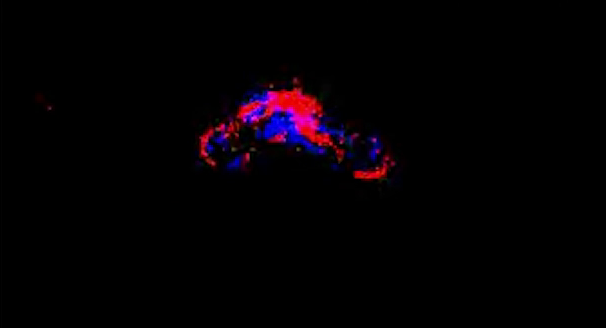} &
\includegraphics[width=\linewidth]{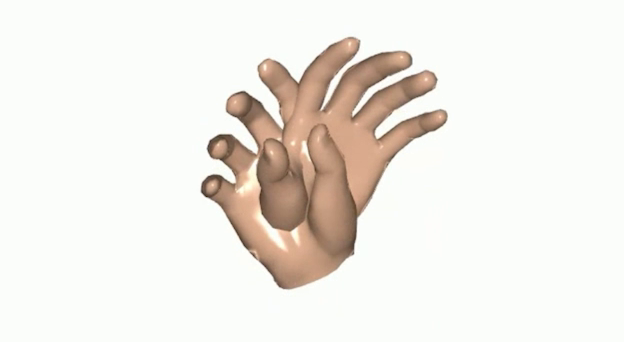} &
\includegraphics[width=0.8\linewidth]{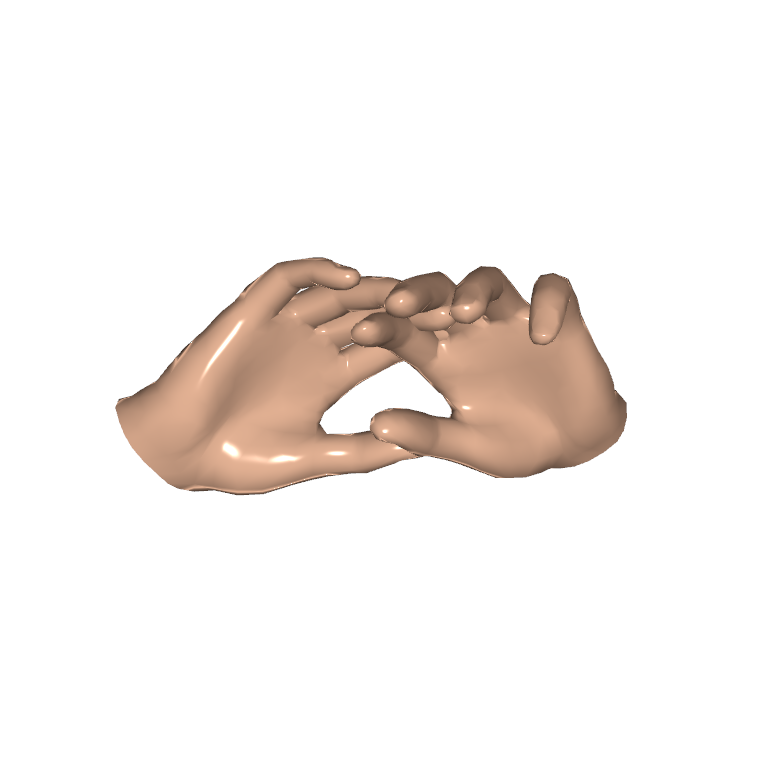} &
\includegraphics[width=\linewidth]{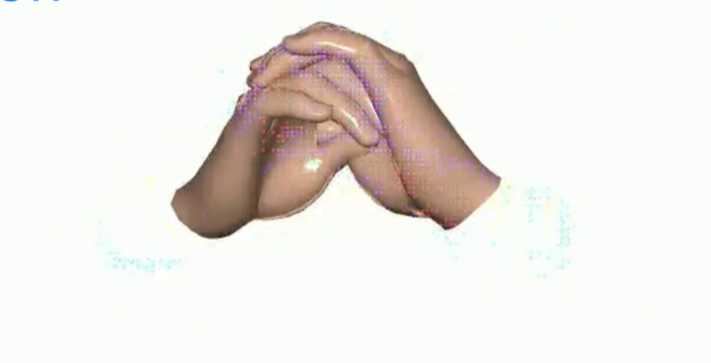} \\

\includegraphics[width=\linewidth]{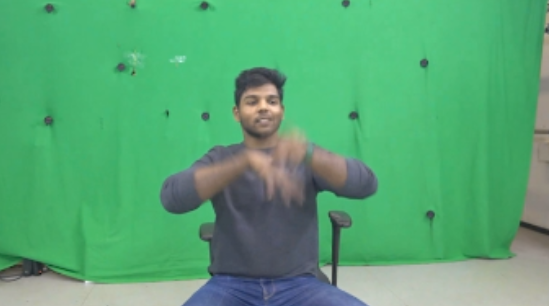}  &
\includegraphics[width=\linewidth]{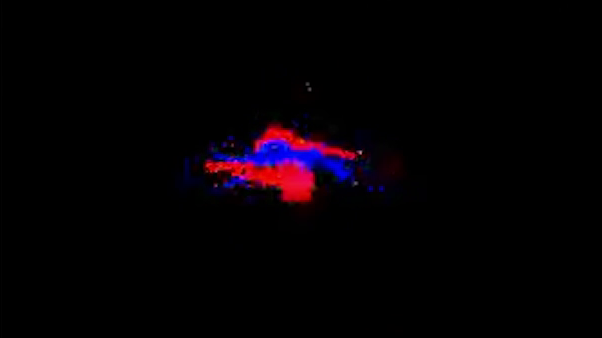} &
\includegraphics[width=\linewidth]{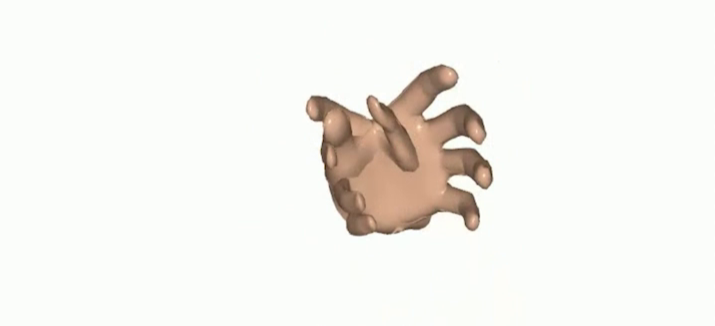} &
\includegraphics[width=0.8\linewidth]{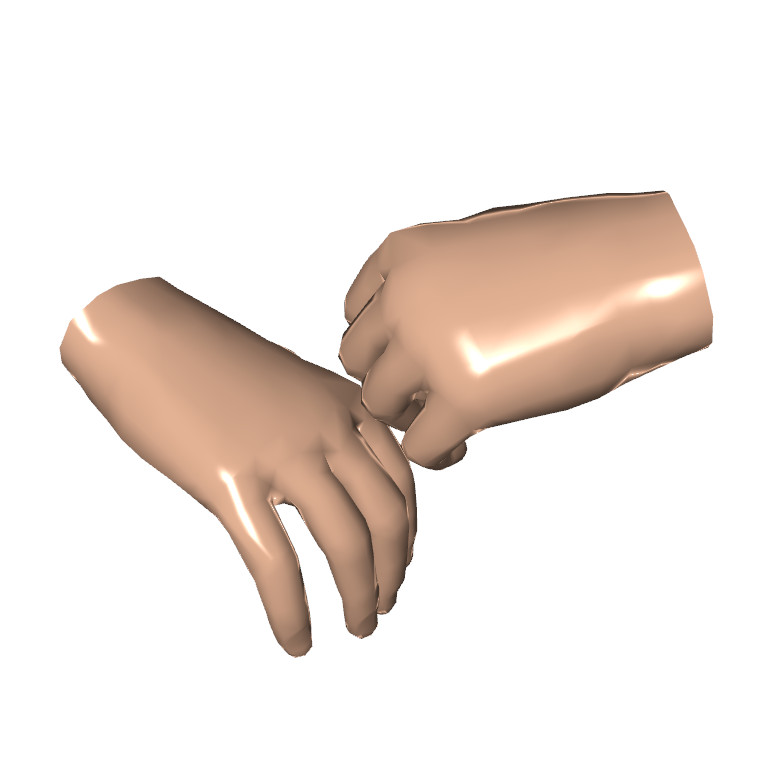} &
\includegraphics[width=\linewidth]{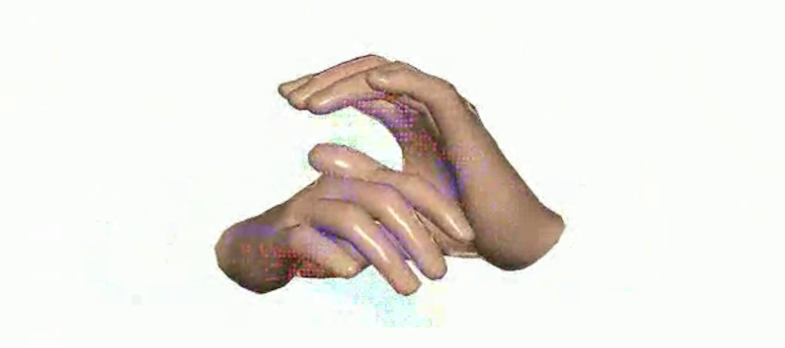} \\

\bottomrule
\end{tabular}
\caption{
Comparisons between our method and Li \textit{et al.}~and Moon \cite{li2022interacting, Moon_2023_CVPR_InterWild}. 
The first two rows show the results obtained from the high shutter speed ($500$ FPS) sequence while the last three rows show the results obtained from the fast motion sequence recorded at $25$ FPS. 
We observe that the predictions from Ev2Hands are better with more precise articulations and inter-hand distances. 
Note: The 3D visualizations of our method are overlaid onto the event stream.
}
\label{fig:highfpsexp_qualitative}
\vspace{4cm}
\end{table*}
\newcolumntype{C}{>{\centering\arraybackslash}m{8.7em}}
\begin{table*}\sffamily
\centering
\begin{tabular}{C*8{C}@{}}
\toprule
RGB & Events & Event Cloud & Segmentation & Predictions \\
\midrule

\includegraphics[width=8em, trim=0 -10cm 0 0]{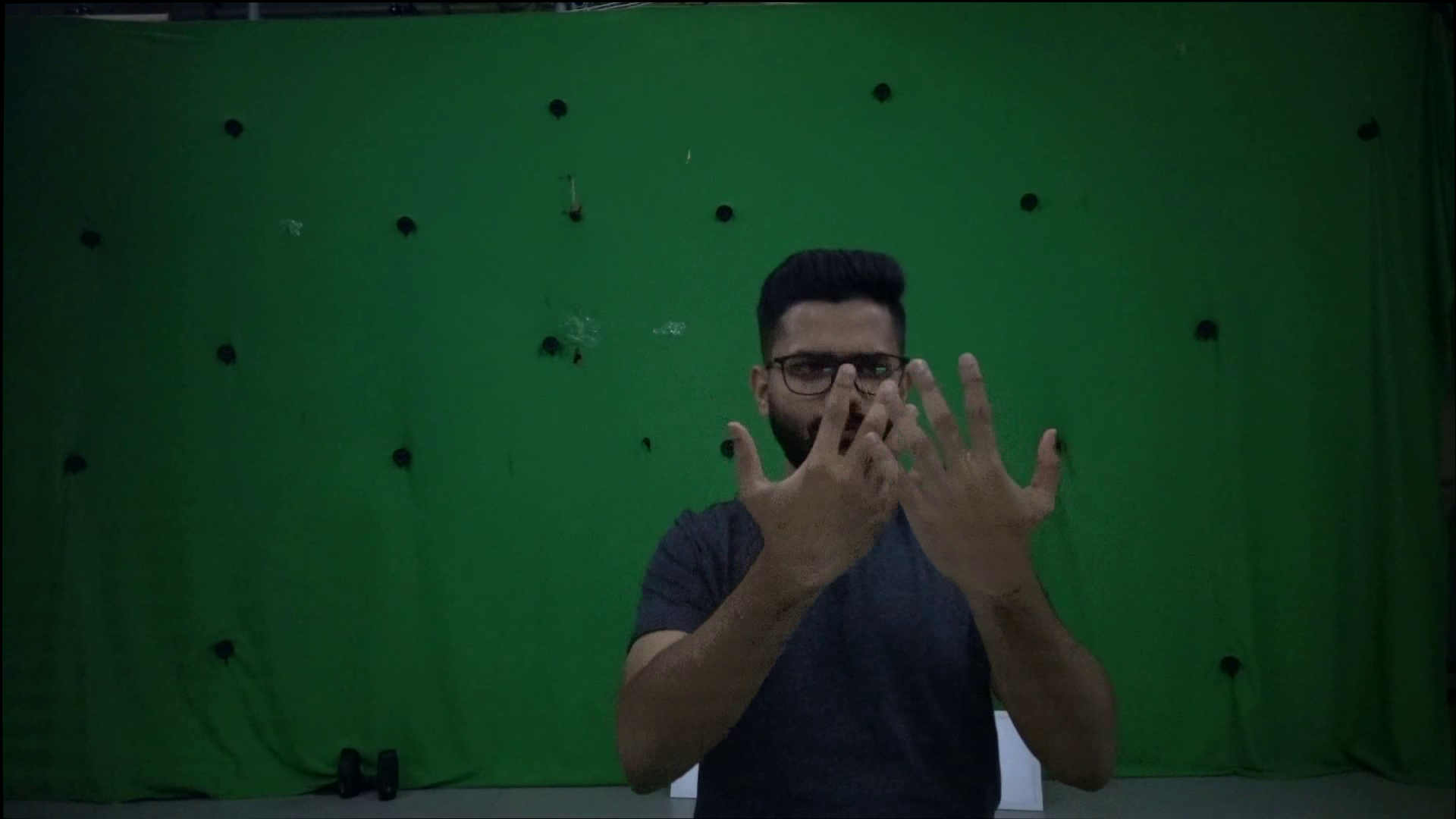}  &
\includegraphics[width=8em]{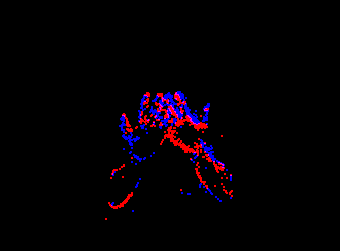} &
\includegraphics[width=8em]{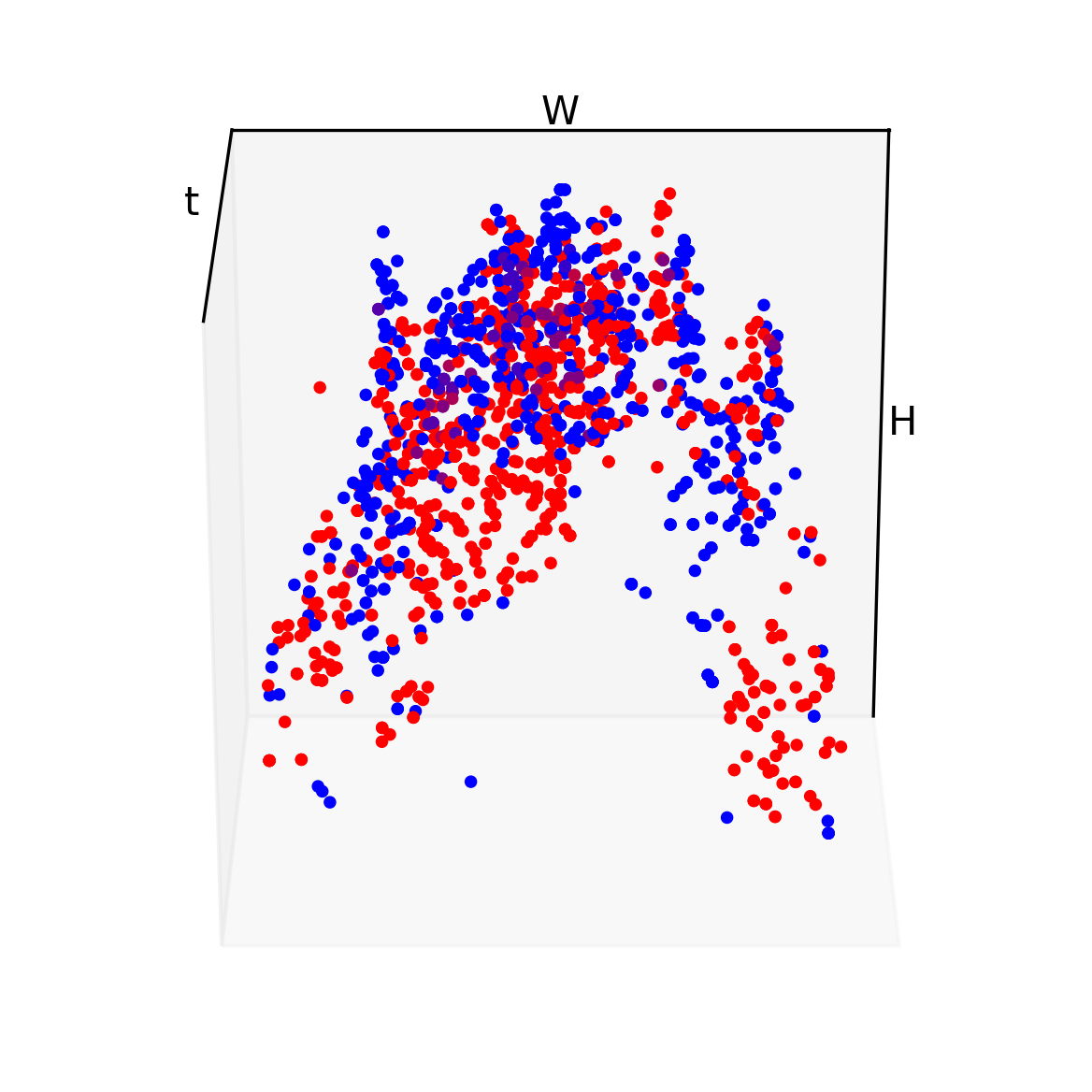} &
\includegraphics[width=8em]{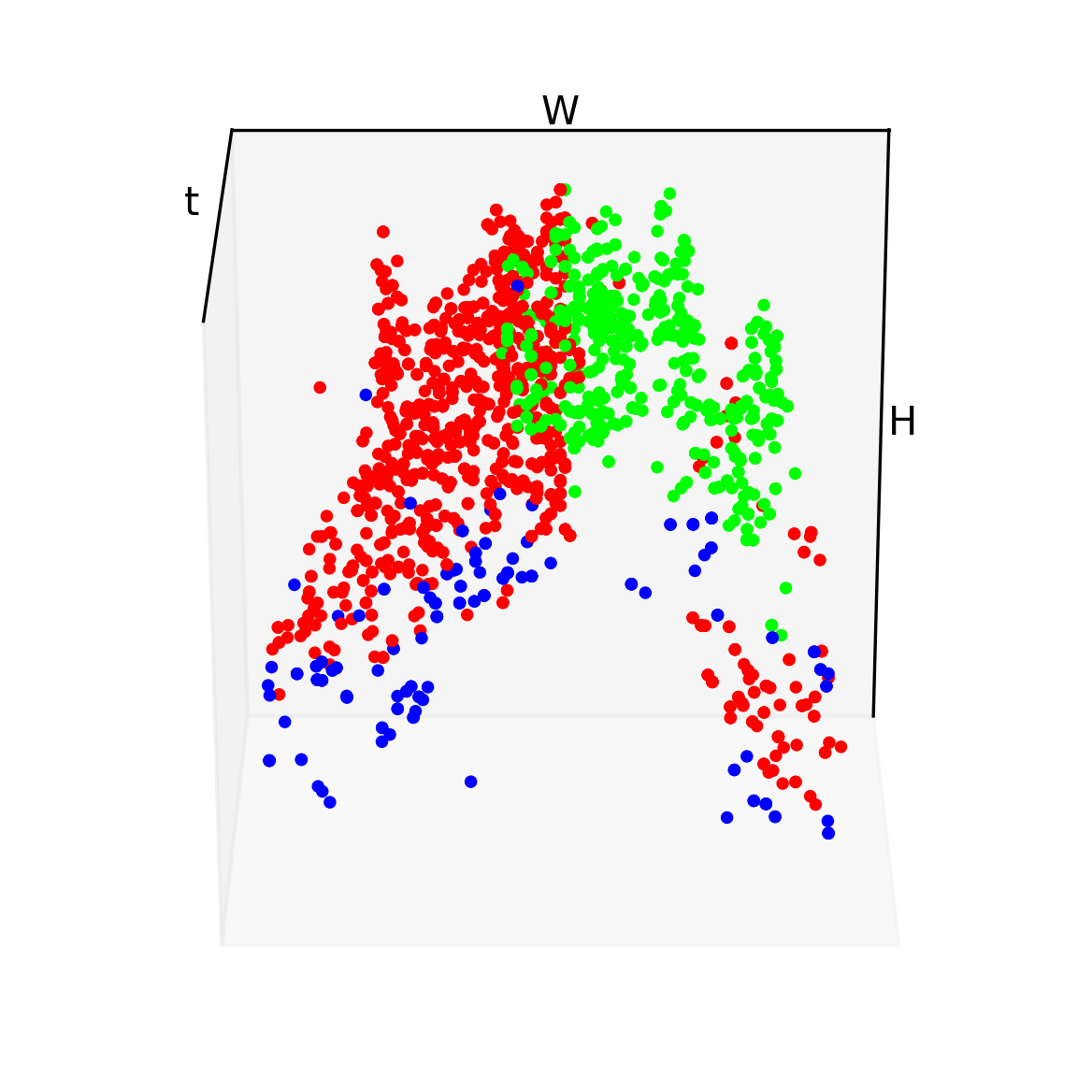} &
\includegraphics[width=8em]{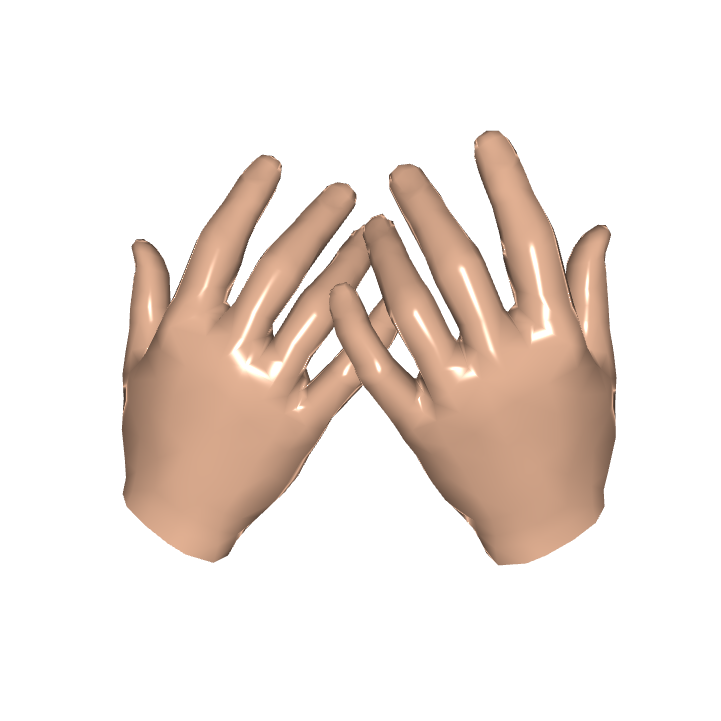} \\

\includegraphics[width=8em, trim=0 -20cm 0 0]{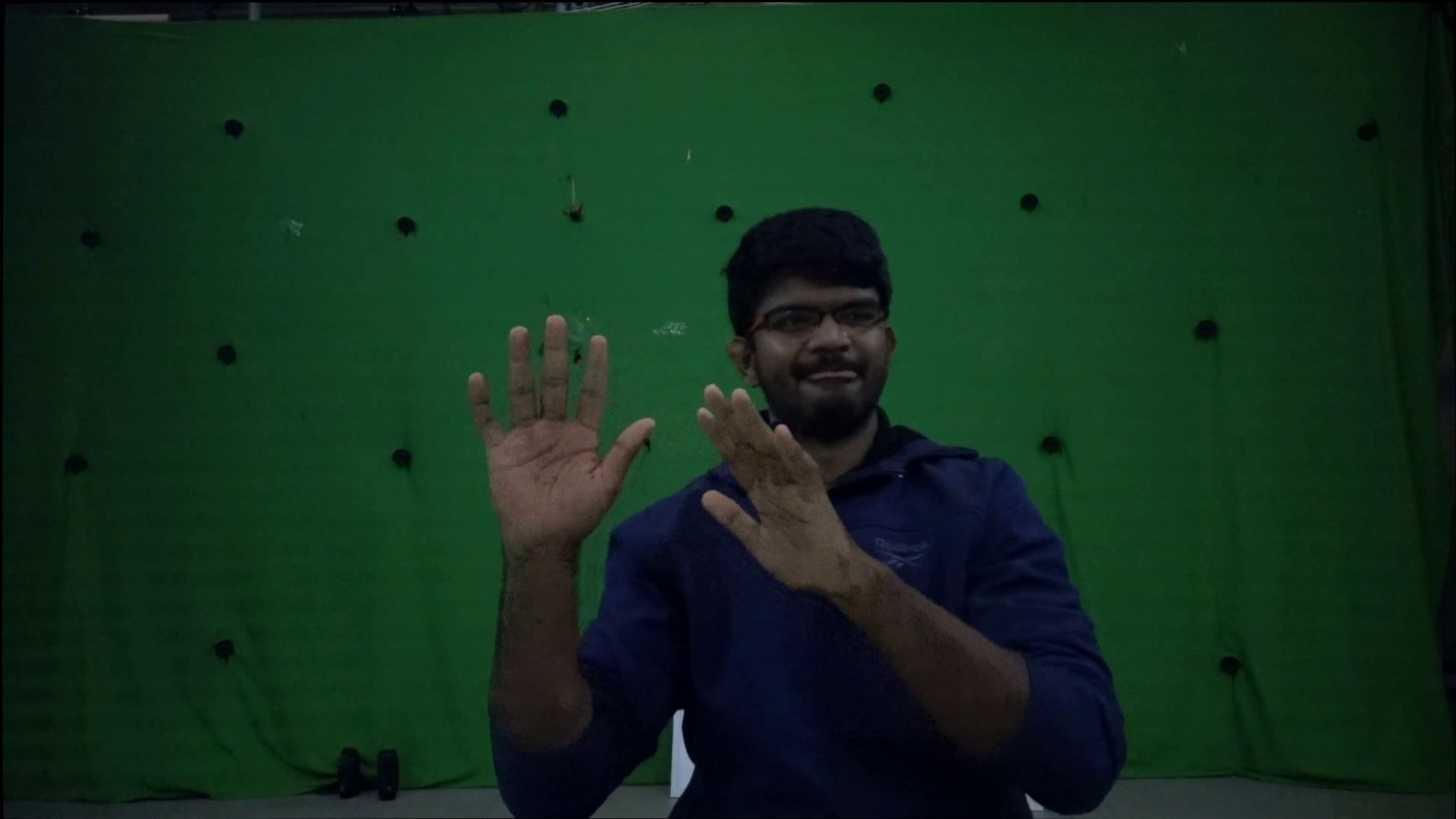}  &
\includegraphics[width=8em]{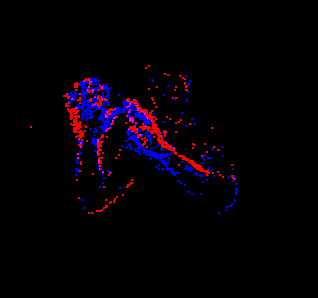} &
\includegraphics[width=8em]{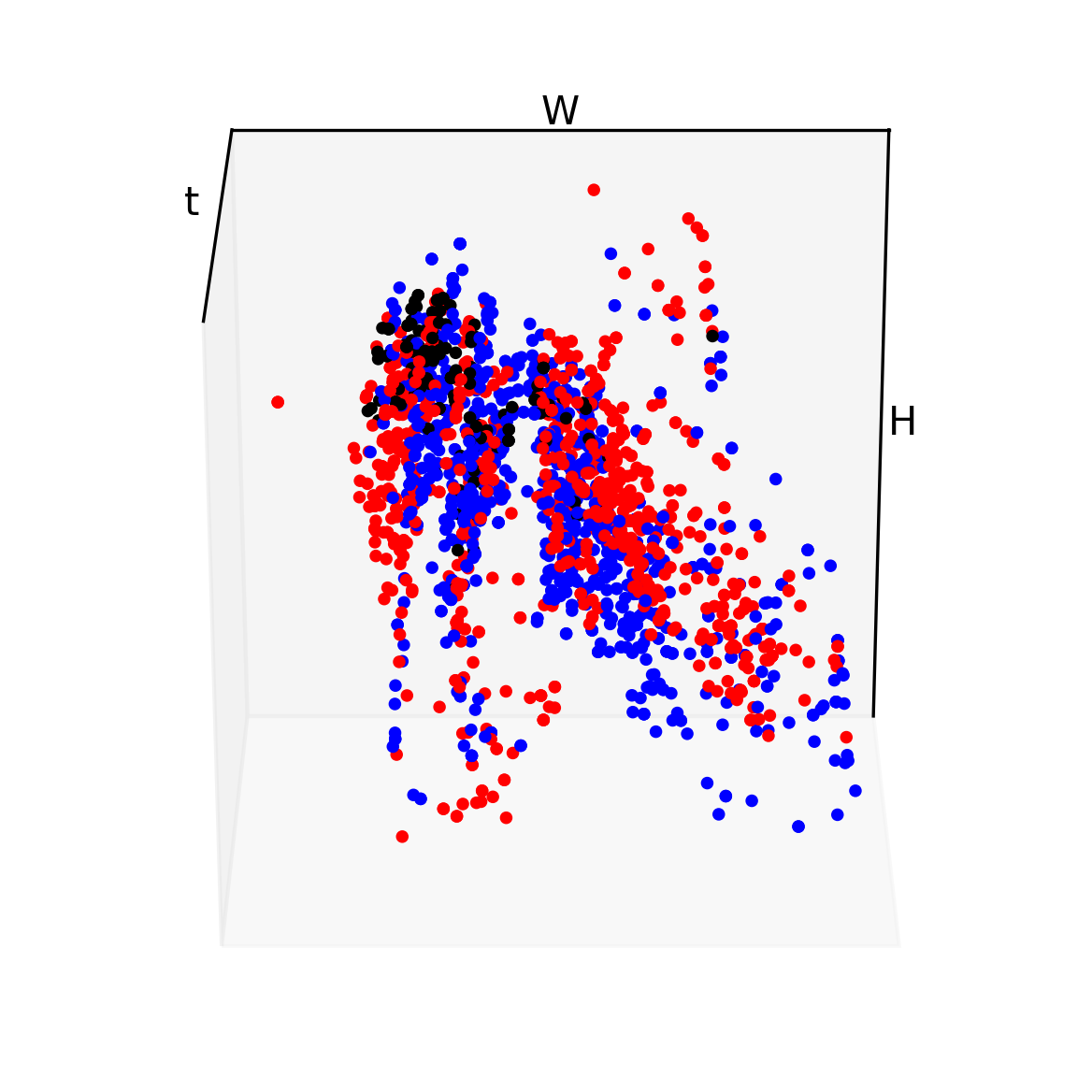} &
\includegraphics[width=8em]{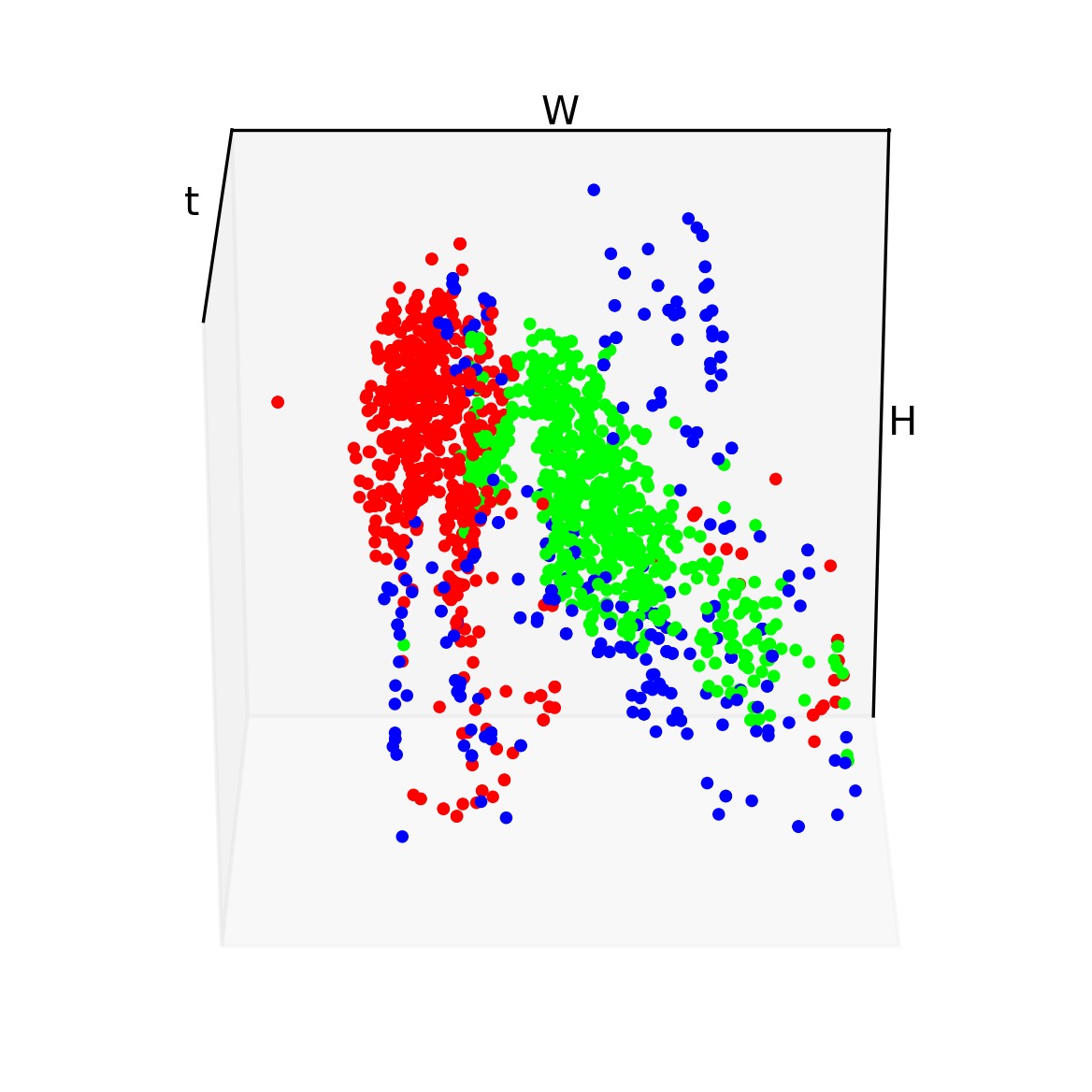} &
\includegraphics[width=8em]{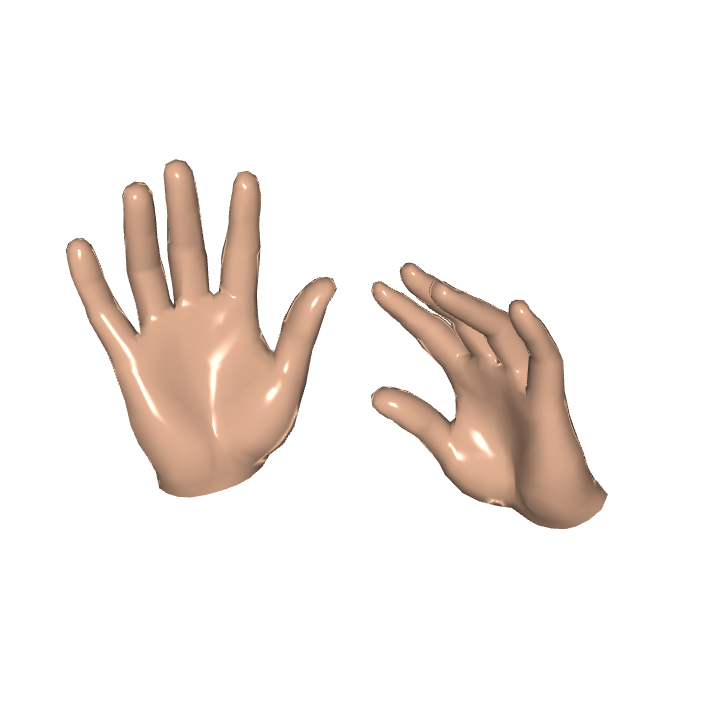} \\

\includegraphics[width=8em, trim=0 0cm 0 0]{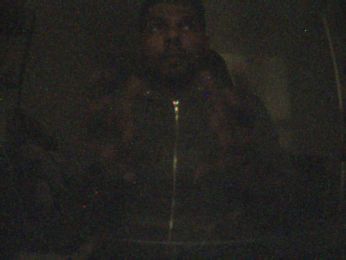}  &
\includegraphics[width=8em]{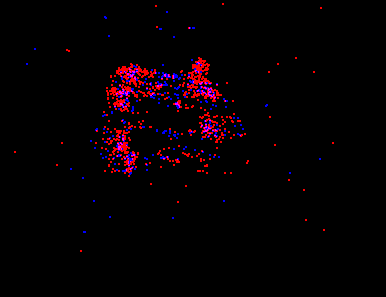} &
\includegraphics[width=8em]{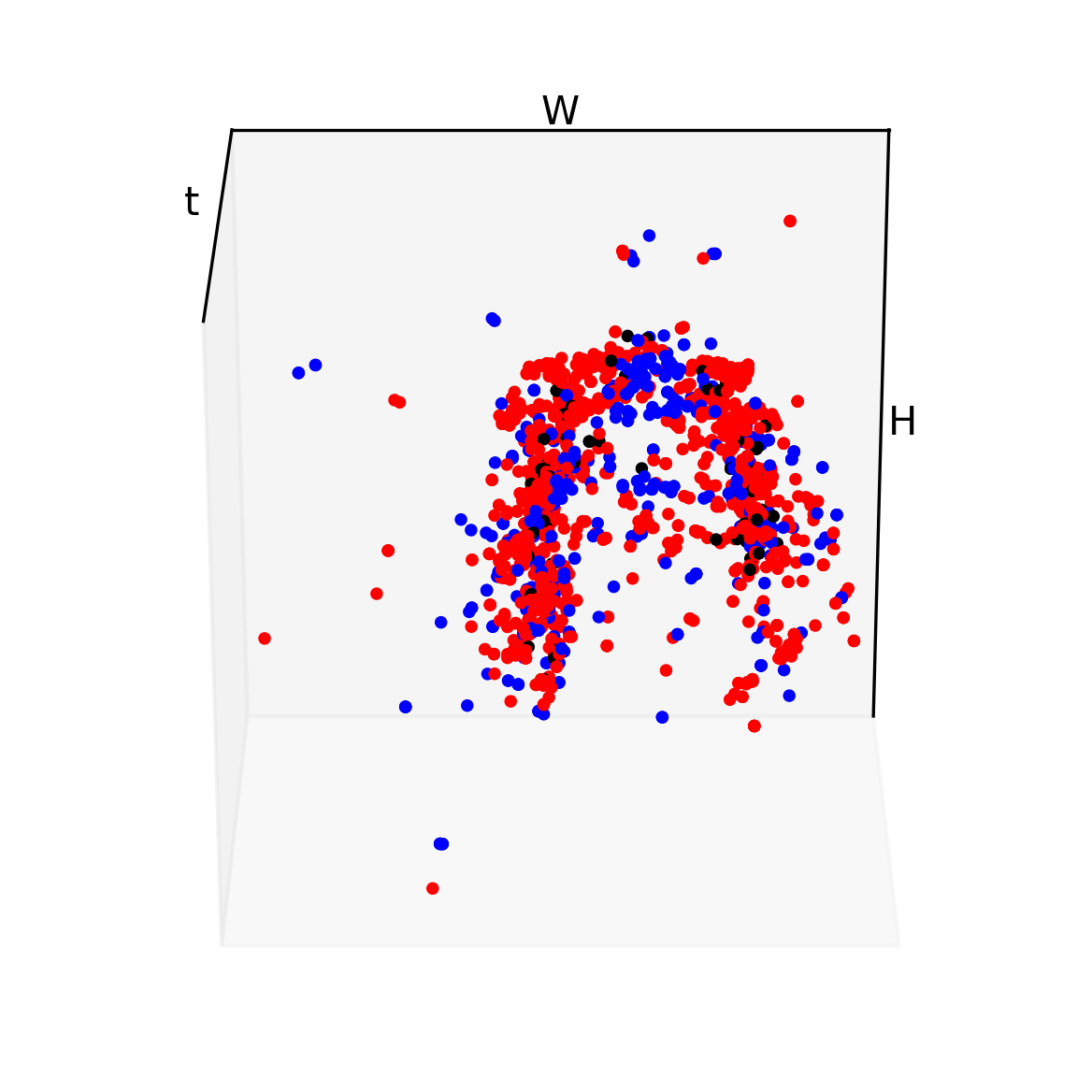} &
\includegraphics[width=8em]{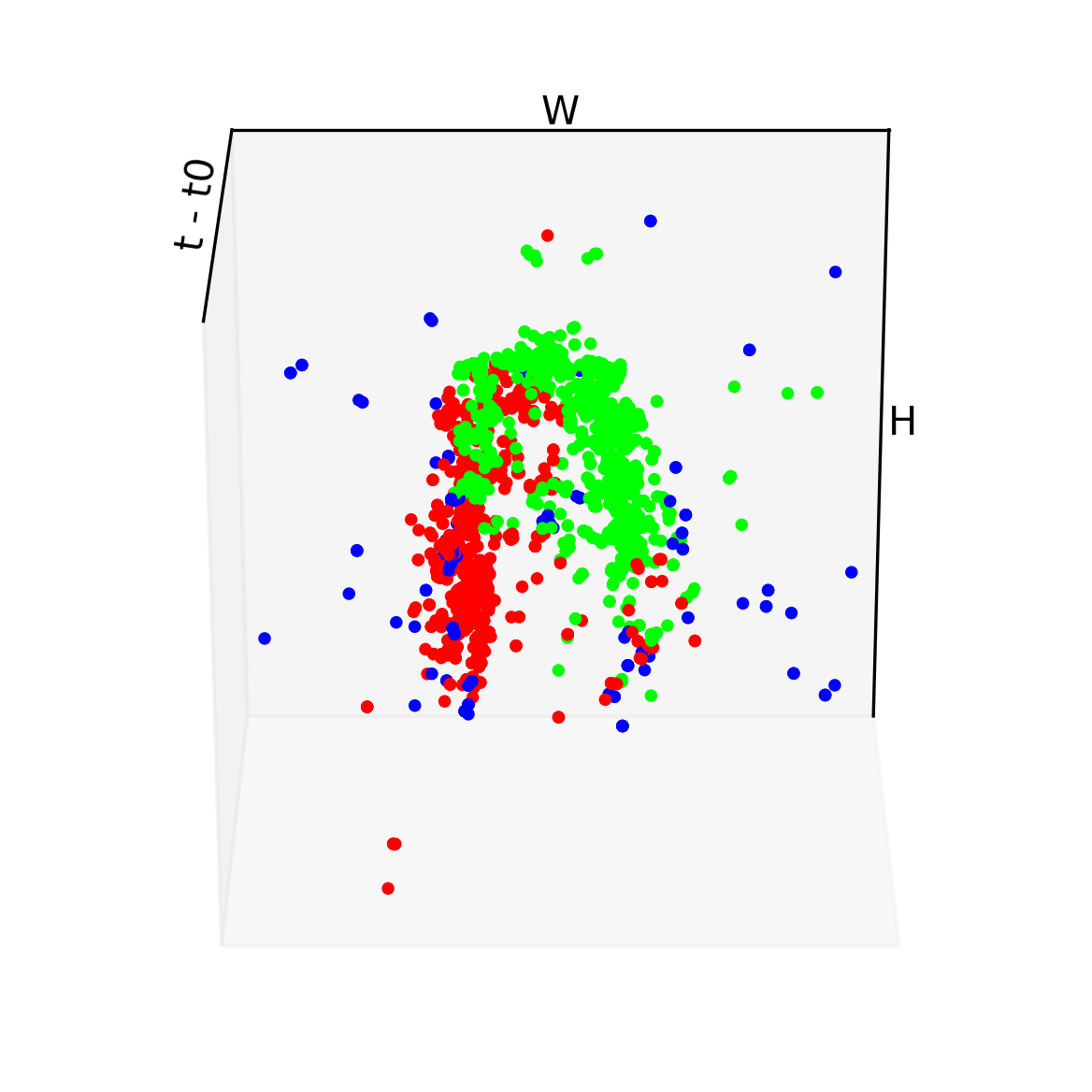} &
\includegraphics[width=8em]{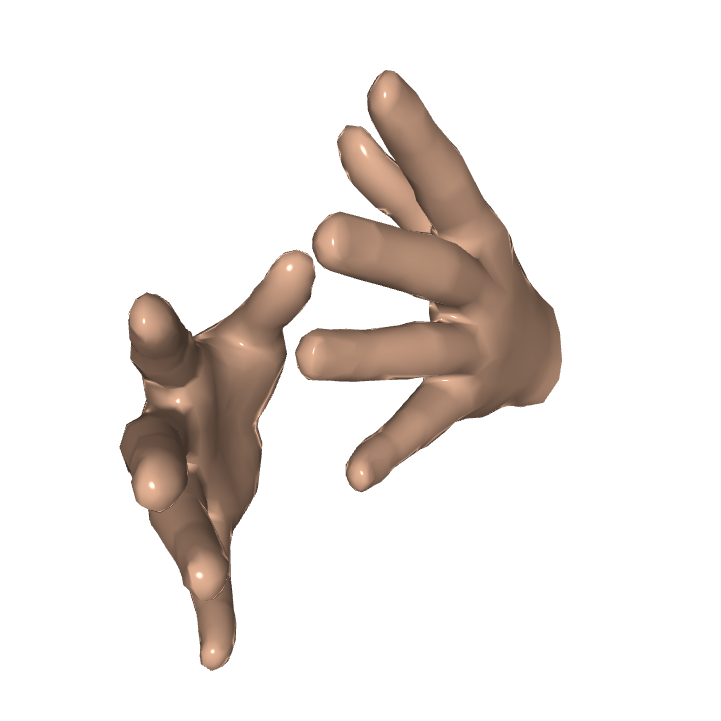} \\

\includegraphics[width=8em, trim=0 0.5cm 0 0]{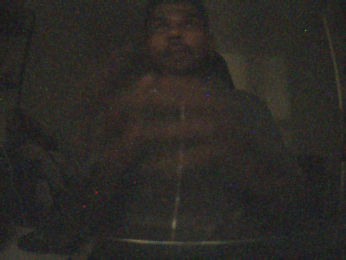}  &
\includegraphics[width=8em]{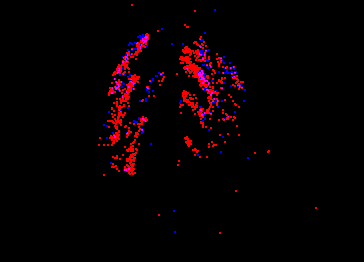} &
\includegraphics[width=8em]{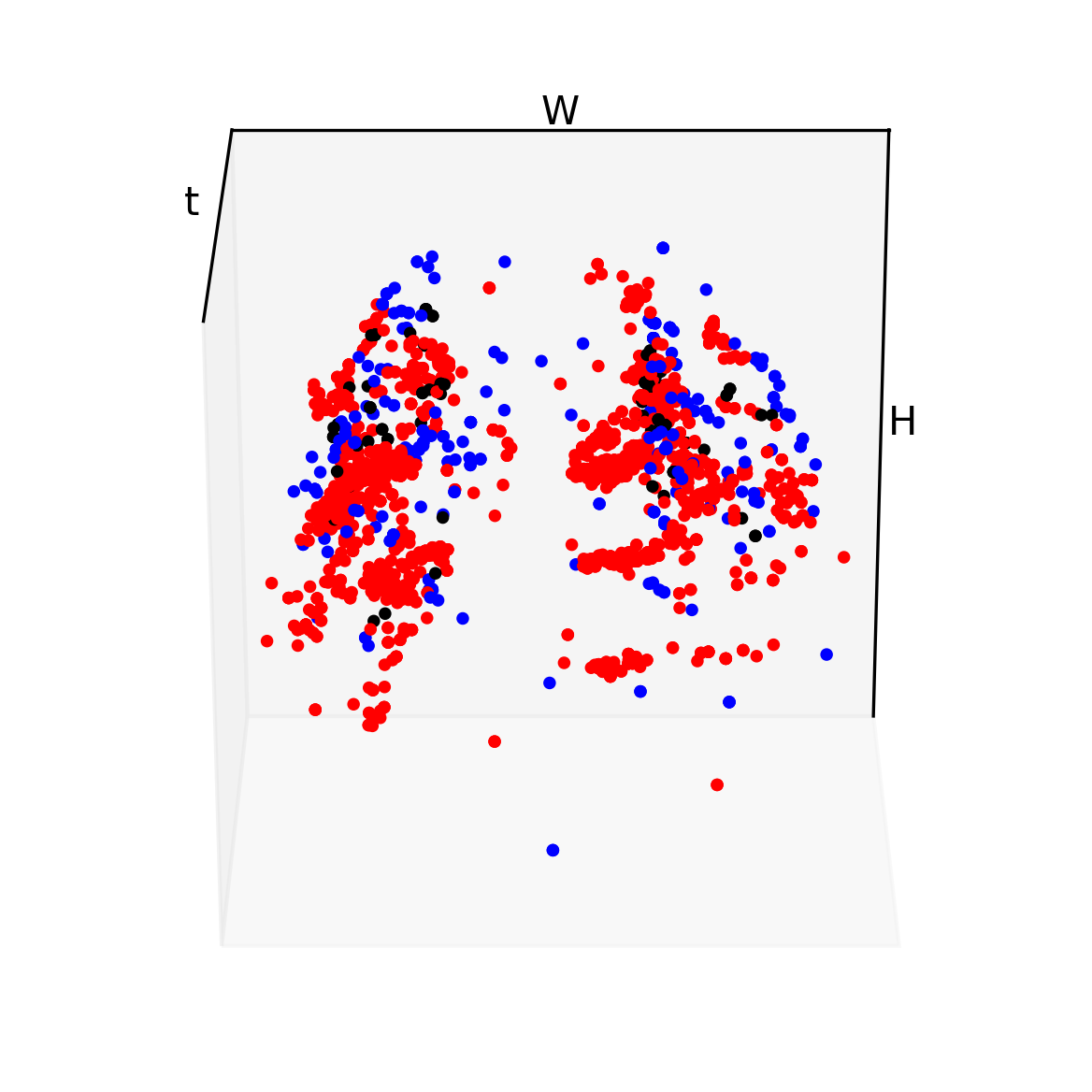} &
\includegraphics[width=8em]{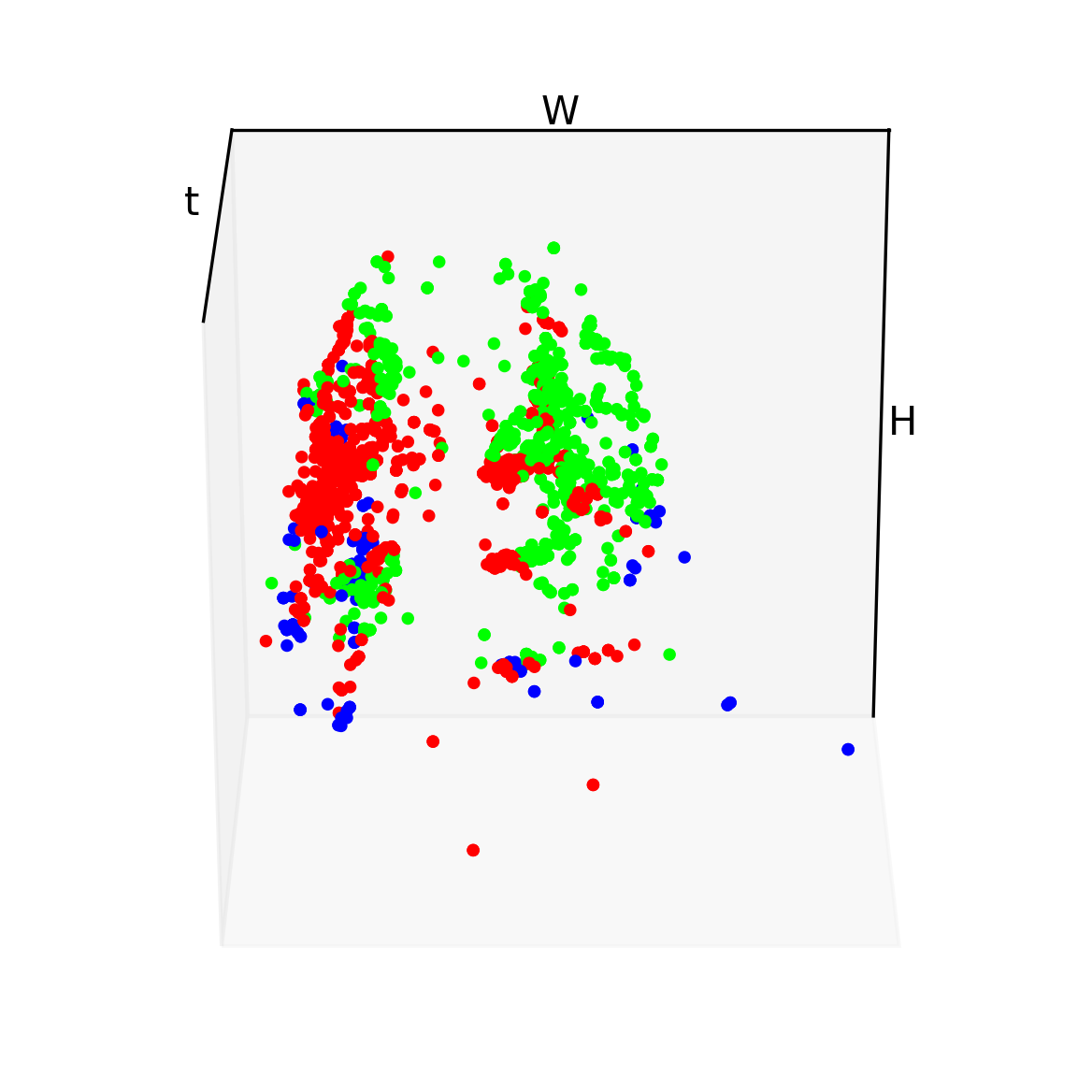} &
\includegraphics[width=8em]{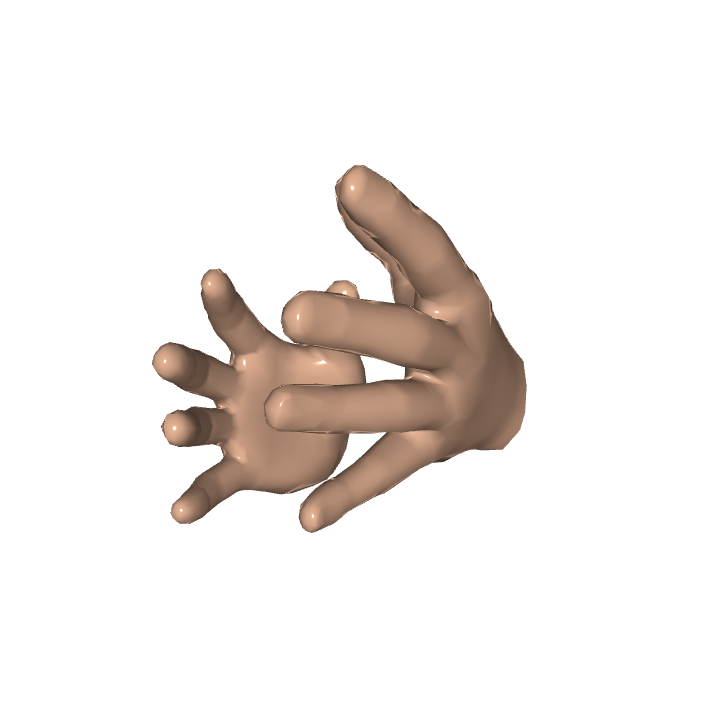} \\

\includegraphics[width=8em, trim=0 0cm 0 0]{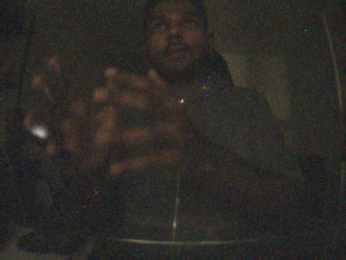}  &
\includegraphics[width=8em]{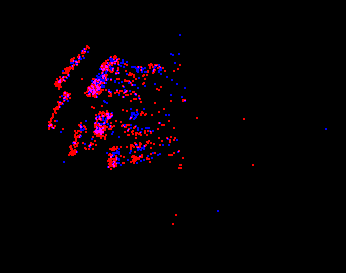} &
\includegraphics[width=8em]{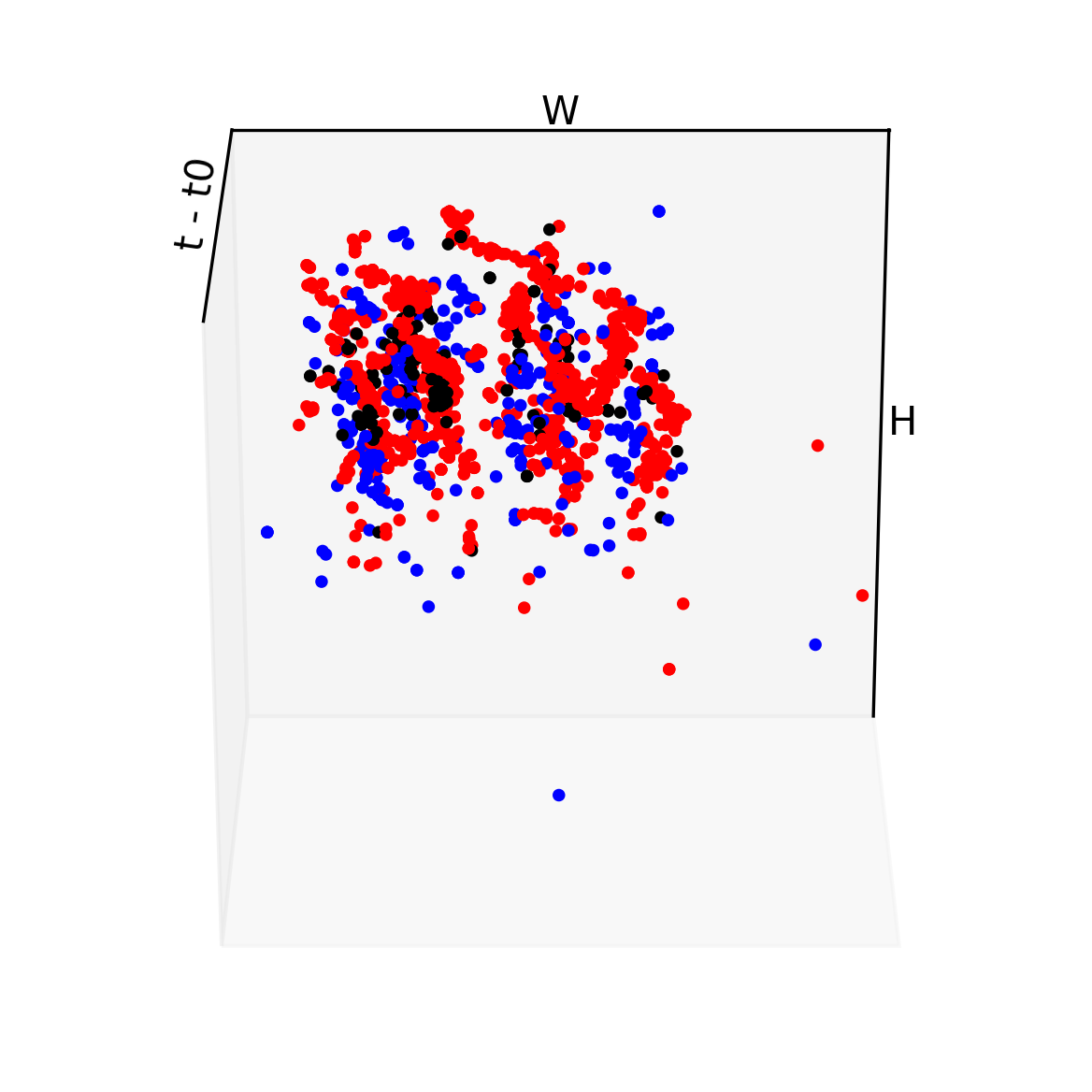} &
\includegraphics[width=8em]{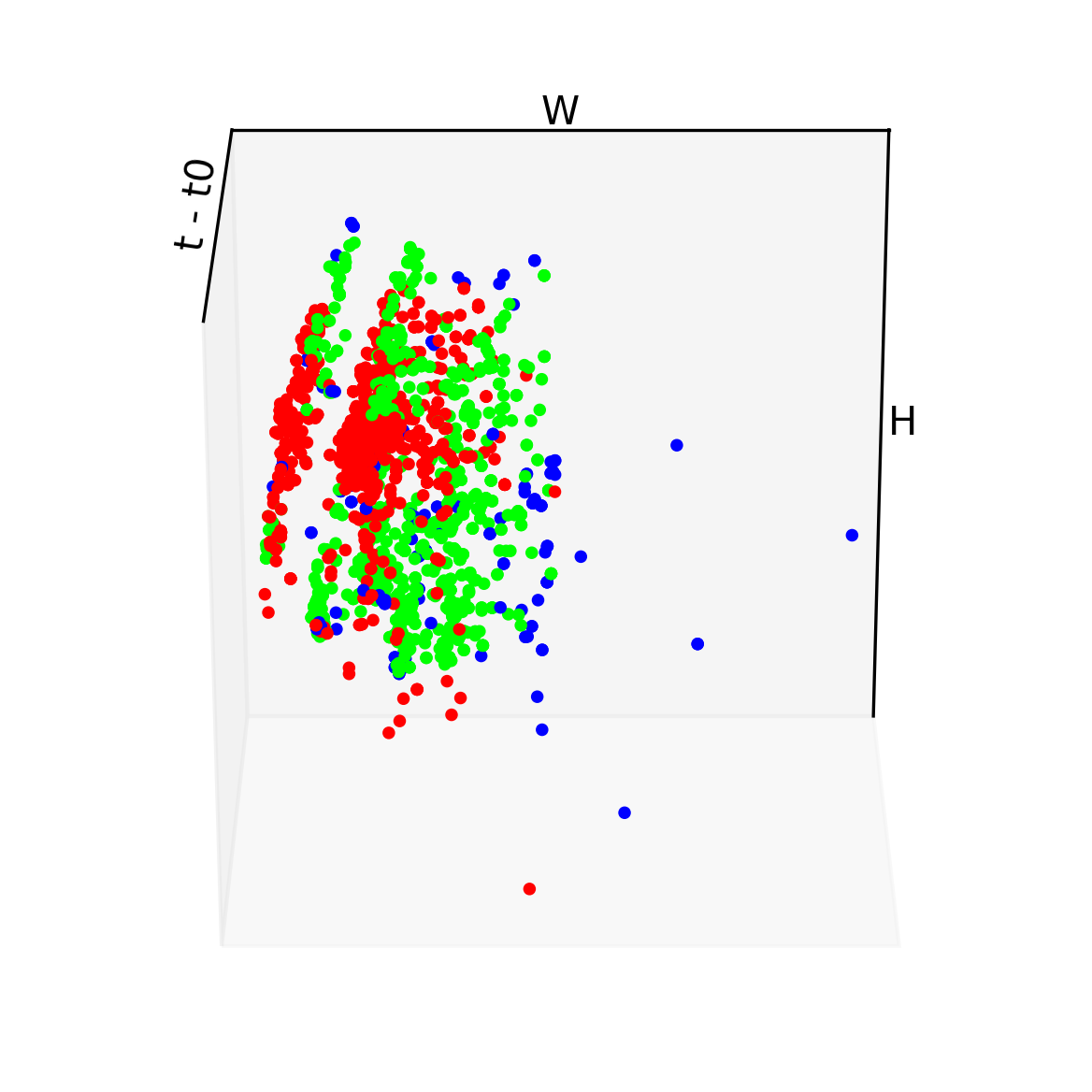} &
\includegraphics[width=8em]{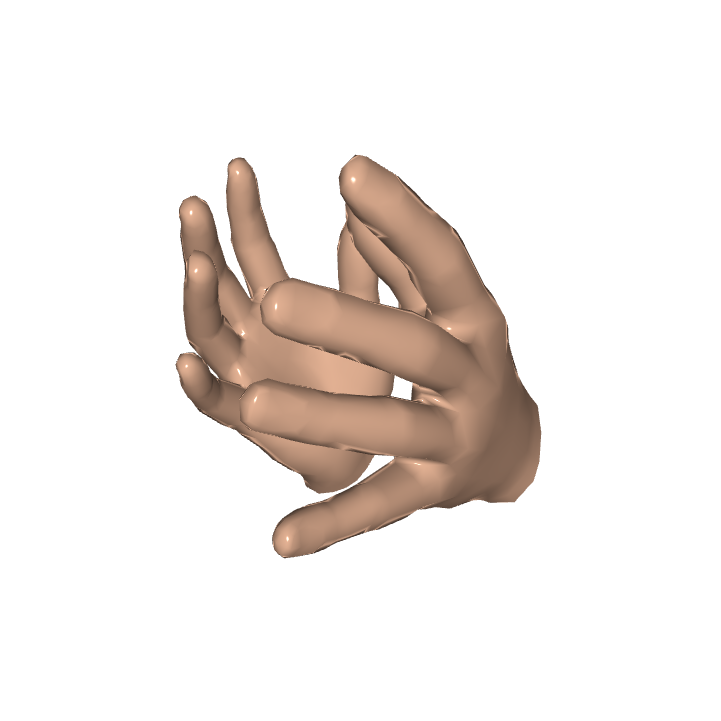} \\

\includegraphics[width=8em, trim=0 -3cm 0 0]{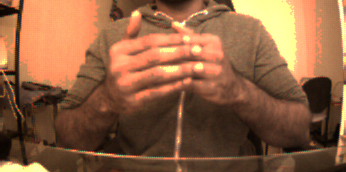}  &
\includegraphics[width=8em]{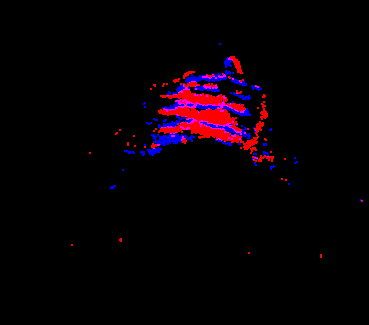} &
\includegraphics[width=8em]{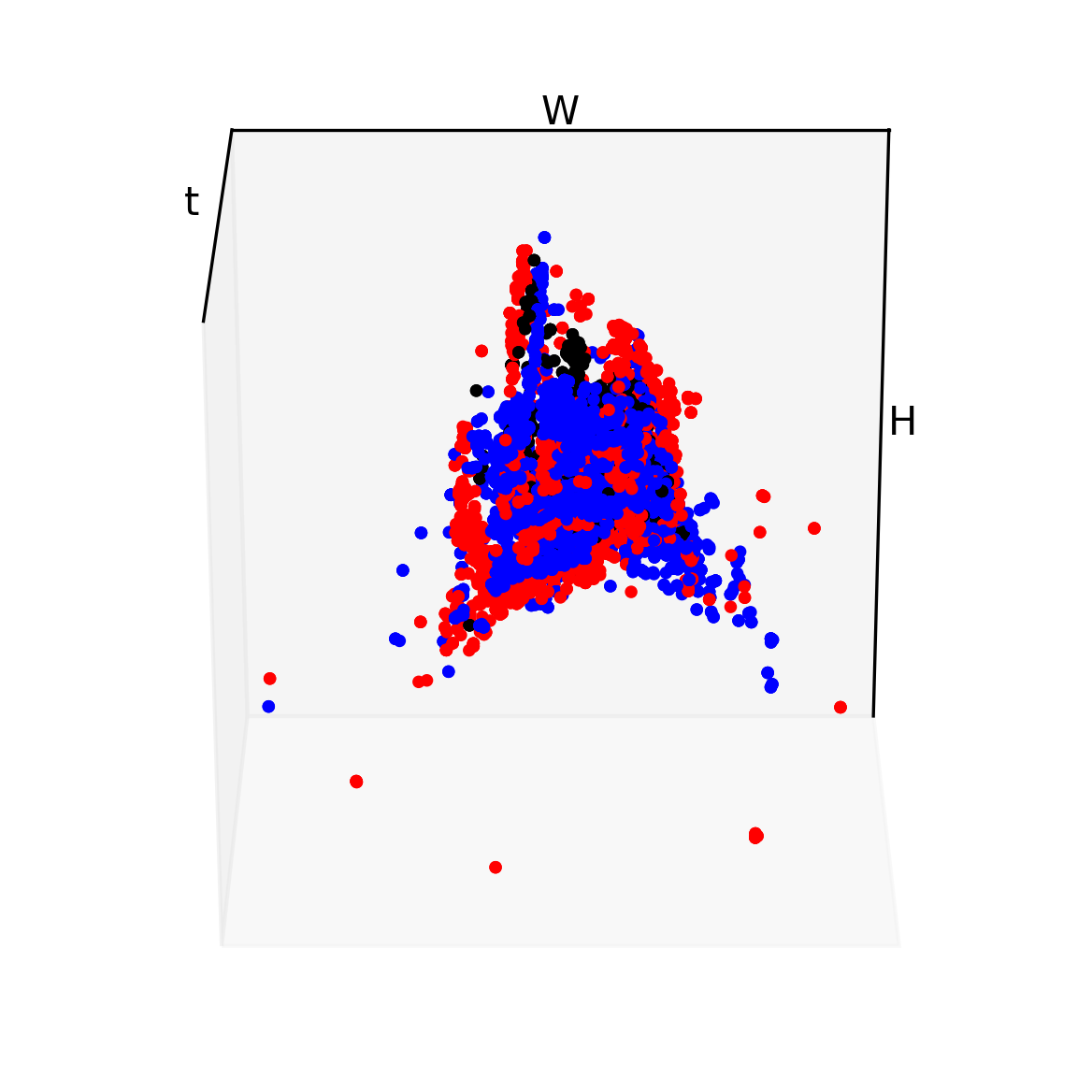} &
\includegraphics[width=8em]{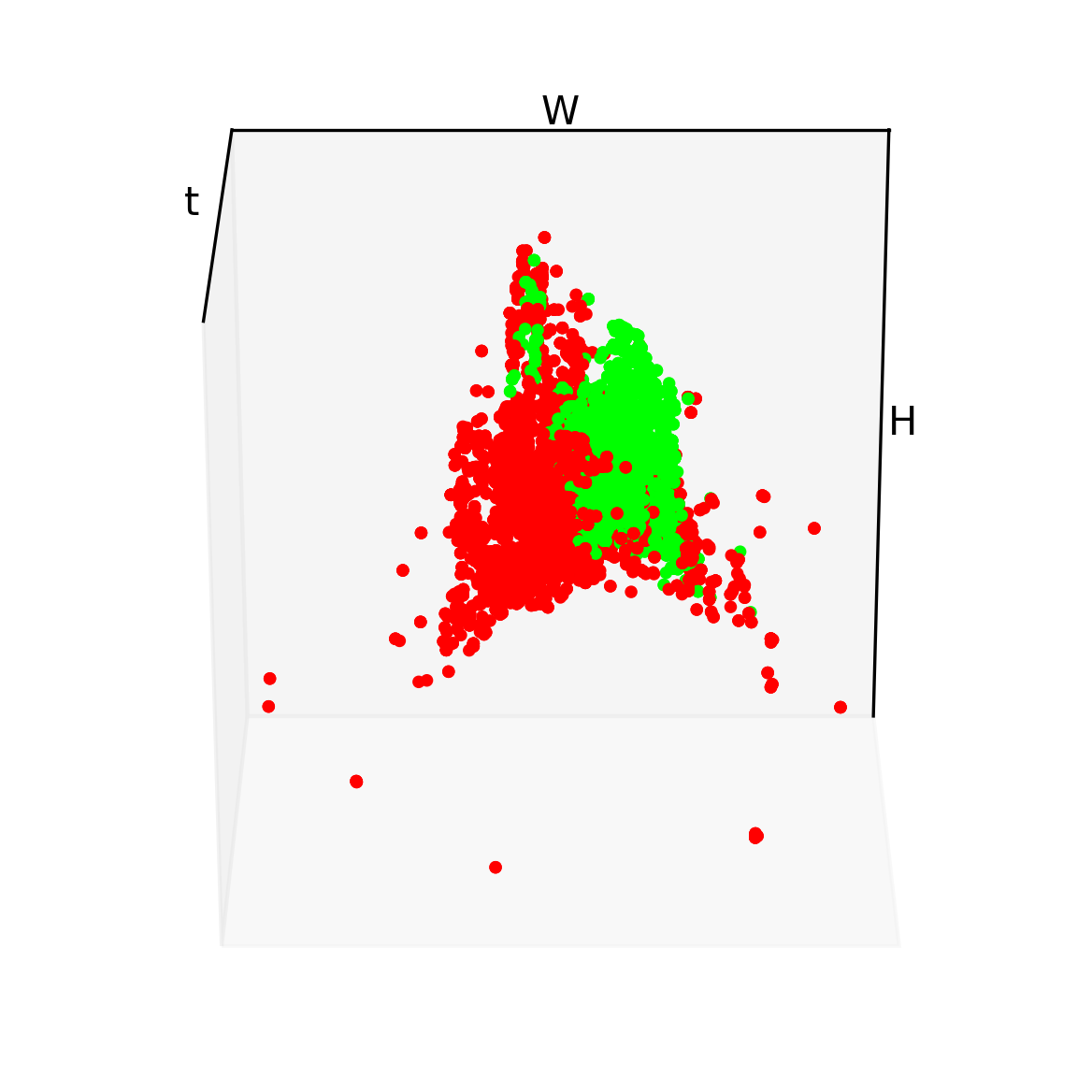} &
\includegraphics[width=8em]{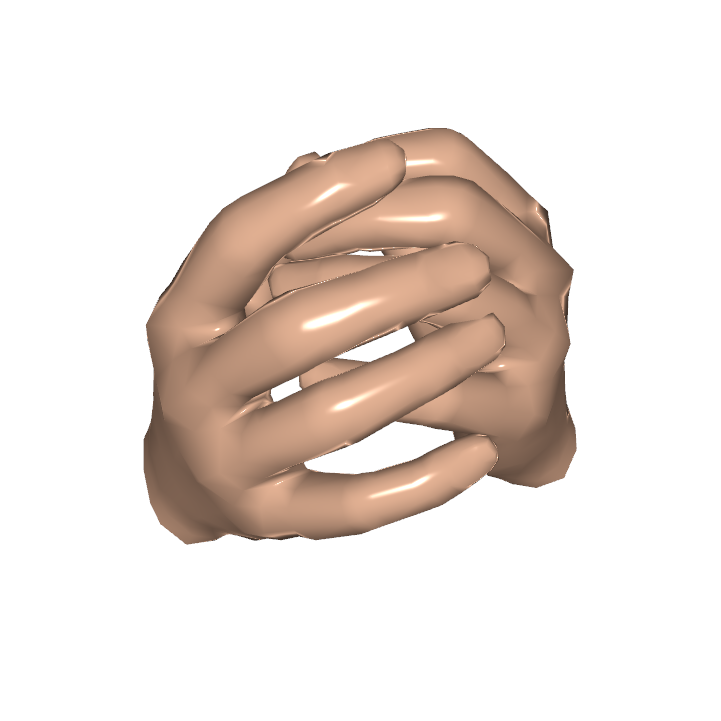} \\

\includegraphics[width=8em, trim=0 -1.1cm 0 0]{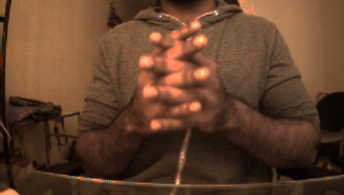}  &
\includegraphics[width=8em]{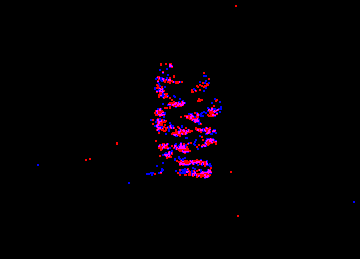} &
\includegraphics[width=8em]{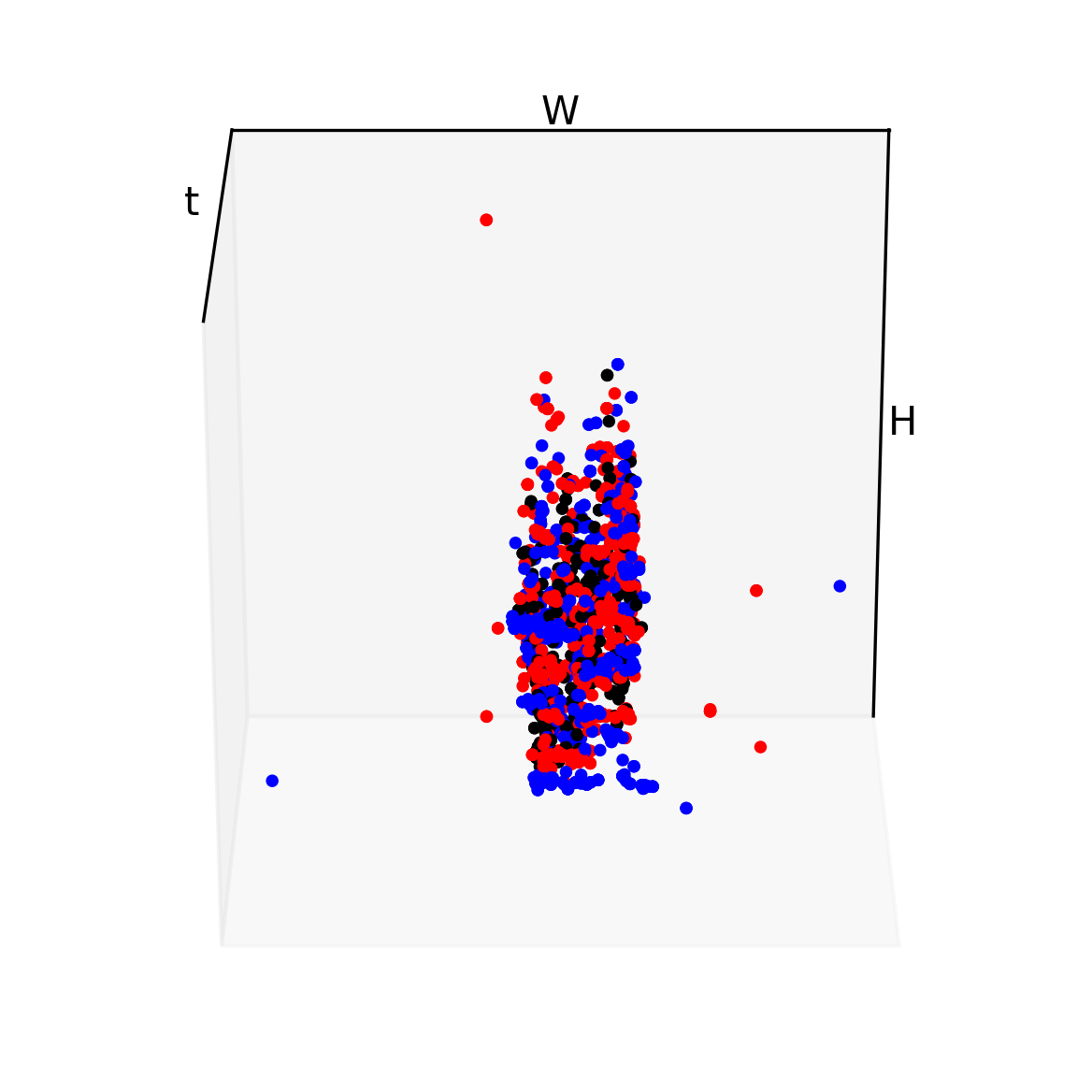} &
\includegraphics[width=8em]{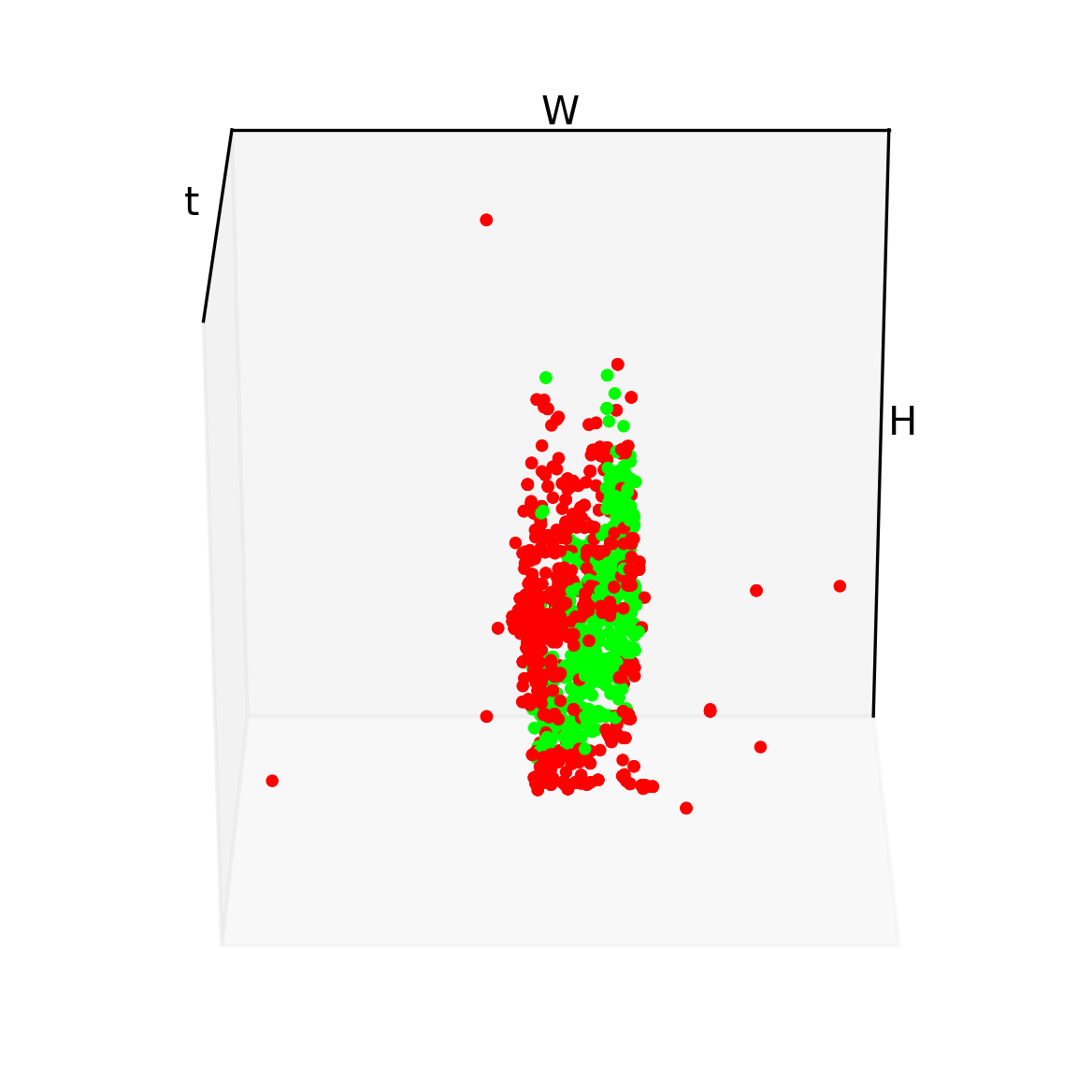} &
\includegraphics[width=8em]{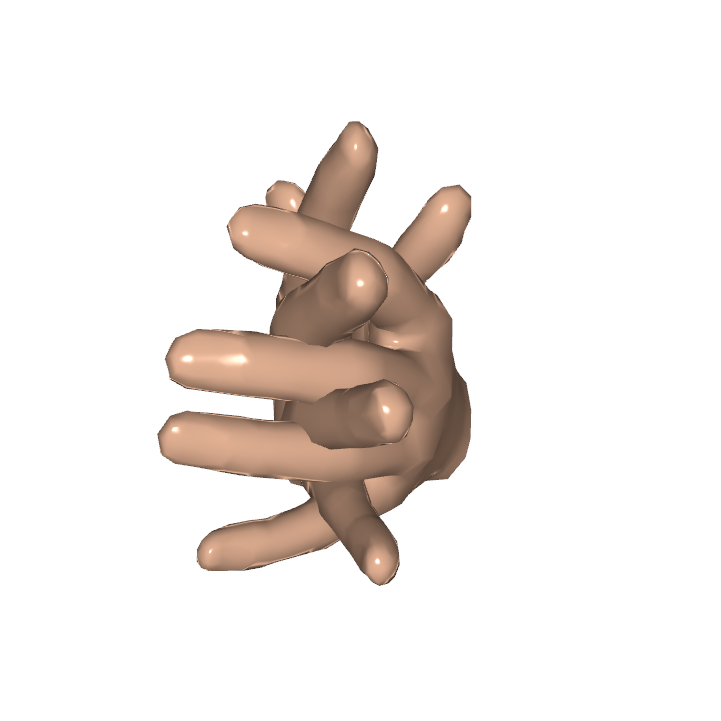} \\

\bottomrule
\end{tabular}
\caption{
Additional visualisations of Ev2Hands along with the visualisations of event clouds, segmentations and the 3D predictions.
}
\label{fig:additionalresults}
\end{table*}

\end{document}